\documentclass[a4paper,fleqn]{cas-dc}

\usepackage[authoryear]{natbib}
\usepackage{algorithm}
\usepackage{algpseudocode}
\usepackage{booktabs}
\usepackage{threeparttable}
\usepackage{graphicx,verbatim}
\usepackage{amsmath, amssymb}
\usepackage{xcolor}
\definecolor{myhexcolor0}{HTML}{FF00FF} 
\definecolor{myhexcolor1}{HTML}{c13eff} 
\definecolor{myhexcolor2}{HTML}{8679ff} 
\definecolor{myhexcolor3}{HTML}{42bdff} 
\usepackage{caption}
\usepackage{subcaption}
\usepackage{url}
\usepackage{xcolor}
\usepackage{soul}
\def\tsc#1{\csdef{#1}{\textsc{\lowercase{#1}}\xspace}}
\tsc{WGM}
\tsc{QE}

\makeatletter
\@ifundefined{iflatexml}{%
  \newif\iflatexml
  \latexmlfalse
}{}
\iflatexml
  \providecommand{\cormark}[1]{}
  \providecommand{\cortext}[2][]{}
  \providecommand{\ead}[1]{}
  \providecommand{\credit}[1]{}
  \providecommand{\printcredits}{}
  \providecommand{\affiliation}[2][]{}
  \providecommand{\shorttitle}[1]{}
  \providecommand{\shortauthors}[1]{}
\fi
\makeatother

\begin{document}
\let\WriteBookmarks\relax
\def\floatpagepagefraction{1}    
\def\textpagefraction{.001}      

\iflatexml

\title{Towards patient-specific optimization for mandibular reconstruction planning based on predicted bone-union propensity}
\author{Hamidreza Aftabi}
\author{John E. Lloyd}
\author{Amanda Ding}
\author{Benedikt Sagl}
\author{Eitan Prisman}
\author{Antony Hodgson}
\author{Sidney Fels}

\else

\shorttitle{}    

\shortauthors{}  

\title [mode = title]{Towards patient-specific optimization for mandibular reconstruction planning based on predicted bone-union propensity}



%

\author[1,2,3]{Hamidreza Aftabi}[orcid=0000-0002-2792-9050]

\cormark[1]

\ead{aftabi@student.ubc.ca}
\ead{h.aftabi@mail.utoronto.ca}

\credit{Conceptualization, Methodology, Software, Writing - original draft, Writing - review \& editing, Visualization, Data curation}

\author[1]{John E. Lloyd}
\credit{Software}

\author[4]{Amanda Ding}
\credit{Data curation}

\author[5,6]{Benedikt Sagl}[orcid=0000-0002-6739-4222]
\credit{Software}

\author[4]{Eitan Prisman}
\credit{Funding acquisition, Supervision, Writing - review \& editing}

\author[7]{Antony Hodgson}[orcid=0000-0002-3107-581X]
\credit{Funding acquisition, Supervision, Writing - review \& editing}

\author[1]{Sidney Fels}[orcid=0000-0001-9279-9021]
\credit{Funding acquisition, Supervision, Writing - review \& editing}

\affiliation[1]{organization={Department of Electrical and Computer Engineering, University of British Columbia},
            city={Vancouver},
            state={BC},
            country={Canada}}

\affiliation[2]{organization={Department of Medical Imaging, University of Toronto},
            city={Toronto},
            state={ON},
            country={Canada}}

\affiliation[3]{organization={The Hospital for Sick Children (SickKids)},
            city={Toronto},
            state={ON},
            country={Canada}}

\affiliation[4]{organization={Department of Surgery, University of British Columbia, Gordon and Leslie Diamond Health Care Centre},
            city={Vancouver},
            state={BC},
            country={Canada}}

\affiliation[5]{organization={Competence Center Artificial Intelligence in Dentistry, University Clinic of Dentistry, Medical University of Vienna},
            city={Vienna},
            country={Austria}}

\affiliation[6]{organization={Comprehensive Center Artificial Intelligence in Medicine, Medical University of Vienna},
            city={Vienna},
            state={Vienna},
            country={Austria}}

\affiliation[7]{organization={Department of Mechanical Engineering, University of British Columbia},
            city={Vancouver},
            state={BC},
            country={Canada}}

\cortext[1]{Corresponding author}

\fi


\begin{abstract}
Mandibular reconstruction with vascularized bone grafts is complicated by donor–host nonunion, and current virtual surgical planning produces a geometric plan rather than optimizing for bone-union propensity at the donor–host interface. We present \textit{OsteoOpt++}, an image-to-decision planning loop for patient-specific mandibular reconstruction. A pre-operative computed tomography (CT) is converted into a personalized digital twin through template-to-patient registration and CT-derived updates of the muscle and temporomandibular-joint parameters. Bayesian optimization with an expected-improvement-plus acquisition rule then searches six clinically controllable cut-plane and donor-positioning variables under an apposition-driven objective and a safety-factor-regularized variant. The workflow was evaluated on three generic defects (body, symphysis, and ramus-body) and four patient-specific cases, three of which were used for optimization and all four for retrospective longitudinal spatial analysis. In the generic cases, against the surgeon's geometric plan, cycle-averaged donor-mandible apposition increased by up to 29 percentage points; in the patient-specific cases, against the surgeon-implemented day-5 postoperative configuration, by up to 26 percentage points. A \(\pm10\%\) sensitivity analysis over eleven modeling parameters capped the change in the apposition-driven objective at \(\sim\)3\% (generic) and \(\sim\)4\% (patient-specific), and across the four longitudinal cases the Dice overlap between predicted apposition and year-1 bone formation ranged from 70.1\% to 84.9\%, with centroid shifts of 0.24 to 1.82~mm. Together, these results support the feasibility-stage use of \textit{OsteoOpt++} to compare candidate reconstructions using apposition-derived predictions of bone-union propensity. The optimization and patient-specific modeling code is open source at \url{https://github.com/hamidreza-aftabi/OsteoOpt}.
\end{abstract}





\maketitle

\section{Introduction}\label{sec:introduction}

Head and neck cancer remains a major clinical burden. When tumors involve the mandible, segmental bone resection may be required to obtain adequate surgical margins~\citep{kumar2016mandibular,li2023current}. These resections can create discontinuity defects that disrupt facial contour, mandibular function, and the donor-host interface required for bone union. Mandibular reconstruction aims to restore jaw continuity, facial form, and long-term functional rehabilitation~\citep{goh2008mandibular,brown2017mandibular,urken1991oromandibular}. Microvascular autologous bone transfer, most commonly using vascularized fibula or scapula grafts, remains a central reconstructive option as it provides viable bone for integration~\citep{disa2000mandible,brown2017mandibular,aftabi2024computational}. Nevertheless, nonunion or partial union at the graft-host interface remains a persistent complication, with reported rates reaching up to 37\% in some cohorts~\citep{swendseid2020natural,sabiq2024evaluating}. These failures can lead to pain, impaired mastication, delayed rehabilitation, and revision surgery~\citep{hundepool2008rehabilitation}. Recent biomechanical and clinical studies have shown that the geometry of the reconstruction and resulting donor-host interface is a major determinant of bone union~\citep{wong2010biomechanics,kim2024optimizing,sabiq2024evaluating}. However, the systematic and automated optimization of clinically controllable geometric variables to promote bone union remains largely unexplored. Current virtual surgical planning (VSP) has mainly emphasized geometric reconstruction and donor-to-contour shape matching~\citep{nakao2017automated,guo2025automated} as an initial planning step, rather than optimizing the cut-plane and donor variables that shape the interface conditions relevant to predicted bone-union propensity.

VSP has substantially improved maxillofacial reconstruction by embedding medical-image-derived anatomy into the preoperative workflow. CT-based segmentation, three-dimensional reconstruction, patient-specific guide design, and planned transfer of donor segments have improved the reproducibility and efficiency of surgical execution~\citep{hanasono2013computer,nguyen2021maxillectomy,vyas2022virtual}. Early clinical studies also reported reductions in operative time and improved surgical transfer using VSP~\citep{jacek20183d,gil2015surgical}, and guided planning can improve segment positioning and bony contact~\citep{wang2016mandibular}. Yet the primary output of most VSP pipelines remains a geometric plan~\citep{tran2023virtual}. Two plans can appear equally feasible and visually acceptable while producing different donor-host contact, donor positioning, and factors affecting bone-union propensity. Accurate transfer of a geometric plan is essential, but it does not by itself establish that the plan is favorable for donor-host union. The surgically controllable variables driving these interface conditions include resection-plane orientation, donor positioning, segment arrangement, and the donor–native interface. These variables are typically assessed through visual fit, surgical experience, and geometric feasibility, with no patient-specific way to prospectively compare feasible alternatives in terms of donor–host union. A successful reconstruction must therefore do more than reproduce the mandibular outline; it must also create conditions that support donor-host bone union.

This motivates adopting an image-driven, simulation-based optimization perspective. Computed tomography (CT) encodes more than a surface for guide fabrication: it provides patient mandibular anatomy, defect geometry, donor geometry, and data on bone composition, all of which can define the reconstruction geometry, feasible planning variables, patient-specific parameters, and objective functions used to evaluate candidate plans. While deep learning and generative AI have transformed data-rich medical-image-analysis tasks~\citep{kazerouni2023diffusion,dayarathna2024deep}, mandibular reconstruction planning is fundamentally different: each patient receives a single realized surgical configuration, and the relevant alternative cut-plane and donor-position choices remain unrealized. Donor-host bone formation is moreover an inherently longitudinal phenomenon, unfolding over months of postoperative remodeling rather than being directly observable from the immediate postoperative image. A patient-specific, physics-based digital twin grounded in bone-remodeling theory can fill this gap by evaluating hypothetical reconstructions generated from the patient's images, assigning interpretable surrogate scores for predicted bone-union propensity, and searching variables that the surgeon can control. This does not replace data-driven image analysis but solves a complementary problem: comparing possible plans when the alternative outcomes are unavailable as training labels.

Several computational studies have investigated bone formation, remodeling, and reconstruction-related outcomes using image-based or simulation-based models~\citep{field2010prediction,zheng2019investigation,zheng2022,wan2022interaction,ferguson2022,wu2021machine,bettin2026identifying}. Related subject-specific mandible and patient-specific temporomandibular joint (TMJ) modeling studies have also shown that simulation can estimate functional quantities that are difficult to measure directly in vivo~\citep{guo2022emg,sagl2019dynamic,ahmadi2026computational}. These studies provide an important foundation for evaluating reconstructed anatomy. However, as they generally analyze a fixed reconstruction, implant, plate, simplified contact model, or generic anatomy, they do not yet provide the systematic optimization and patient-specific modeling pipeline needed for surgical planning. In particular, existing approaches do not provide a CT-driven workflow that uses patient anatomy and donor information to search clinically controllable reconstruction variables across alternative surgical plans. The unresolved planning question is therefore how to close this image-to-decision loop before surgery.

Building on our prior craniofacial reconstruction model and Bayesian optimization workflow~\citep{aftabi2024extent,aftabi2025optimizing,aftabi2025osteoopt}, we introduce \textit{OsteoOpt++}, an image-to-decision planning loop for patient-specific mandibular reconstruction that combines automated optimization with patient-specific physics-based and finite-element modeling. The pipeline converts CT-derived anatomy and donor information into a patient-specific digital twin, evaluates candidate reconstructions using surrogate objective functions for predicted bone-union propensity, and uses Bayesian optimization to search clinically controllable reconstruction variables. In the retrospective patient-specific setting, pre-op CT is used to construct the planning model, day-5 post-op CT is used to recover the realized surgical baseline implemented by the surgeon, and year-1 postoperative CT is used for retrospective spatial comparison with observed bone-formation patterns. \textit{OsteoOpt++} thereby extends geometric VSP toward patient-specific, image-driven optimization while keeping the model outputs interpretable and tied to apposition-based surrogate metrics. Accordingly, the workflow is framed as a feasibility study for retrospective planning and longitudinal consistency analysis rather than as a validated prospective clinical tool.

To assess this image-to-decision workflow, we evaluate \textit{OsteoOpt++} on the three most frequent defect classes using three generic/synthetic cases and three real patient-specific cases selected to match those defect patterns. Each optimization case is evaluated under two objective functions and repeated across multiple optimization runs, allowing the generic and patient-specific optimization behavior to be compared directly. Retrospective longitudinal spatial analysis was performed on four patient-specific cases: the three patient-specific optimization cases and a fourth patient. Because postoperative MRI was available only for the fourth patient, that case alone was used for the MRI-to-CT assessment of the temporomandibular-joint disc approximation.

The main contributions of this work are:
\begin{enumerate}
    \item \textit{OsteoOpt++}, a patient-specific Bayesian optimization workflow that moves mandibular reconstruction planning beyond static geometric contour-matching to optimization under dynamic, functional loading. Established solvers are reused; the methodological contribution is an image-to-decision framework that automatically regenerates, remeshes, and dynamically simulates each candidate reconstruction, then ranks it using complementary objectives that promote high, balanced interface apposition while penalizing mechanically unsafe loading. The optimization and patient-specific modeling code is open source at \url{https://github.com/hamidreza-aftabi/OsteoOpt}.
    \item A CT-driven registration pipeline that turns patient imaging into a dynamic functional digital twin. It transfers muscle and ligament attachments, derives subject-specific muscle-force and bone-material maps from CT, and introduces an anatomy-guided TMJ disc and capsule prior reconstructed from CT alone and compared against magnetic resonance imaging (MRI), recovering soft tissue not visible on CT.
    \item A systematic generic-versus-patient-specific evaluation across the three most frequent defect classes, showing consistent optimization behavior under two objectives and repeated runs.
    \item A retrospective longitudinal image-based spatial analysis comparing day-5 predicted apposition against year-1 bone formation, with sensitivity analysis and an MRI-to-CT assessment of the TMJ-disc approximation.
\end{enumerate}

\begin{figure*}[t]
    \hspace{-1cm}
    \includegraphics[width=1\textwidth]{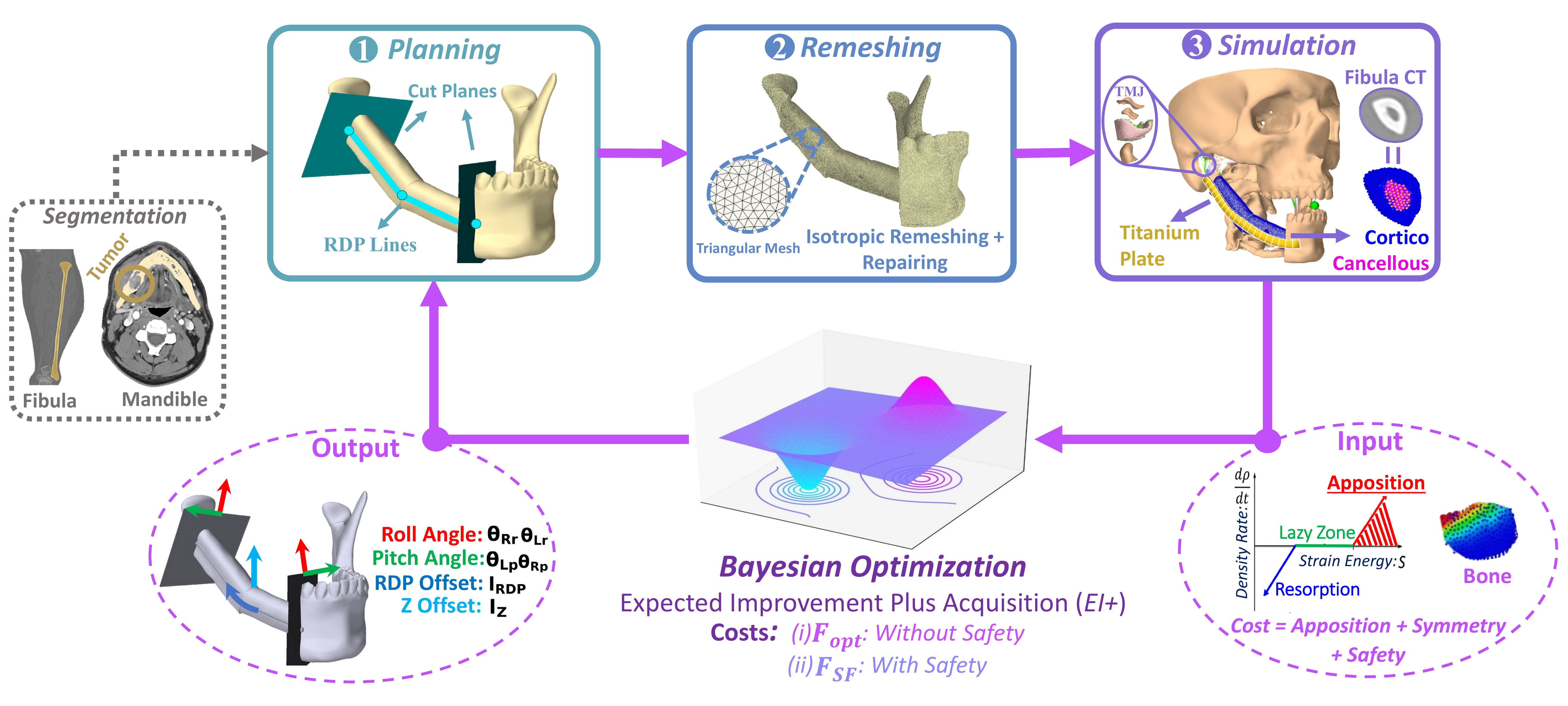}
    \caption{\textbf{Planning and optimization workflow.} The process starts from CT-based segmentation of the mandible and donor fibula, followed by construction of the corresponding 3D surface models and definition of the tumor or resection region. For each candidate design, the virtual-planning stage generates the defect-specific resection geometry, donor-segment configuration, and fixation-plate layout. The arrowheads indicate the positive axis directions of the design variables. For the cut-plane variables, positive and negative rotations follow the right-hand rule about the indicated local axes; these axes are shown schematically for visualization and are not necessarily centered on the planes, and the local cut-plane \(z\)-axis points toward the defect and is omitted from the illustration for clarity. Roll and pitch are tilts of the resection plane about its x (red) and y (green) axes; rotation about the long-axis-aligned z-axis is excluded as it leaves the plane unchanged. The resulting anatomical components are then repaired and remeshed to obtain simulation-ready geometries. In the simulation stage, the reconstructed mandible is dynamically simulated to evaluate donor-native apposition and related cost function, which are returned to the Bayesian optimizer to guide selection of the next candidate configuration.}
    \label{fig:structopt_workflow}
\end{figure*}

\section{Materials and Methods}

This section is organized into four main subsections: the optimization workflow (Section~\ref{subsec:opt_workflow}), the Bayesian optimization formulation (Section~\ref{subsec:bo_formulation}), the patient-specific modeling pipeline (Section~\ref{sec:functional_patient_specific_modeling}), and the experimental setup (Section~\ref{subsec:experimental_design}). We first describe the planning workflow used to generate and evaluate candidate mandibular reconstructions from anatomical surface geometry. We then present the Bayesian optimization strategy used to search this design space. Next, we describe the template-to-patient pipeline used to construct a patient-specific digital twin from patient CT imaging as a one-time preprocessing step, after which the same optimization loop is applied without additional registration cost. Finally, we summarize the experimental setup used to assess the workflow.  In this setting, pre-op CT was used for patient-specific planning, day-5 post-op CT was used to recover the realized surgical reference configuration, and year-1 post-op CT was used only for longitudinal analysis. The workflow builds on the validated dynamical craniofacial model of \citet{aftabi2024extent,aftabi2025osteoopt}.

Ethics approval was obtained for the human data components of this research from the University of British Columbia Clinical Research Ethics Board for the project entitled \textit{Assessing Functional and Structural Outcomes of the Mandible Post-Reconstruction Surgery} (Certificate No.~H24-01840).

\subsection{Optimization Workflow}\label{subsec:opt_workflow}

This subsection describes the computational workflow used to generate and evaluate candidate surgical configurations. The workflow comprises virtual planning, mesh refinement, and simulation, which are executed consistently within each optimization iteration and are summarized in Figure~\ref{fig:structopt_workflow}.

\subsubsection{Virtual Planning}

Virtual planning is performed on 3D anatomical models generated from CT. The mandible and donor fibula are segmented from the CT images and converted into surface models that provide the geometric basis for planning. Using these models, the surgeon defines the resection plane positions according to the defect type, while the remaining angular and length-based variables are optimized. Virtual planning then follows the reconstruction workflow introduced in our earlier work~\citep{aftabi2025optimizing}. In the present study, we focus on three defect classes selected since they are frequently encountered and collectively sample different anatomical regions of the mandible: body (B), symphysis (S), and ramus-body (RB) defects according to Urken's classification~\citep{urken1991oromandibular}. The number and position of donor segments are determined from the mandibular contour using the Ramer-Douglas-Peucker (RDP) algorithm, which approximates the image-derived contour with piecewise linear segments~\citep{shkedy2020predicting}. For a contour point \(\mathbf{p}_k\) and the line segment joining the current endpoints \(\mathbf{p}_1\) and \(\mathbf{p}_n\), the perpendicular deviation is
\begin{equation}
d_{\perp}\!\left(\mathbf{p}_k,\overline{\mathbf{p}_1\mathbf{p}_n}\right)
=
\frac{\left\|(\mathbf{p}_n-\mathbf{p}_1)\times(\mathbf{p}_1-\mathbf{p}_k)\right\|_2}
{\left\|\mathbf{p}_n-\mathbf{p}_1\right\|_2},
\label{eq:structopt_rdp_distance}
\end{equation}
and the contour is recursively split at the point with the largest deviation whenever this value exceeds a prescribed tolerance. In this way, the contour representation determines both the number of donor segments and the candidate intermediate cut locations. One donor segment is used for B and S defects, whereas RB defects require two segments; however, the formulation is not restricted to these specific cases. This contour-based segmentation is automatic and purely geometric, in contrast to anatomical landmark-guided strategies that position donor segments relative to functional landmarks such as the dental arch and implant sites~\citep{wu2025anatomical}. The two are complementary: the clinical considerations those landmarks encode enter our workflow through the feasible-region constraints on the subsequent optimization.

To increase bony contact, a brute-force search is performed on the reconstructed donor and resection surfaces. Overlapping planar polygons extracted from these surfaces are analyzed to identify cut configurations that maximize apposition. This image-derived geometric matching step was used to increase contact between the donor and the native mandible, which is favorable for bone union~\citep{aftabi2025optimizing,aftabi2025osteoopt}.

Once the donor segments are positioned, a coarse finite element (FE) fixation plate is generated along the reconstructed mandibular contour by projecting points from contour markers and surface normals to define a smooth plate path. The plate volume is then hexahedrally meshed and, as a deformable titanium FE body, conformed to the mandibular curvature through tuned spring attachments together with plate-mandible contact under a forward simulation. The plate is regenerated for every candidate configuration, with the locking screws positioned on the regenerated plate, so that each candidate reconstruction can be simulated under comparable fixation conditions.

\subsubsection{Mesh Refinement}
The anatomical components produced during virtual planning are subsequently refined to ensure mesh quality and consistency for simulation. Surface refinement is performed on the reconstruction geometries using the MeshLab API~\citep{pymeshlab,cignoni2008meshlab}, applying isotropic explicit remeshing with a target edge length of 0.50 mm over 50 refinement iterations. A mesh sensitivity analysis on the maximum principal stress (MPS) was conducted to determine the optimal donor edge length, with refinement stopped when changes in MPS fell below 5\%, balancing computational cost and accuracy. Additional processing merges nearby vertices, snaps mismatched borders, removes duplicate faces, repairs non-manifold elements, and closes small holes to preserve structural integrity.

This stage is essential since the optimization repeatedly generates geometries with different cut orientations and donor positions. Robust remeshing prevents surface artifacts introduced during repeated geometric processing from contaminating the mechanical response. The refined meshes are then passed to the simulation stage, where the objective functions are evaluated.

\subsubsection{Simulation}

The simulation stage is implemented in ArtiSynth~\citep{lloyd2012artisynth} (www.artisynth.org), an open-source physics engine that combines multibody dynamics with finite-element modeling. It builds on the previously validated craniofacial model for mandibular reconstruction presented in \citet{aftabi2024extent}, which was developed from prior jaw-modeling work~\citep{hannam2010comparison,stavness2010predicting,sagl2019dynamic}. The base model comprises rigid maxillary, mandibular, and hyoid bodies, 24 Hill-type muscles, ligaments, a scar-tissue representation, and a temporomandibular joint representation with a disc and finite-element capsule. In the patient-specific setting, the simulation is driven by digital twins constructed from patient CT imaging. Since these components have already been described, they are not repeated here. Instead, we focus on the components introduced or adapted for the present image-driven planning workflow: defect-specific reconstruction, donor and fixation-plate modeling, image-derived donor properties, and patient-specific functional updating.

\paragraph{Defect-Specific Reconstruction Model.}

To represent different resections, the muscles attached to the resected regions are removed according to the defect class defined by Urken's classification~\citep{aftabi2024extent}; the specific muscles omitted for each defect type follow our prior work and are listed in Appendix Table~\ref{tab:resected_muscles}. Consistent with the surgical technique used in our cohort, they are not re-attached to the donor bone or fixation plate. The resulting soft-tissue response is captured using a defect-specific scar-tissue representation modeled as a six-dimensional spring-damper system following prior works~\citep{aftabi2024extent,hannam2010comparison,stavness2010predicting,corr2009biomechanical}. The donor bone and fixation plate are incorporated as FE bodies derived from the reconstructed geometry. The donor is tetrahedralized from the refined donor surface mesh using TetGen~\citep{hang2015tetgen}, while the fixation plate is modeled as linear elastic titanium with density \(\rho = 4.42\) g/cm\textsuperscript{3}, Young's modulus \(E = 100\) GPa, and Poisson ratio \(\nu = 0.3\). The plate is anchored to the donor bone using rigid locking screws. The plate was attached to the resected mandible by directly linking its nodes to the mandible's surface~\citep{aftabi2025optimizing,aftabi2025osteoopt}. The effect of this simplification on the safety-factor objective is examined in Section~\ref{design_variables}.

The donor bone is represented as a cortico-cancellous structure derived from CT attenuation. Regions with \(\mathrm{HU} > 1000\) are operationally classified as cortical bone~\citep{mahesh2013essential}, whereas the remaining regions are classified as cancellous bone. The average Hounsfield unit (HU) value is then computed separately for the two image-defined regions, and the apparent density assigned to each region is obtained by linear mapping according to
\begin{equation}
\rho = \rho_{\min} + (\rho_{\max} - \rho_{\min})\
\frac{\mathrm{HU} - \mathrm{HU}_{\min}}{\mathrm{HU}_{\max} - \mathrm{HU}_{\min}},
\label{eq:structopt_density}
\end{equation}
where \(\rho_{\min} = 0.7\) g/cm\textsuperscript{3}, \(\rho_{\max} = 1.8\) g/cm\textsuperscript{3}, \(\mathrm{HU}_{\min}=350\), and \(\mathrm{HU}_{\max}=1700\)~\citep{field2010prediction}. For measured mean HU values outside this calibration interval, the assigned density was bounded by \(\rho_{\min}\) and \(\rho_{\max}\). The cortical and cancellous regions are then assigned material properties \((E=13.7\ \mathrm{GPa}, \nu=0.3)\) and \((E=1.1\ \mathrm{GPa}, \nu=0.3)\), respectively~\citep{field2010prediction}. Within the optimization workflow, this region-wise density assignment preserves computational tractability across repeated optimization calls.

\paragraph{Contact and Bone Union Metric.}

Contact between the donor and the native mandible is modeled using vertex penetration on the reconstructed surfaces, a standard penalty-based elastic foundation contact device in which the small surface overlap is mapped to a restoring pressure rather than representing literal bone interpenetration~\citep{bei2004multibody,lloyd2025modelguide}. The AJL contour collider implemented in ArtiSynth is used to identify local penetration regions and estimate the associated contact area from the surface geometry. This contact response is then regularized with the elastic foundation contact formulation adopted from \citet{aftabi2024extent,bei2004multibody}. The contact pressure is written as
\begin{equation}
p_{\mathrm{contact}}=
- \frac{(1-\nu)E}{(1+\nu)(1-2\nu)}
\ln{\left[1-\frac{d}{t_{\mathrm{contact}}}\right]},
\label{eq:structopt_contact_pressure}
\end{equation}
where \(E = 30\) kPa, \(\nu = 0.3\), \(t_{\mathrm{contact}} = 0.2\) mm is the thickness of the contact layer, and \(d\) is the penetration depth~\citep{bei2004multibody,ferguson2022,aftabi2024extent,lloyd2025modelguide}. These parameters approximate the soft tissue region and residual gap that are often present in the early postoperative period, and because the contact pressure diverges as the penetration approaches \(t_{\mathrm{contact}}\), the admissible overlap remains bounded within this thin compliant layer. The 30~kPa layer is a fixed early-stage surrogate: the workflow uses the early-phase apposition pattern as a spatial correlate of the year-1 bone formation rather than reproducing the fully healed interface; as it stiffens, the objective changes monotonically for every candidate, providing a consistent comparative model across candidates, although configuration-dependent ranking effects cannot be excluded.

Bone-union propensity is quantified using a density-normalized mechanical stimulus derived from the elemental strain energy density (SED), following established bone-remodeling formulations motivated by Wolff's law~\citep{zheng2019investigation,zheng2022,wan2022interaction,field2010prediction,ferguson2022,wu2021machine}. Specifically,
\begin{equation}
\mathrm{SED} = \frac{1}{2}\sum_{i=1}^{3}\sum_{j=1}^{3}\sigma_{ij}\varepsilon_{ij},
\qquad
S = \frac{\mathrm{SED}}{\rho},
\label{eq:structopt_sed}
\end{equation}
where \(\sigma_{ij}\) and \(\varepsilon_{ij}\) denote the elemental stress and strain tensor components. Bone apposition, used here as a surrogate for bone remodeling and union propensity, is assumed to occur when the contact-transmitted chewing load drives the stimulus \(S\) above the remodeling threshold \(S_0(1+\delta)\), with \(S_0=0.036\ \mathrm{mJ/g}\) and \(\delta=0.1\)~\citep{field2010prediction,zheng2022,ferguson2022,lin2010bone,rungsiyakull2011loading}. The corresponding lazy-zone width is therefore \(2\delta=0.2\). To assess remodeling at the donor-mandible interface, a single layer of donor elements is extracted at each resection side, and the apposition fraction is defined as the ratio of elements exceeding the threshold to the total number of elements in that layer. This thin-layer choice reflects the interface-localized character of bone apposition observed clinically, with longitudinal CT studies of fibula-flap mandibular reconstruction showing that bone mineral density and cortical bridging concentrate within a narrow region adjacent to the resection interface during the early healing phase~\citep{zheng2022}.

This choice of objective signal is central to the optimization strategy. Unlike purely geometric criteria, the apposition fraction couples image-derived interface geometry to a mechanically meaningful remodeling stimulus. It is used as a comparative planning surrogate: for a fixed patient and feasible surgical design space, the optimizer ranks candidate configurations by SED/apposition-derived predicted bone-union propensity rather than predicting the full biological healing process.

Forward-dynamics chewing simulations were then performed in ArtiSynth over one chewing cycle of approximately 0.62 s, including the food-bolus engagement interval, at a temporal resolution of 0.001 s. In the forward simulation, chewing is unilateral on the non-reconstructed side, the side patients predominantly use early after surgery~\citep{hannam2010comparison,roumanas2006masticatory}; muscle activation is set to pre-reconstruction levels to represent this stage~\citep{aftabi2024extent}, identically for the generic and patient-specific models. For the patient-specific cases, the optimization was additionally repeated under an adapted activation profile from inverse simulation, in which the muscles remaining after resection, with the scar tissue, follow the pre-reconstruction jaw trajectory, approximating the neuroplastic adaptation that follows reconstruction; this tests whether the recovered configuration is stable under a loading that differs from the one used for optimization. The forward and inverse formulations and equations are detailed in Appendix~\ref{app:forward_inverse}, where the normal forward and adapted inverse activation profiles are also shown (Fig.~\ref{fig:fwd_inv_b}).

\subsection{Bayesian Optimization Formulation}\label{subsec:bo_formulation}

Having defined the simulation that scores each candidate reconstruction, the surgical design problem is posed as a Bayesian optimization over the variables that govern donor placement and resection orientation. Each objective evaluation requires geometry generation from anatomical surfaces, remeshing, and dynamic simulation, making the objective expensive, nonlinear, and mildly noisy. Bayesian optimization is therefore used to learn a probabilistic surrogate of the objective and to select informative candidates with a limited number of evaluations~\citep{shahriari2015taking}.

\subsubsection{Design Variables and Objective Functions}
\label{design_variables}

The angular variables correspond to the roll and pitch angles of the left and right image-defined resection planes, while the length variables govern donor positioning:
\begin{equation}
\boldsymbol{\theta} =
(\theta_{Lr}, \theta_{Lp}, \theta_{Rr}, \theta_{Rp})^T,\qquad
\boldsymbol{\ell} = (l_Z, l_{RDP})^T,
\label{eq:structopt_theta_l}
\end{equation}
and the complete design vector is
\begin{equation}
\boldsymbol{\phi} = (\boldsymbol{\theta}^T,\boldsymbol{\ell}^T)^T \in \mathbb{R}^{6}.
\label{eq:structopt_phi}
\end{equation}
In the angular subscripts, \(L/R\) denotes the left or right resection plane, and \(r/p\) denotes roll or pitch. As illustrated in Figure~\ref{fig:structopt_workflow}, the arrowheads indicate the positive axis directions of these variables. Positive and negative angular changes are defined by the right-hand rule about the corresponding local cut-plane axes, while the local \(z\)-axis of each cut plane points toward the defect and is omitted from the drawing for clarity. Here roll and pitch tilt the resection plane about its x and y axes; the local z-axis is aligned with the bone's long axis and is not optimized, since rotating a plane about its normal leaves the osteotomy unchanged. The variable \(l_Z\) is the vertical donor offset and \(l_{RDP}\) is the RDP point offset controlling the location of the intermediate cut in the contour-derived two-segment RB reconstruction. The feasible region is written as
\begin{equation}
\mathcal{X}=
\left\{
\boldsymbol{\phi}\in\mathbb{R}^{6}\ \middle|
\begin{array}{l}
\theta_{Lr}\in[-\alpha_r,\alpha_r],\quad
\theta_{Lp}\in[-\alpha_p,\alpha_p],\\
\theta_{Rr}\in[-\beta_r,\beta_r],\quad
\theta_{Rp}\in[-\beta_p,\beta_p],\\
l_Z\in[-z,z],\\
l_{RDP}\in[-r,r]
\end{array}
\right\}.
\label{eq:structopt_feasible}
\end{equation}
The ranges were chosen based on physical, surgical, and anatomical constraints, including preservation of the clinically selected donor orientation for vascular considerations, avoidance of unnecessary muscle sacrifice, minimization of tooth-root compromise, maintenance of a 10--25 mm vertical gap between the fibula and maxillary teeth for implant placement~\citep{tran2023dental}, and a 20 mm minimum fibular segment length in two-segment reconstructions based on reported revascularization considerations~\citep{knitschke2022osseous}. The defect-specific bounds used in the experiments are described in Section~\ref{subsec:experimental_design}; the intermediate-cut offset \(l_{RDP}\) enters only for the two-segment RB reconstruction and is omitted for the single-segment B and S defects. These constraints encode, as feasibility bounds on the optimization, the functional considerations that anatomical landmark-guided strategies instead address through explicit landmark targets~\citep{wu2025anatomical}; incorporating such landmark-based functional objectives is left for future work.

Let \(L_i(\boldsymbol{\phi})\), \(R_i(\boldsymbol{\phi})\), and \(M_i(\boldsymbol{\phi})\) denote the apposition fractions at time step \(i\) on the left, right, and middle resection planes, respectively. Their chewing-cycle average is written in the unified form
\begin{equation}
\overline{X}(\boldsymbol{\phi}) =
\frac{1}{n}\sum_{i=1}^{n} X_i(\boldsymbol{\phi}),
\qquad X\in\mathcal{S}=\{L,R,M\}.
\label{eq:structopt_cycle_average}
\end{equation}
where \(n\) denotes the number of samples spanning one chewing cycle of approximately \(0.62\) seconds. Bone-remodeling formulations are time-integrated rather than peak-only: in Prasad's model the formation rate ($B$) scales jointly with strain magnitude above threshold and the number of loading cycles ($B\propto(\varepsilon-\varepsilon_{\mathrm{thres}})N^{q}$)~\citep{prasad2019invertible,rubin1984regulation}. Cycle-averaged apposition is therefore the natural geometric analog of the established remodeling stimulus, capturing both how much donor-host contact is present and for how long, whereas the instantaneous peak captures only the former. This argument applies equally to the strain-energy-density formulation adopted here, which is one of the accepted scalar measures of the same loading field~\citep{prasad2019invertible,field2010prediction}. Consistent with this rationale, the longitudinal analysis in Section~\ref{subsec:results_validation} shows spatial overlap between the cycle-averaged apposition pattern and the year-1 bone-formation distribution observed on CT, supporting its use as a comparative optimization surrogate. Based on these cycle-averaged interface measures extracted from the reconstructed geometry, the optimization is posed using the two objective functions introduced below.

\begin{figure*}[t]
    \centering
    \includegraphics[width=1\textwidth]{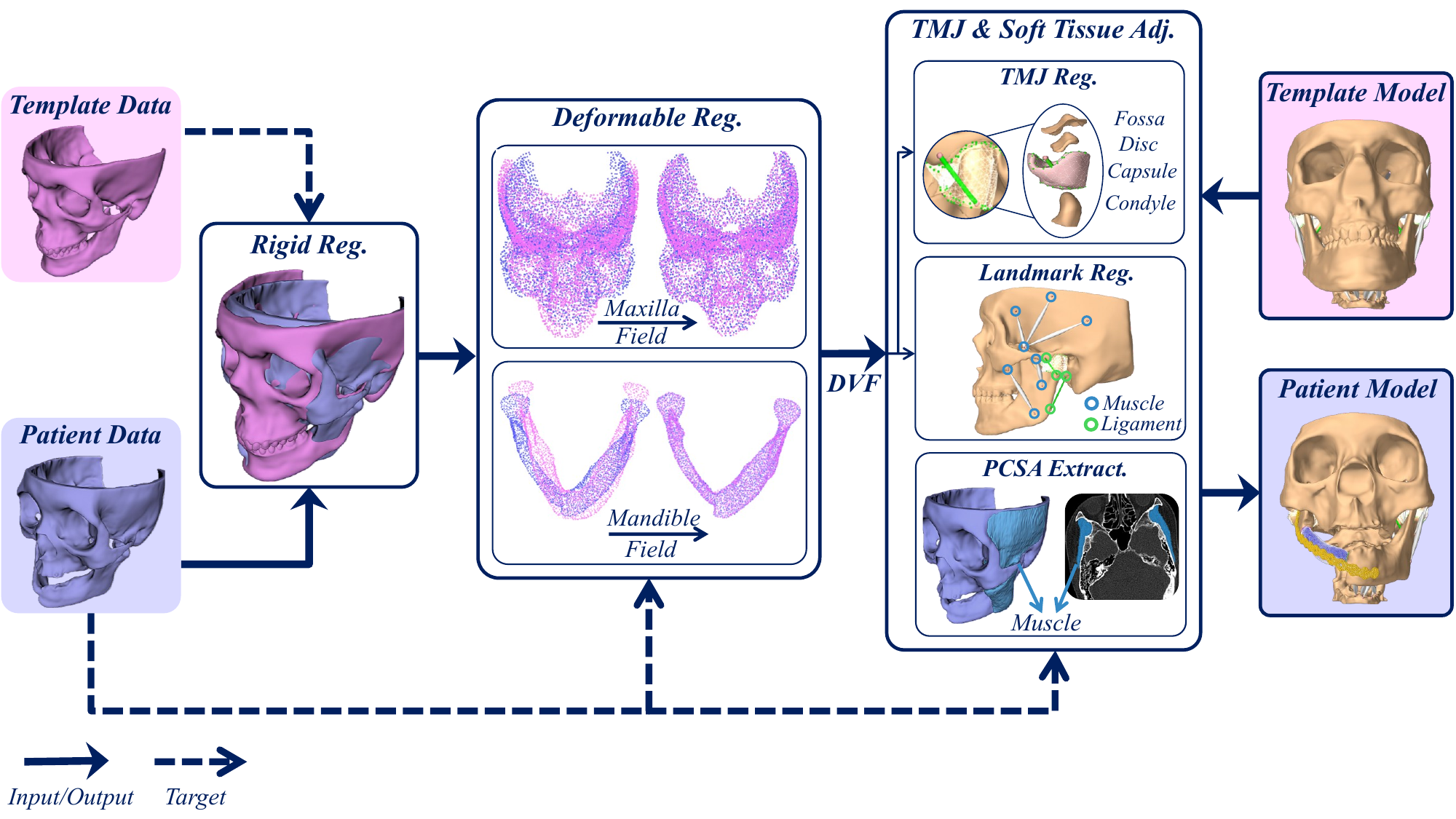}
    \caption{\textbf{Overview of the automated patient-specific workflow.} The diagram summarizes the template-to-patient registration pipeline described in Algorithm~\ref{alg:psm_pipeline}. A healthy generic craniofacial model provides the anatomical prior, while patient CT provides the patient-specific skeletal geometry and muscle information used for personalization. These two streams are coupled through maxilla-based rigid initialization, coherent point drift registration, transfer of musculoligamentous attachment sites, updating of muscle and ligament model parameters including PCSA-scaled muscle forces, and anatomy-guided registration of the finite-element TMJ disc and capsule, yielding a patient-specific digital twin for subsequent functional analysis and structural optimization.}
    \label{fig:psm_overview}
\end{figure*}

\paragraph{Objective 1: \(F_{\mathrm{opt}}\).}
For the generalized multi-interface case, the primary objective is
\begin{equation}
F_{\mathrm{opt}}(\boldsymbol{\phi}) =
W_1\!\sum_{X\in\mathcal{S}}\!\overline{X}(\boldsymbol{\phi})
- W_2\!\sum_{\substack{X,Y\in\mathcal{S}\\X<Y}}\!
|\overline{X}(\boldsymbol{\phi})-\overline{Y}(\boldsymbol{\phi})|.
\label{eq:structopt_fopt}
\end{equation}
Here, \(W_1=W_2=0.5\). The first term favors large average apposition across the resection interfaces, whereas the second penalizes imbalance between them. In this way, the objective does not merely reward high contact at one interface, but instead promotes a more uniformly favorable mechanical environment for union. The optimization is posed as a minimization of \(-F_{\mathrm{opt}}\).
For single-segment reconstructions, the middle interface is absent, and \(F_{\mathrm{opt}}\) reduces directly to the corresponding left-right two-interface form.

\paragraph{Objective 2: \(F_{\mathrm{SF}}\).}
To discourage reconstructions that improve apposition at the expense of unfavorable local bone loading, the optimization also considers
\begin{equation}
F_{\mathrm{SF}}(\boldsymbol{\phi}) = F_{\mathrm{opt}}(\boldsymbol{\phi}) - \overline{C}(\boldsymbol{\phi}),
\label{eq:structopt_fsf}
\end{equation}
where the penalty term is
\begin{equation}
\overline{C}(\boldsymbol{\phi}) =
\frac{1}{n}\sum_{i=1}^{n}
\sum_{s\in\{\mathrm{left},\mathrm{right}\}}
w^s
\bigl[\max(0, SF_{\mathrm{desired}}-SF^s_{\mathrm{worst},i})\bigr]^2 
\label{eq:structopt_penalty}
\end{equation}
with \(SF_{\mathrm{desired}}=1\) and \(w^s=0.5\). Here, \(SF\) denotes the safety factor, defined as the ratio between bone yield stress and the corresponding maximum principal stress; values below \(1\) therefore indicate that the estimated local stress exceeds the nominal bone limit. The penalty becomes active only when the safety factor on a given side falls below the desired threshold. The worst-case safety factor on side \(s\) is defined by
\begin{equation}
SF^s_{\mathrm{worst},i} =
\min
\left\{
\frac{\sigma_{\mathrm{yield,cortical}}}{\sigma^s_{\mathrm{maxP,cortical},i}},
\frac{\sigma_{\mathrm{yield,cancellous}}}{\sigma^s_{\mathrm{maxP,cancellous},i}}
\right\},
\label{eq:structopt_sf}
\end{equation}
where \(\sigma_{\mathrm{yield,cortical}}=100\ \mathrm{MPa}\) and \(\sigma_{\mathrm{yield,cancellous}}=5\ \mathrm{MPa}\) are the cortical and cancellous yield stresses, and \(\sigma^s_{\mathrm{maxP},i}\) is the MPS in each region on side \(s\) at time \(i\) during the chewing cycle. This definition is intentionally conservative: at each time step and on each interface side, the lower of the cortical and cancellous safety factors governs the penalty, as local risk is determined by the more vulnerable bone region rather than by the average response. Considering this worst-case scenario discourages solutions that achieve favorable apposition but produce locally unfavorable loading patterns that could lead to bone resorption or fracture. This penalty depends on the maximum principal stress in the donor cortical and cancellous bone at the resection interface, not on plate or screw stress, so the rigid idealization of the plate attachments enters only through the load path; applied identically to every candidate, this simplification is expected to affect the absolute $F_{\mathrm{SF}}$ values more than the relative ranking used for optimization. Consequently, \(F_{\mathrm{SF}}\) retains the apposition-driven character of \(F_{\mathrm{opt}}\) while favoring safer interface conditions.

\subsubsection{Gaussian-Process Surrogate and Acquisition Function}

Let \(f(\boldsymbol{\phi})\) denote the expensive objective to be minimized over \(\mathcal{X}\), with \(f\in\{-F_{\mathrm{opt}},-F_{\mathrm{SF}}\}\). Since each evaluation is generated by the full geometry-generation and simulation pipeline, the observations are treated as noisy: \(y_j=f(\boldsymbol{\phi}_j)+\varepsilon_j\), with \(\varepsilon_j\sim\mathcal{N}(0,\sigma^2)\). The accumulated observations are used to update a Gaussian-process (GP) surrogate with prior mean and covariance \((\mu,k)\). The resulting posterior mean and predictive variance, \((\mu_N,\sigma_Q^2)\), provide the estimated objective value and uncertainty at each untested design. An automatic relevance determination (ARD) Matérn 5/2 kernel is used so that separate length scales encode anisotropic sensitivity across the surgical design variables. The full GP posterior-update and kernel equations are provided in Appendix~\ref{app:gp_surrogate}.

The search is initialized with \(n_{\text{samples}}=25\) Sobol' points, generated with a skip of 1000 and a leap of 100~\citep{renardy2021sobol}, and linearly mapped to the feasible region \(\mathcal{X}\). This initialization establishes quasi-uniform coverage of \(\mathcal{X}\), reducing bias before Bayesian optimization begins. For either objective \(f\in\{-F_{\mathrm{opt}},-F_{\mathrm{SF}}\}\), the dataset at iteration \(N\) can be written as \(\mathcal{D}_N=
\{(\boldsymbol{\phi}_j,y_j)\}_{j=1}^{n_{\text{samples}}}
\cup
\{(\boldsymbol{\phi}_j,y_j)\}_{j=n_{\text{samples}}+1}^{N},\) where the first term corresponds to the Sobol' initialization and the subsequent evaluations are incorporated iteratively to refine the search. At each iteration, the next point is chosen using an expected improvement plus (EI+) acquisition rule. EI+ retains the expected-improvement form while adding a safeguard against overexploitation. Let \(\tilde{f}_{\min}=\min\{y_j:(\boldsymbol{\phi}_j,y_j)\in\mathcal{D}_N\}\), define \(S(\boldsymbol{\phi})=\sigma_Q^2(\boldsymbol{\phi})\), and write \(z(\boldsymbol{\phi})=(\tilde{f}_{\min}-\mu_N(\boldsymbol{\phi}))/\sqrt{S(\boldsymbol{\phi})}\). The acquisition score for minimization is
\begin{equation}
\operatorname{EI+}(\boldsymbol{\phi}) =
\Bigl(\tilde{f}_{\min} - \mu_N(\boldsymbol{\phi})\Bigr)
\Phi\!\left(z(\boldsymbol{\phi})\right)
+
\sqrt{S(\boldsymbol{\phi})}\,
\varphi\!\left(z(\boldsymbol{\phi})\right),
\label{eq:structopt_ei}
\end{equation}
where \(\Phi(\cdot)\) and \(\varphi(\cdot)\) denote the cumulative distribution function and probability density function of the standard normal distribution, respectively~\citep{qin2017improving}. The next sampling point is selected by maximizing \(\operatorname{EI+}(\boldsymbol{\phi})\) over \(\mathcal{X}\).

The ``+'' in EI+ refers to this additional safeguard. Let \(t_{\sigma}>0\) denote the exploration ratio. If the point selected by expected improvement satisfies \(\sigma_F(\boldsymbol{\phi}_{N+1}) < t_{\sigma}\sigma\), its posterior uncertainty is considered too small relative to the observation-noise level. In that case, the kernel hyperparameters are multiplicatively inflated before accepting the next point, increasing uncertainty away from sampled regions and encouraging further exploration. The exploration-ratio hyperparameter was tuned on the generic defect cases and then fixed before all patient-specific evaluations. This tuning yielded \(t_{\sigma}=0.5\) for the one-segment B and S defects and \(t_{\sigma}=0.6\) for the two-segment RB defect. The higher value for RB is consistent with the broader search space introduced by the additional \(l_{RDP}\) variable and with the less sharply localized objective landscape observed during generic-case tuning.

\subsection{Patient-Specific Modeling}\label{sec:functional_patient_specific_modeling}

This subsection describes the automated procedure used to construct a patient-specific digital twin from patient CT imaging. In the retrospective studies, the same pipeline was applied to pre-op CT for patient-specific planning. Executed once as a preprocessing step, the pipeline produces $\mathcal{M}_{\mathrm{ps}}$, which is then passed directly to the optimization loop described in Sections~\ref{subsec:opt_workflow}--\ref{subsec:bo_formulation} without additional registration cost at each iteration.

Patient-specific mandibular and maxillary surfaces obtained from the corresponding CT volume were first conditioned to obtain smooth target geometries suitable for correspondence estimation. The generic model, corresponding to the same model used in Section~\ref{subsec:opt_workflow}, was then adapted to the planned or realized reconstruction pattern according to the Urken defect classification by excluding muscles and ligamentous structures attached to anatomy removed by the defect. The corresponding defect-specific scar-tissue representation inherited from the generic model was retained so that the personalized model remained consistent with the reconstructed defect type.

The patient-specific digital twin was then obtained by using the patient CT meshes as the simulated skeletal geometry and registering the generic template onto them to transfer the functional scaffold that segmentation cannot define, namely skeletal attachment sites, musculoligamentous parameters, and the temporomandibular joint disc and capsule. The deformed template itself is not simulated; it serves only to carry these functional components onto the patient anatomy, keeping the resulting model compatible with the optimization workflow. Figure~\ref{fig:psm_overview} summarizes the workflow, and Algorithm~\ref{alg:psm_pipeline} lists the main computational steps.

\subsubsection{Rigid and Deformable Template-to-Patient Registration}

Registration was performed separately for the mandible and the maxilla/skull segment. The generic model was first rigidly initialized from the intact maxilla/skull using centroid alignment followed by iterative closest point refinement~\citep{besl1992method}. This maxilla-based transform, denoted by \((\mathbf{R}_0,\mathbf{t}_0)\), was propagated to the remaining template geometries to provide a stable starting point while avoiding bias from the surgically altered mandible.

Nonrigid template-to-patient correspondence was then estimated with coherent point drift (CPD)~\citep{myronenko2010cpd}. CPD treats the template control points as centroids of a Gaussian mixture model and estimates a smooth displacement field by expectation-maximization (EM)~\citep{myronenko2010cpd,dempster1977maximum}. Separate mandibular and maxillary fields, \(\mathcal{T}_{\mathrm{mand}}\) and \(\mathcal{T}_{\mathrm{max}}\), were retained because the reconstructed mandible and cranial base provide distinct anatomical constraints and support different attachment-site classes. The rigid-initialization and CPD deformation equations are provided in Appendix~\ref{app:registration_equations}.

\begin{algorithm}[t]
\scriptsize
\caption{Template-to-Patient Registration Pipeline for Patient-Specific Craniofacial Modeling}
\label{alg:psm_pipeline}
\begin{algorithmic}[1]
\Require generic model $\mathcal{M}_0$; generic skeletal surfaces $\mathcal{S}^{0}_{\mathrm{mand}},\mathcal{S}^{0}_{\mathrm{max}}$; patient CT volume $I_{\mathrm{CT}}$; patient skeletal surfaces $\mathcal{S}_{\mathrm{mand}},\mathcal{S}_{\mathrm{max}}$; generic and patient condylar/fossa surfaces $\mathcal{S}^{0}_{\mathrm{cond}},\mathcal{S}_{\mathrm{cond}},\mathcal{S}^{0}_{\mathrm{fossa}},\mathcal{S}_{\mathrm{fossa}}$; defect descriptor $\mathcal{D}$; attachment set $\mathcal{A}_0$; generic TMJ soft tissues $\mathcal{X}_{\mathrm{disc}},\mathcal{X}_{\mathrm{caps}}$
\State $\mathcal{M}_0\gets \textsc{Prune}(\mathcal{M}_0,\mathcal{D})$ \Comment{Urken-based defect adaptation}
\State $(\mathbf{R}_0,\mathbf{t}_0)\gets \textsc{RigidInit}(\mathcal{S}^{0}_{\mathrm{max}},\mathcal{S}_{\mathrm{max}})$ \Comment{Intact maxilla-based alignment}
\State $(\widetilde{\mathcal{S}}^{0}_{\mathrm{mand}},\widetilde{\mathcal{S}}^{0}_{\mathrm{max}})\gets (\mathbf{R}_0\mathcal{S}^{0}_{\mathrm{mand}}+\mathbf{t}_0,\mathbf{R}_0\mathcal{S}^{0}_{\mathrm{max}}+\mathbf{t}_0)$ \Comment{Propagate rigid transform}
\State $\mathcal{T}_{\mathrm{mand}}\gets \textsc{CPD}(\widetilde{\mathcal{S}}^{0}_{\mathrm{mand}},\mathcal{S}_{\mathrm{mand}})$ \Comment{Mandibular field}
\State $\mathcal{T}_{\mathrm{max}}\gets \textsc{CPD}(\widetilde{\mathcal{S}}^{0}_{\mathrm{max}},\mathcal{S}_{\mathrm{max}})$ \Comment{Maxillary field}
\For{each point $\mathbf{p}\in\mathcal{A}_0$ with anatomical parent $k$} \Comment{Transfer landmarks}
  \State $\mathbf{p}'\gets \mathbf{p}+\Delta_k(\mathbf{p})$ \Comment{Deformation transfer}
  \State $\widehat{\mathbf{p}}\gets \Pi_{\mathcal{S}_k}(\mathbf{p}')$ \Comment{Surface refinement}
\EndFor
\State $\mathcal{M}_{\mathrm{mus}}\gets \textsc{TotalSegmentator}(I_{\mathrm{CT}})$ \Comment{CT-based muscle segmentation}
\State $\mathrm{SCS}_m\gets \textsc{ExtractSections}(\mathcal{M}_{\mathrm{mus}})$ \Comment{Muscle cross section}
\State $\widehat{\mathrm{PCSA}}_{m}^{(r)}\gets a_{m}^{(r)}\mathrm{SCS}_{m}+b_{m}^{(r)},\quad r\in\{\mathrm{WPCS},\mathrm{BPCS}\}$ \Comment{PCSA regression \(r\)}
\State $\widehat{\mathrm{PCSA}}_m\gets \tfrac{1}{2}\!\left(\widehat{\mathrm{PCSA}}_{m}^{(\mathrm{WPCS})}+\widehat{\mathrm{PCSA}}_{m}^{(\mathrm{BPCS})}\right)$ \Comment{PCSA mean estimate}
\State $F'_{\max,m}\gets 40\,[\mathrm{N/cm}^2]\times\widehat{\mathrm{PCSA}}_m$ \Comment{Update muscle force}
\State $\ell'_{\mathrm{opt},m}\gets \ell'_{\mathrm{cur},m}, \quad \ell'_{\max,m}\gets r_m\,\ell'_{\mathrm{cur},m}$ \Comment{Update muscle lengths}
\State $l'_{0,g}\gets l'_{\mathrm{cur},g}+\delta_g$ \Comment{Update ligament rest length}
\State $\Delta_{\mathrm{cond}}\gets \textsc{CPD}(\mathcal{S}^{0}_{\mathrm{cond}},\mathcal{S}_{\mathrm{cond}})$ \Comment{Condyle-guided TMJ field}
\State $\Delta_{\mathrm{fossa}}\gets \textsc{CPD}(\mathcal{S}^{0}_{\mathrm{fossa}},\mathcal{S}_{\mathrm{fossa}})$ \Comment{Fossa-guided TMJ field}
\For{each $\mathbf{x}\in\mathcal{X}_{\mathrm{disc}}\cup\mathcal{X}_{\mathrm{caps}}$}
  \State $\widehat{\mathbf{x}}\gets \mathbf{x}+w_{\mathrm{cond}}(\mathbf{x})\Delta_{\mathrm{cond}}(\mathbf{x})+w_{\mathrm{fossa}}(\mathbf{x})\Delta_{\mathrm{fossa}}(\mathbf{x})$ \Comment{Blend disc/capsule deformation}
\EndFor
\State $\widehat{\mathcal{X}}_{\mathrm{TMJ}}\gets\{\widehat{\mathbf{x}}\}$ \Comment{Assemble patient-specific TMJ}
\State $\mathcal{M}_{\mathrm{ps}}\gets (\mathcal{S}_{\mathrm{mand}},\mathcal{S}_{\mathrm{max}},\widehat{\mathcal{X}}_{\mathrm{TMJ}},\widehat{\Theta})$ \Comment{Model with functional parameters $\widehat{\Theta}$}
\end{algorithmic}
\end{algorithm}

\subsubsection{Functional Transfer of Attachment Sites}

Once the skeletal correspondence fields had been established, they were reused to perform functional transfer of anatomical attachment landmarks from the generic model to the patient-specific model, corresponding to the central stages of Figure~\ref{fig:psm_overview}. This step was used to update muscle origins and insertions, condylar reference points, and ligament attachment sites so that the personalized model preserved subject-specific functional anatomy. The bony attachment sites of the principal masticatory muscles are anatomically conserved across individuals, which justifies transferring these landmarks from the generic template to patient anatomy through registration rather than estimating them de novo for each subject~\citep{akita2019masticatory,mezey2022masseter,yu2021morphology}. For a template landmark $\mathbf{p}$ attached to anatomical support $k$, a normalized deformation transfer was used:
\begin{equation}
\Delta_k(\mathbf{p})=
\frac{
\sum_{m=1}^{M_k}
G_{\beta_k}\!\left(\mathbf{p},\mathbf{y}^{(k)}_m\right)
\Delta_k\!\left(\mathbf{y}^{(k)}_m\right)}
{\sum_{m=1}^{M_k}
G_{\beta_k}\!\left(\mathbf{p},\mathbf{y}^{(k)}_m\right)},
\quad
\mathbf{p}' = \mathbf{p} + \Delta_k(\mathbf{p}),
\label{eq:psm_landmark_transfer}
\end{equation}
with $\Delta_k(\mathbf{y}^{(k)}_m)=\mathcal{T}_k(\mathbf{y}^{(k)}_m)-\mathbf{y}^{(k)}_m$, and $G_{\beta_k}$ denoting the Gaussian kernel. In contrast to assigning each attachment to a single nearest correspondence, this normalized transfer preserves the local character of the learned deformation and provides a smoother update for functional muscle and ligament attachment landmarks.

Since deformation transfer alone may place some landmarks slightly off the patient surface, the transformed points were subsequently refined by projection to the closest anatomically valid surface vertex on the corresponding patient-specific rigid body,
\begin{equation}
\widehat{\mathbf{p}}
=
\Pi_{\mathcal{S}_k}(\mathbf{p}')
=
\arg\min_{\mathbf{q}\in\mathcal{S}_k}
\|\mathbf{q}-\mathbf{p}'\|_2,
\label{eq:psm_projection}
\end{equation}
where $\mathcal{S}_k$ denotes the patient-specific surface associated with the anatomical parent of the attachment. This surface-constrained refinement step was important for maintaining anatomically plausible functional muscle origins, insertions, and ligament attachments after nonrigid transfer to the image-derived anatomy. Finally, when the hyoid was outside the available CT field of view, it and its associated landmarks were updated using a mandible-derived similarity transform to preserve geometric consistency in the submandibular region.

\begin{figure*}[t!]
	\centering
	{\includegraphics[width=1\linewidth]{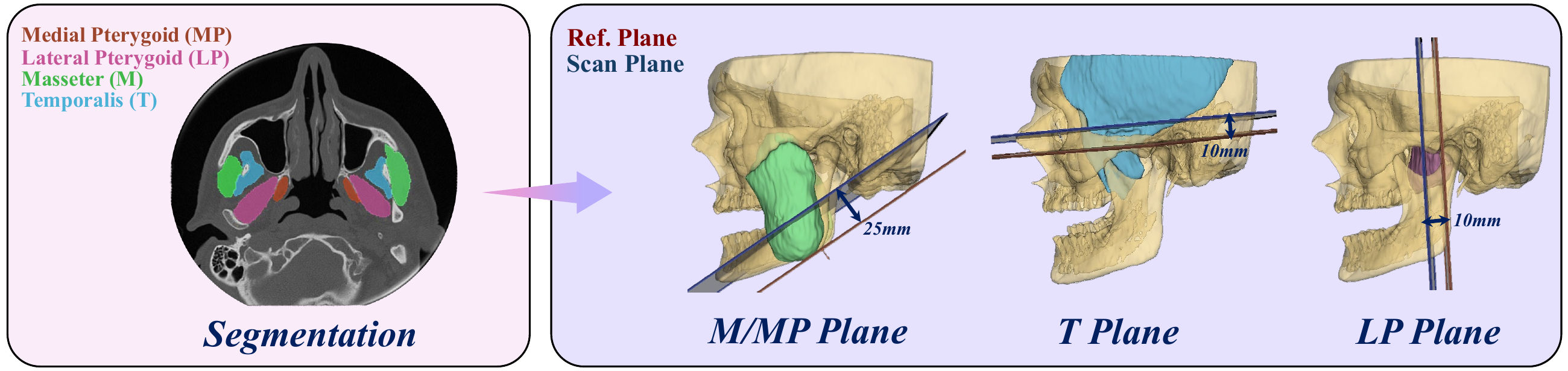}}
		\caption{\textbf{Muscle segmentation and scan cross-section plane definition for PCSA estimation.} TotalSegmentator~\citep{wasserthal2023totalsegmentator} is used to segment the principal masticatory muscles from patient CT. Landmark-defined reference planes and neighboring parallel planes are then used to extract scan cross sections (SCSs) for the PCSA update.}
	\label{Muscle_Ref}
\end{figure*}

\subsubsection{Patient-Specific Muscle and Ligament Updating}

Each masticatory muscle was represented by a Hill-type model~\citep{hill1953mechanics} in which the produced force takes the form $F = F_{\max}\,A(\ell;\ell_{\mathrm{opt}},\ell_{\max})$, with $A(\cdot)$ a length-activation function, $F_{\max}$ the maximum force, and $\ell_{\mathrm{opt}}$ and $\ell_{\max}$ the optimal and maximum lengths, respectively. These three patient-specific parameters were updated directly from the CT image~\citep{zheng2019investigation,aftabi2024extent}.

The lengths were derived from the post-registration muscle path. Let $r_m$ denote the maximum-to-optimal length ratio inherited from the generic model for muscle $m$, listed in Appendix~\ref{app:functional_modeling_parameters}, Table~\ref{tab:func_supp_9}. If $\ell^{\prime}_{\mathrm{cur},m}$ is the post-registration muscle length, the updated parameters were defined as
\begin{equation}
\ell^{\prime}_{\mathrm{opt},m} = \ell^{\prime}_{\mathrm{cur},m},
\quad
\ell^{\prime}_{\max,m} = r_m\,\ell^{\prime}_{\mathrm{cur},m},
\label{eq:psm_muscle_lengths}
\end{equation}
which preserves the original length ratio while aligning the patient-specific muscle with the registered anatomy. The maximum force $F_{\max}$ was obtained from CT image-derived physiological cross-sectional area (PCSA), as summarized in Figure~\ref{Muscle_Ref}. Automated segmentation of the patient CT was performed using TotalSegmentator~\citep{wasserthal2023totalsegmentator}, yielding subject-specific segmentations of the four main masticatory muscles: masseter, temporalis, medial pterygoid, and lateral pterygoid. For each muscle, the scan cross section (SCS) was measured from a landmark-defined reference plane and from neighboring parallel planes; the full SCS extraction workflow is described in Appendix~\ref{app:pcsa_scs_workflow}. The selected CT-derived SCS was then converted to PCSA using two regression formulations reported for these four muscle groups~\citep{weijs1984relationship}: the Weber method, denoted here by WPCS, and the Buchner method, denoted by BPCS:
\begin{equation}
\widehat{\mathrm{PCSA}}_{m}^{(r)} = a_{m}^{(r)}\,\mathrm{SCS}_{m} + b_{m}^{(r)},
\qquad r\in\{\mathrm{WPCS},\mathrm{BPCS}\},
\label{eq:psm_pcsa_regression}
\end{equation}
where the muscle-specific slopes and intercepts for the Weber (WPCS) and Buchner (BPCS) regressions are listed in Appendix~\ref{app:functional_modeling_parameters}, Table~\ref{tab:func_supp_11}. Since direct PCSA estimation requires detailed muscle fiber orientation information, which cannot be obtained from routine CT and would require diffusion tensor imaging (DTI) that was not available, both regression-based estimates were computed and the nominal patient-specific PCSA was defined as
\begin{equation}
\widehat{\mathrm{PCSA}}_{m}
=
\frac{
\widehat{\mathrm{PCSA}}_{m}^{(\mathrm{WPCS})}
+
\widehat{\mathrm{PCSA}}_{m}^{(\mathrm{BPCS})}
}{2}.
\label{eq:psm_pcsa_mean}
\end{equation}
The corresponding maximum muscle force was then scaled as \(F'_{\max,m}=40\,[\mathrm{N/cm}^2]\times\mathrm{PCSA}_m\) when PCSA is expressed in \(\mathrm{cm}^2\), consistent with the estimated intrinsic strength of the human jaw muscles~\citep{peck2000dynamic,weijs1985strength,zheng2019investigation}. Through this linear scaling, any residual error in the image-derived PCSA maps directly onto \(F_{\max}\); its effect on the objective is examined in the sensitivity analysis (Section~\ref{subsec:results_sensitivity}) and, as shown there, remains small. For each of the four masticatory muscle groups with patient-specific PCSA estimates, the group-level force capacity was then allocated across its corresponding branches using fixed proportions inherited from the generic functional model and consistent with the dynamic model of \citet{hannam2008dynamic,aftabi2024extent} and \citet{sagl2019dynamic}; these branch-allocation factors are listed in Appendix~\ref{app:functional_modeling_parameters}, Table~\ref{tab:func_supp_10}. Rest-length updating was also performed for the transferred stylomandibular and sphenomandibular ligaments inherited from the generic model. Denoting these ligament families by $g\in\{\mathrm{stm},\mathrm{sphm}\}$, the updated rest length was defined as
\begin{equation}
l'_{0,g}=l'_{\mathrm{cur},g}+\delta_g,
\label{eq:psm_lig_rest}
\end{equation}
where $l'_{\mathrm{cur},g}$ is the post-registration current ligament length and $\delta_{\mathrm{stm}}=1.5$ and $\delta_{\mathrm{sphm}}=5.5$ are bilateral slack offsets inherited from the generic model construction~\citep{aftabi2024extent,sagl2019dynamic}. This update preserves the intended passive slack behavior of these jaw-supporting ligaments after anatomical transfer and keeps the joint-support structures consistent with the registered geometry.

\subsubsection{Anatomy-Guided Registration of the Temporomandibular Joint}

The final stage of Figure~\ref{fig:psm_overview} concerns anatomy-guided registration of the TMJ disc and capsule. Direct delineation of the disc generally requires high-resolution MRI, which is not part of the planning pipeline, and these soft tissues are not reliably available from routine CT. The TMJ representation was therefore derived from the surrounding osseous anatomy rather than from direct soft-tissue segmentation. This approach is compared against an MRI-segmented disc in Section~\ref{subsec:results_validation} and is also consistent with prior patient-specific TMJ modeling work in which disc geometry was adapted from condylar and fossa anatomy~\citep{ahmadi2026computational}.

The generic disc and capsule were retained as anatomical priors, while their geometry was adapted using two complementary deformation fields inferred from the condyle and the articular fossa. For a TMJ point \(\mathbf{x}\), the blended update was defined as
\begin{equation}
\widehat{\mathbf{x}}_{\mathrm{TMJ}}
=
\mathbf{x}
+
w_{\mathrm{cond}}(\mathbf{x})\,\Delta_{\mathrm{cond}}(\mathbf{x})
+
w_{\mathrm{fossa}}(\mathbf{x})\,\Delta_{\mathrm{fossa}}(\mathbf{x}).
\label{eq:psm_tmj_blend}
\end{equation}
Here, \(w_{\mathrm{cond}}(\mathbf{x})+w_{\mathrm{fossa}}(\mathbf{x})=1\), and \(\Delta_{\mathrm{cond}}\) and \(\Delta_{\mathrm{fossa}}\) denote the condyle- and fossa-derived deformation fields, respectively. The spatial weights were determined from the relative proximity of \(\mathbf{x}\) to the condylar and fossa reference regions, with \(w_{\mathrm{fossa}}(\mathbf{x})=1-w_{\mathrm{cond}}(\mathbf{x})\):
\begin{equation}
w_{\mathrm{cond}}(\mathbf{x})
=
\frac{(d_{\mathrm{cond}}(\mathbf{x})+\varepsilon)^{-q}}
{(d_{\mathrm{cond}}(\mathbf{x})+\varepsilon)^{-q}
+
(d_{\mathrm{fossa}}(\mathbf{x})+\varepsilon)^{-q}}.
\label{eq:psm_tmj_weights}
\end{equation}
where \(d_{\mathrm{cond}}\) and \(d_{\mathrm{fossa}}\) denote local \(k\)-nearest-neighbor distances to the condylar and fossa reference point sets, \(q>0\) controls the decay of influence, and \(\varepsilon\) is a small stabilization constant. In the present implementation, the regional blending used $k_{\mathrm{NN}}=20$ neighboring reference points with $\varepsilon=10^{-8}$, while the distance-decay exponent was set to $q=0.5$ for the disc and $q=1.0$ for the capsule. This construction allowed the disc and capsule to follow the patient-specific joint neighborhood while maintaining smooth transitions between mandibular and cranial influence regions.

\begin{table*}[t]
\centering
\caption{Feasible-region bounds for the range parameters in Eq.~\ref{eq:structopt_feasible}. The angular range parameters $\alpha_r$, $\alpha_p$, $\beta_r$, and $\beta_p$ bound the design variables $\theta_{Lr}$, $\theta_{Lp}$, $\theta_{Rr}$, and $\theta_{Rp}$, respectively; for example, $\alpha_r=25^\circ$ denotes $\theta_{Lr}\in[-25^\circ,25^\circ]$. The length range parameters $z$ and $r$ bound $l_Z$ and $l_{RDP}$ in millimeters. A dash (---) indicates that $l_{RDP}$ is omitted for single-segment defects.}
\label{tab:design_bounds}
\footnotesize
\setlength{\tabcolsep}{3pt}
\renewcommand{\arraystretch}{1.1}
\begin{tabular*}{\textwidth}{@{\extracolsep{\fill}}lccccccc}
\toprule
Case & Defect & $\alpha_r$ (\textdegree) & $\alpha_p$ (\textdegree) & $\beta_r$ (\textdegree) & $\beta_p$ (\textdegree) & $z$ (mm) & $r$ (mm) \\
\midrule
Generic 1 & B  & 25 & 25 & 20 & 20 & 3.5  & --- \\
Generic 2 & S  & 15 & 15 & 15 & 15 & 5.0  & --- \\
Generic 3 & RB & 25 & 25 & 15 & 15 & 5.0  & 7   \\
Patient 1 & B  & 20 & 20 & 25 & 25 & 4.0  & --- \\
Patient 2 & S  & 10 & 15 & 10 & 15 & 5.0  & --- \\
Patient 3 & RB & 25 & 25 & 25 & 25 & 5.0 & 7   \\
\bottomrule
\end{tabular*}
\end{table*}

\begin{table*}[t]
\centering
\caption{CT-derived donor-bone descriptors and study-role indicators for all cases. For both the generic and patient-specific cases, mean cortical and cancellous HU values were computed from the donor segmentation, and the corresponding apparent densities were obtained from Eq.~\ref{eq:structopt_density}. Checkmarks indicate whether each case was used for optimization, longitudinal spatial analysis, and MRI/TMJ-disc evaluation.}
\label{tab:patient_bone_ct}
\scriptsize
\setlength{\tabcolsep}{6pt}
\renewcommand{\arraystretch}{1.2}
\begin{tabular*}{\textwidth}{@{\extracolsep{\fill}}@{}llccccccc@{}}
\toprule
Case & Donor & Optimization & Longitudinal Eval. & MRI/TMJ Eval. & Cortical HU & Cancellous HU & Cortical $\rho$ (g/cm$^3$) & Cancellous $\rho$ (g/cm$^3$) \\
\midrule
Generic 1--3 & Fibula  & $\checkmark$ & -- & -- & $\sim$1600 & $\sim$350 & 1.70 & 0.70 \\
Patient 1 & Fibula  & $\checkmark$ & $\checkmark$ & -- & $\sim$1400 & $\sim$550 & 1.54 & 0.85 \\
Patient 2 & Fibula  & $\checkmark$ & $\checkmark$ & -- & $\sim$1400 & $\sim$500 & 1.56 & 0.82 \\
Patient 3 & Fibula  & $\checkmark$ & $\checkmark$ & -- & $\sim$1500 & $\sim$300 & 1.64 & 0.70 \\
Patient 4 & Scapula & -- & $\checkmark$ & $\checkmark$ & $\sim$1250 & $\sim$350 & 1.45 & 0.70 \\
\bottomrule
\end{tabular*}
\end{table*}

The resulting disc and capsule should therefore be interpreted as patient-specific, anatomy-guided approximations rather than direct segmentations of the true soft tissues~\citep{sagl2019dynamic,sagl2021silico}. To quantify the effect of this approximation, we additionally analyzed the sensitivity of the apposition fraction over one chewing cycle to using the anatomy-guided disc versus a real disc segmentation transferred by MRI-to-CT registration. Once the skeletal correspondence, attachment transfer, parameter updates, and TMJ blending have all been applied, the resulting patient-specific digital twin $\mathcal{M}_{\mathrm{ps}}$ remains fully compatible with the generic model used in Section~\ref{subsec:opt_workflow}, and can therefore be substituted directly into the same virtual-planning, mesh-refinement, simulation, and Bayesian-optimization loop described earlier.

\subsection{Experimental Setup}\label{subsec:experimental_design}

This subsection summarizes the experimental design used to evaluate the proposed workflow. The study is organized around four components: (i) \emph{generic-model optimization} on synthetic body (B), symphysis (S), and ramus-body (RB) defect scenarios; (ii) \emph{patient-specific optimization} on three real reconstruction cases matched to the generic defect categories; (iii) a systematic \emph{sensitivity analysis} on key model parameters; and (iv) \emph{longitudinal analysis} of the predicted bone-union outcome on four cases with paired day-5 and year-1 postoperative CT. For the retrospective patient-specific experiments, pre-op CT was used to construct the planning model, day-5 post-op CT was used to recover the actual surgical cut-plane configuration implemented by the surgeon and to define the patient-specific baseline (zero) setting, and year-1 post-op CT was used only in the longitudinal spatial analysis cases to assess the observed union outcomes. The overall workflow is intended as an offline planning tool rather than an intraoperative one. The experimental parameters defining the feasible region for each case are summarized in Table~\ref{tab:design_bounds}, and the corresponding CT-derived donor-bone descriptors and cohort roles are summarized in Table~\ref{tab:patient_bone_ct}.

\subsubsection{Generic-Model Optimization}

The optimization workflow was first evaluated using three generic-model defect scenarios representing B, S, and RB defects. Although these cases were synthetically imposed on a generic craniofacial model, they were defined using real imaging-based anatomy and defect-specific surgical constraints, as described in Section~\ref{design_variables}, thereby preserving realistic reconstruction geometry. This controlled setting isolates the contribution of the optimization approach while relying on a baseline craniofacial model that has already been validated across multiple defect scenarios~\citep{aftabi2024extent}.

For each defect class, Bayesian optimization was repeated for five random seeds, with 25 Sobol' initial samples followed by 50 sequential iterations under both objective functions $F_{\mathrm{opt}}$ and $F_{\mathrm{SF}}$. To examine the influence of the design variables on the cost function, one additional 150-iteration run was performed for each defect, and the results were summarized using parallel-coordinate analysis. The feasible-region bounds are listed in Table~\ref{tab:design_bounds} and follow the same physical, surgical, and anatomical criteria described in Section~\ref{subsec:bo_formulation}. For the synthetic cases, the reference cut-plane configuration was defined from common surgical practice and served as the baseline setting.

\subsubsection{Patient-Specific Optimization}

The patient-specific optimization pipeline was evaluated on the three real cases marked for patient-specific optimization in Table~\ref{tab:patient_bone_ct}, selected to mirror the B, S, and RB generic defect classes while preserving case-specific anatomy, donor properties, and cohort role. Across these cases, the B defect was located on the patient left, the S defect was located at the midline, and the RB defect was located on the patient right, indicating that the workflow is not restricted to a single defect side. For all three patients, pre-op CT was used to construct the patient-specific digital twin described in Section~\ref{sec:functional_patient_specific_modeling} and to drive the optimization loop. Day-5 post-op CT was used retrospectively to recover the actual surgical cut-plane configuration implemented by the surgeon. This realized surgical configuration defined the patient-specific baseline setting and the corresponding case-specific optimization bounds listed in Table~\ref{tab:design_bounds}. It was produced with the same geometric, RDP-based VSP workflow and executed through patient-specific 3D-printed cutting guides, the current clinical standard rather than a freehand or arbitrary setting. The same optimization protocol as in the generic cases was applied, namely five random seeds with 25 Sobol' points followed by 50 sequential iterations under both objective functions $F_{\mathrm{opt}}$ and $F_{\mathrm{SF}}$.

\subsubsection{Sensitivity Analysis}

Sensitivity analysis was performed to assess robustness to modeling uncertainty. Eleven parameters spanning bone material properties, elastic-contact parameters, and muscle parameters were independently perturbed by $\pm 10\%$, and the resulting change in $F_{\mathrm{opt}}$ was evaluated for all three generic and all three patient-specific cases. Each perturbation was repeated five times and averaged, yielding a total of $11 \times 2 \times 5 \times 6 = 660$ simulations. The analysis was restricted to $F_{\mathrm{opt}}$ as $F_{\mathrm{SF}}$ adds a safety-factor penalty to the same apposition-driven signal and is therefore expected to show the same dominant trends. Robustness to the chewing-activation assumption was assessed separately by repeating the optimization under the adapted activation profile from inverse simulation (Appendix~\ref{app:forward_inverse}) and comparing the recovered design variables with the normal-activation optimum.

\subsubsection{Longitudinal Analysis}

The SED-based bone-remodeling stimulus adopted in Section~\ref{subsec:opt_workflow}, motivated by Wolff’s law, is well established and has been extensively validated in the broader bone-remodeling literature~\citep{field2010prediction,zheng2022,ferguson2022,wu2021machine,lin2010bone,rungsiyakull2011loading}. Therefore, the aim of the present analysis was not to re-validate the remodeling formulation itself, but rather to assess its applicability within the proposed optimization and patient-specific modeling workflow. The objective is not to forecast the full patient-specific remodeling trajectory or absolute clinical union, since faithful prediction of remodeling kinetics would require additional intermediate-time-point imaging to inverse-identify subject-specific remodeling parameters~\citep{zheng2022,zheng2019investigation}; instead, we examine whether the apposition metric extracted from the day-5 patient-specific model is spatially consistent with corticalized bone formation observed at year 1. This assessment requires longitudinal imaging data between the early and late postoperative stages, e.g., between day 5 post-op and year 1 post-op. Since such paired scans are difficult to obtain routinely, the evaluation was performed on the four cases for which both scans were available. The analysis therefore focused on spatial agreement between the predicted donor-mandible apposition pattern and observed follow-up bone formation, since alternative cut-plane configurations cannot be implemented and compared within the same patient.

The four longitudinal cases (Patients~1--4, Table~\ref{tab:patient_bone_ct}) included the three patient-specific optimization cases and Patient~4; together they spanned the B, S, and RB defects and both fibula and scapula donors, and each had paired day-5 and year-1 post-op CT. The day-5 post-op scans were used to construct the patient-specific digital twins and to run the retrospective longitudinal analysis following the pipelines described in Sections~\ref{subsec:opt_workflow}--\ref{sec:functional_patient_specific_modeling}, without any additional modeling steps beyond those sections. The year-1 post-op scans provided the observed union outcomes. For Patient~4, a postoperative T1-weighted MRI was additionally available and was registered to the day-5 post-op CT using Elastix~\citep{klein2009elastix} to enable direct segmentation of the TMJ disc. Only Patient~4 had postoperative MRI, so only Patient~4 was used for the MRI-based TMJ-disc evaluation. This MRI-based disc geometry, rarely obtainable in this setting, served as a reference against which the anatomy-guided disc approximation of Section~\ref{sec:functional_patient_specific_modeling} was evaluated by comparing the apposition fraction over one chewing cycle obtained with the anatomy-guided disc to that obtained with the MRI-segmented disc.

The apposition-based surrogate was evaluated over a full chewing cycle by tracking the apposition fraction at the donor-mandible interface on both resection sides. To assess spatial agreement, the year-1 post-op CT was segmented and a thin region of interest near the mandible-donor resection interface was extracted to quantify the observed bone formation. The thickness of this region of interest was matched to the element edge length of the interface layer in the simulation, so that the imaging-derived and simulation-derived signals were compared on geometrically consistent supports. Within this region, cortical bone percentage ($\mathrm{cort}\%$) was computed using an operational HU threshold of $1000$~\citep{mahesh2013essential}, a known marker of mature bone formation and successful union under sustained interface loading~\citep{zheng2022}, yielding a binary bone-formation label map. The same $\mathrm{cort}\%$ was then applied in the simulation to rank and select the corresponding top fraction of interface elements that reached the apposition threshold and maintained sustained contact over the chewing cycle~\citep{prasad2019invertible,aftabi2024extent}, representing regions of prolonged bone-to-bone interaction. The centroids of these elements were finally interpolated onto the imaging voxel grid, allowing Dice overlap and centroid difference to quantify agreement between the predicted-apposition and observed-bone-formation patterns. Because the predicted and observed regions were matched in area, these metrics assess the spatial localization of subsequent bone formation rather than its total amount. To assess robustness to these analysis choices, the Dice overlap and centroid difference were recomputed over three segmentation HU thresholds (900, 1000, and 1100) and three interface-layer thicknesses (one, two, and three times the element edge length).

\subsubsection{Computational Implementation and Runtime}
The core optimization controller and patient-specific modeling pipeline were implemented in MATLAB. MATLAB coordinated the workflow by calling the required external components, including the virtual-planning and forward-simulation routines in ArtiSynth~\citep{lloyd2012artisynth}, the surface-remeshing routines through the PYMeshLab API~\citep{pymeshlab}, and the template-to-patient coherent point drift registration through the PyCPD library~\citep{gatti2022pycpd}. The public repository contains the workflow code for virtual reconstruction, remeshing, ArtiSynth-based simulation setup and execution, patient-specific digital-twin construction, and Bayesian optimization, together with setup instructions for ArtiSynth and the associated dependencies. For the patient-specific cases, the template-to-patient pipeline of Section~\ref{sec:functional_patient_specific_modeling} was executed once before optimization to produce $\mathcal{M}_{\mathrm{ps}}$, and the resulting digital twin was then reused at every subsequent iteration without further registration cost. For each candidate point, the workflow generated a new reconstruction, remeshed the relevant surfaces, performed a dynamic chewing simulation, and evaluated the objective over one chewing cycle; candidates whose geometry or plate generation failed were rejected by exception handling and excluded from the objective. The dynamic simulation used a constrained backward-Euler integrator with a time step of $0.001$~s, and all experiments were carried out on a desktop workstation (Intel Core i7-10700F @ 2.9~GHz, 16~GB RAM, NVIDIA RTX~3060), resulting in an average serial runtime of approximately 6 minutes per iteration. The patient-specific registration and optimization run automatically. The one-time operator input, about 30~minutes per case and not repeated during optimization, comprises the semi-automatic mandible and donor segmentation with a muscle-segmentation check, the mandibular-contour landmarks, and the resection-plane positions, which the surgeon defines from the tumor location and required margin.
Because the workflow is intended for offline preoperative planning rather than intraoperative use, this runtime should be read against the pre-surgical planning interval rather than a real-time budget. Each evaluation required only about 2~GB of memory, keeping the workflow within the capacity of a standard desktop workstation. The patient-specific twin is built once and reused across iterations, and the 75-evaluation budget adopted here is a conservative demonstration cap rather than a clinical requirement; in practice the search converges in significantly fewer evaluations, as demonstrated later. Several properties keep the wall-clock cost modest: the 25 Sobol initialization points are mutually independent and parallelize to about 30--45 minutes with 4--6 workers, the generic-model optima provide warm starts for the patient-specific runs, and the sequential Bayesian-optimization stage can be batched or run asynchronously. Because patients are processed independently, the cost also scales linearly across a larger cohort, and further acceleration through early stopping and parallel candidate evaluation remains available.

\section{Results and Discussion}

This section reports and interprets the four experimental components introduced in Section~\ref{subsec:experimental_design}.

\subsection{Generic-Model Optimization}\label{subsec:results_generic}

\begin{figure*}[t]
\centering

\hspace{-1cm}\includegraphics[width=.21\textwidth]{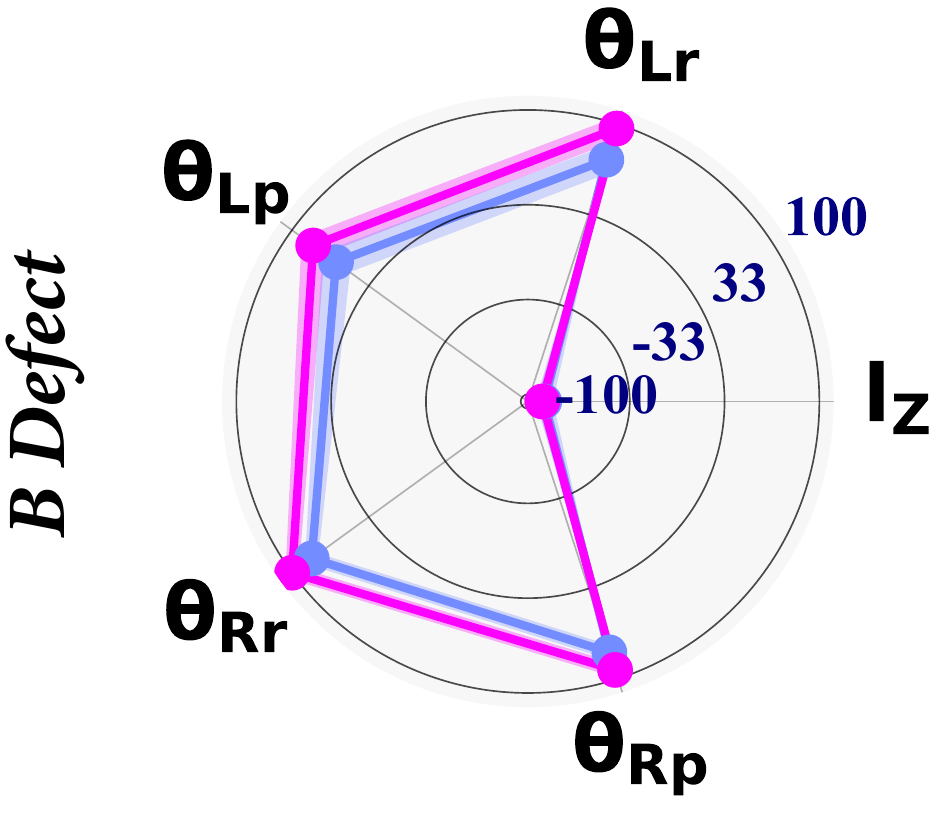}
\hspace{.6cm}
\includegraphics[width=.23\textwidth]{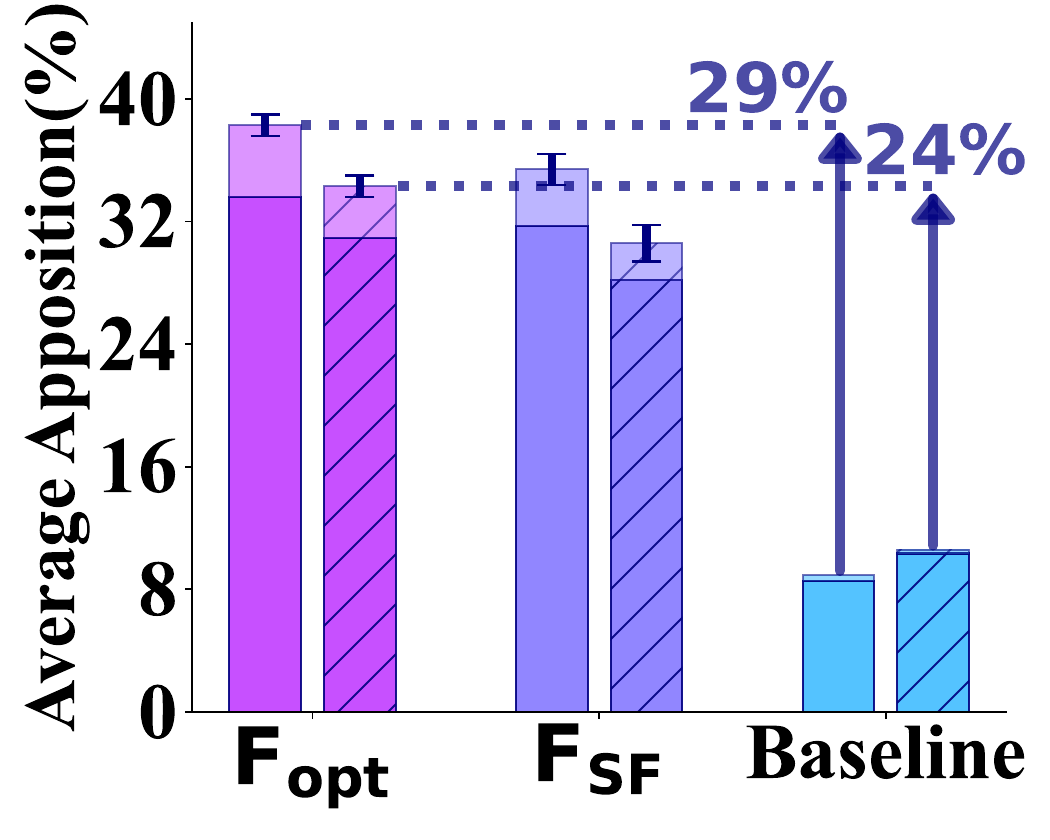}
\hspace{.7cm}
\raisebox{-1mm}{\includegraphics[width=.23\textwidth]{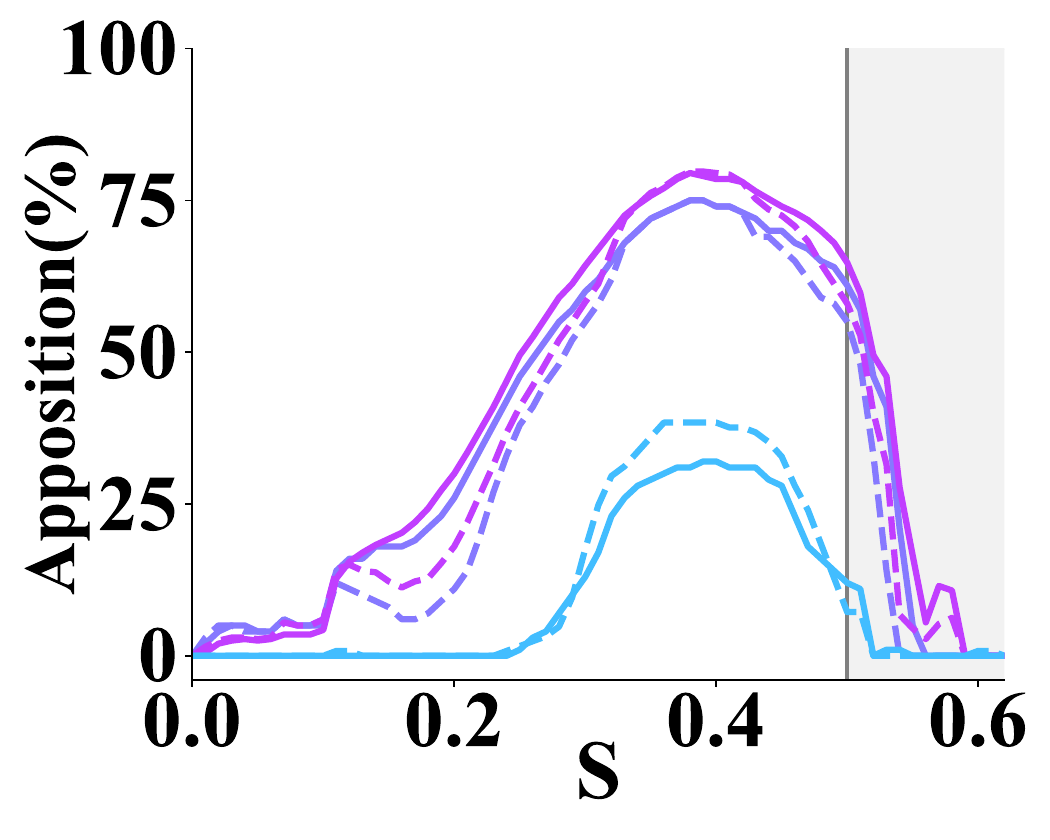}}

\vspace{-.05cm}

\hspace{-1cm}\includegraphics[width=.21\textwidth]{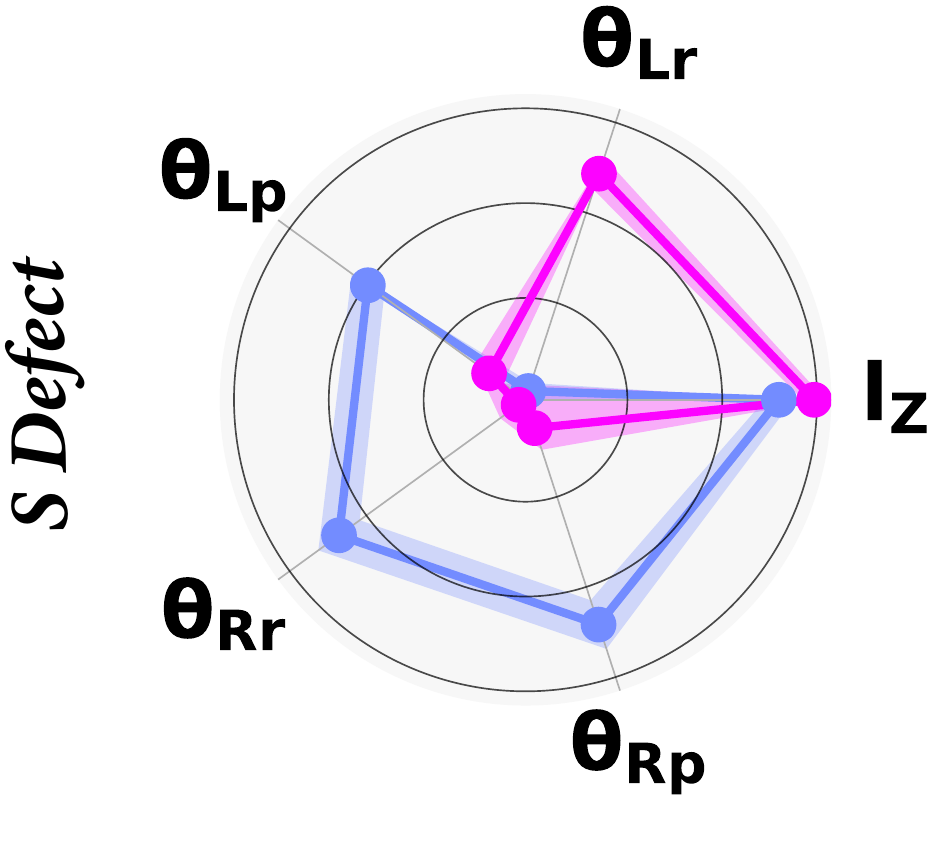}
\hspace{.6cm}
\includegraphics[width=.23\textwidth]{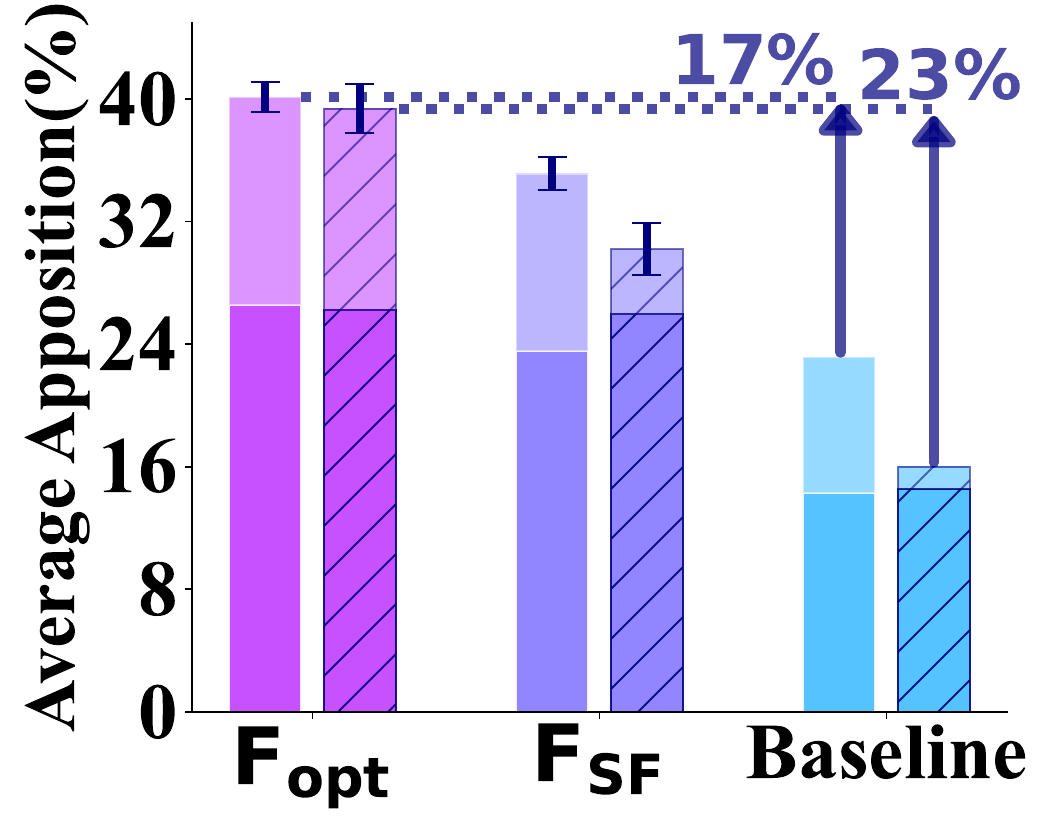}
\hspace{.7cm}
\raisebox{-1mm}{\includegraphics[width=.23\textwidth]{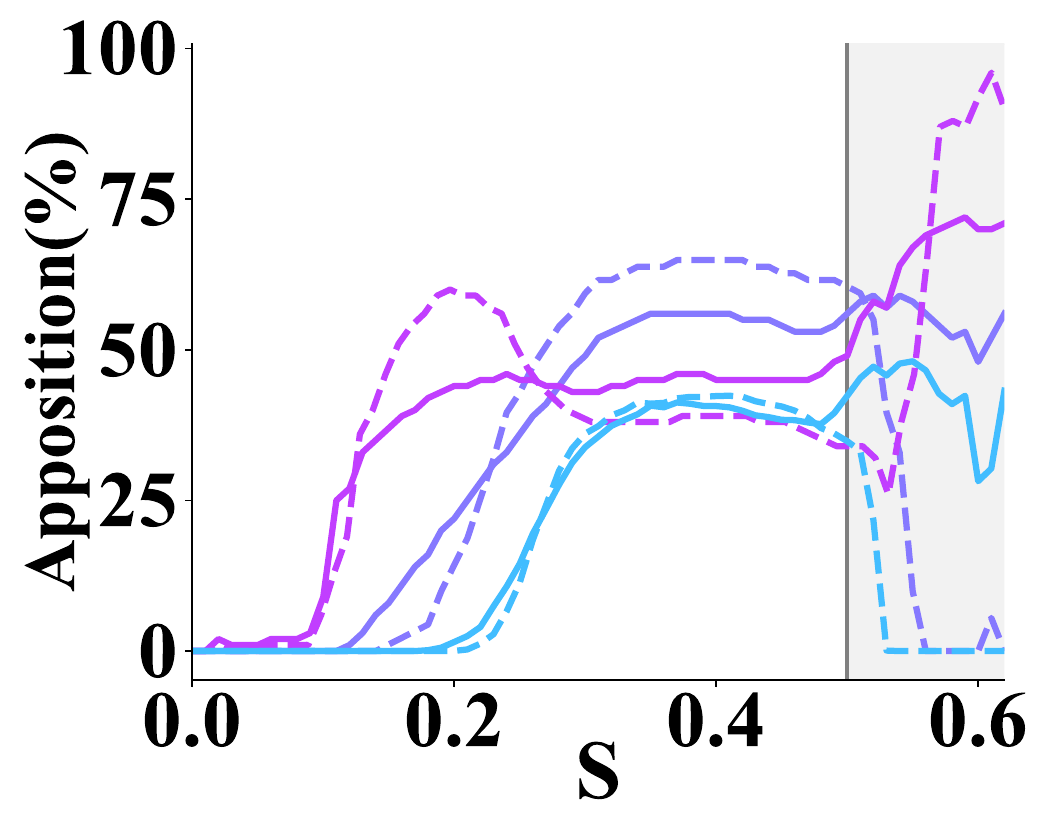}}

\vspace{-.05cm}

\hspace{-1cm}\includegraphics[width=.21\textwidth]{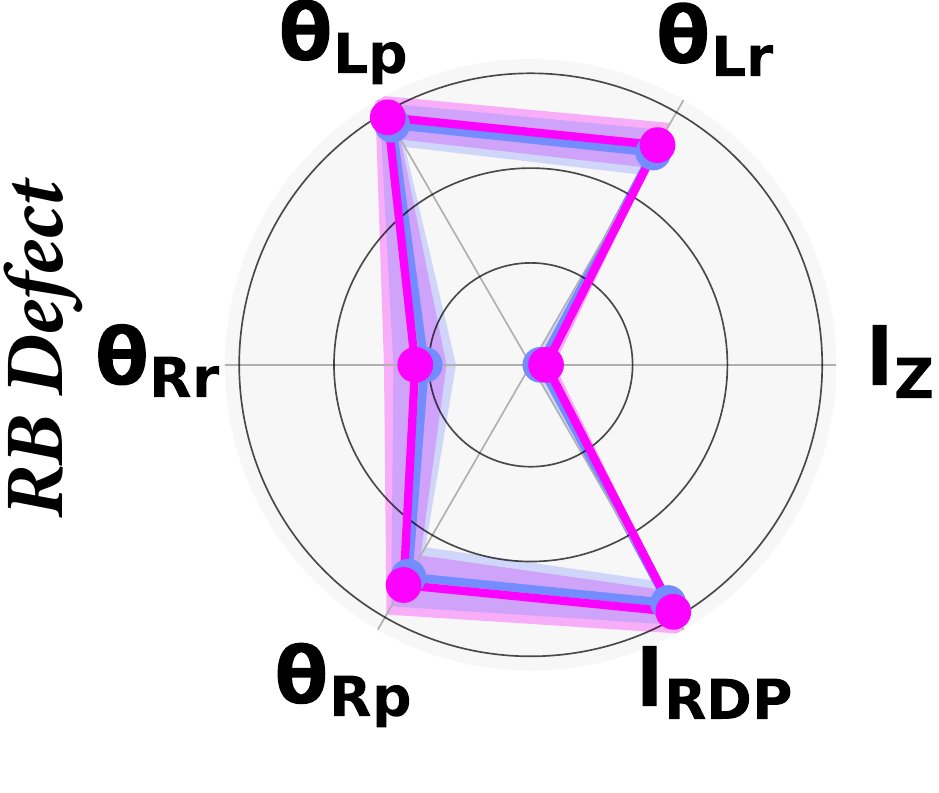}
\hspace{.6cm}
\includegraphics[width=.23\textwidth]{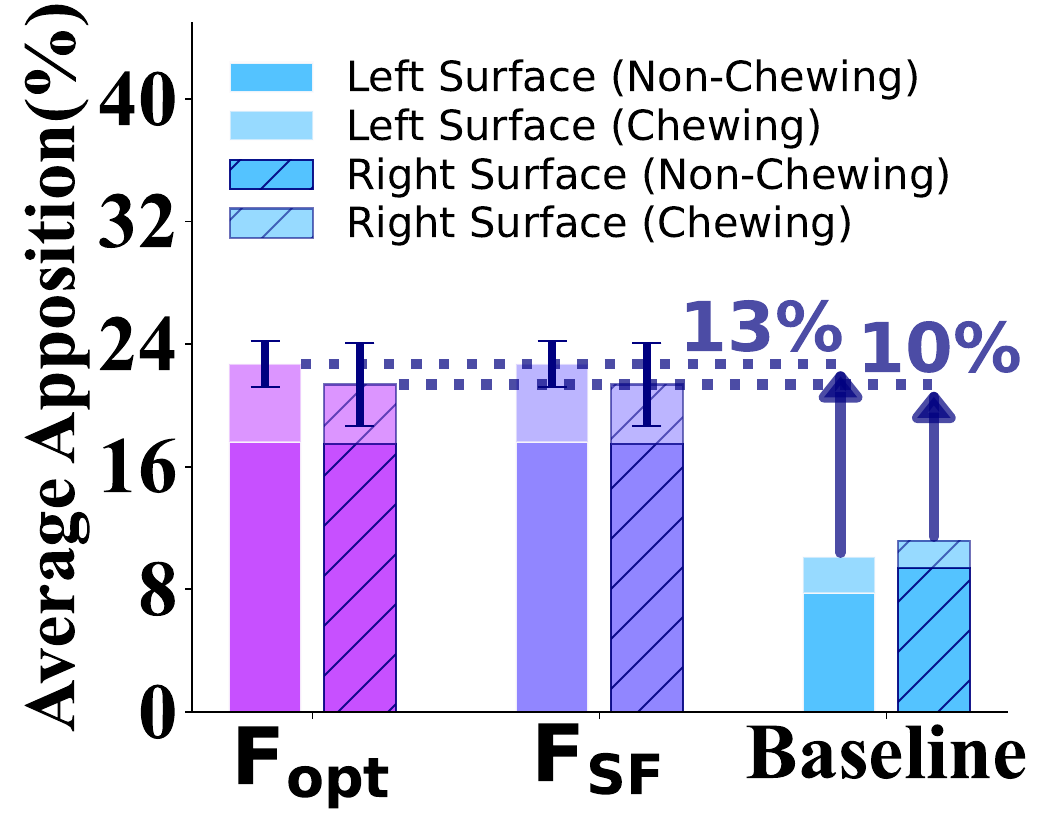}
\hspace{.7cm}
\raisebox{-1mm}{\includegraphics[width=.23\textwidth]{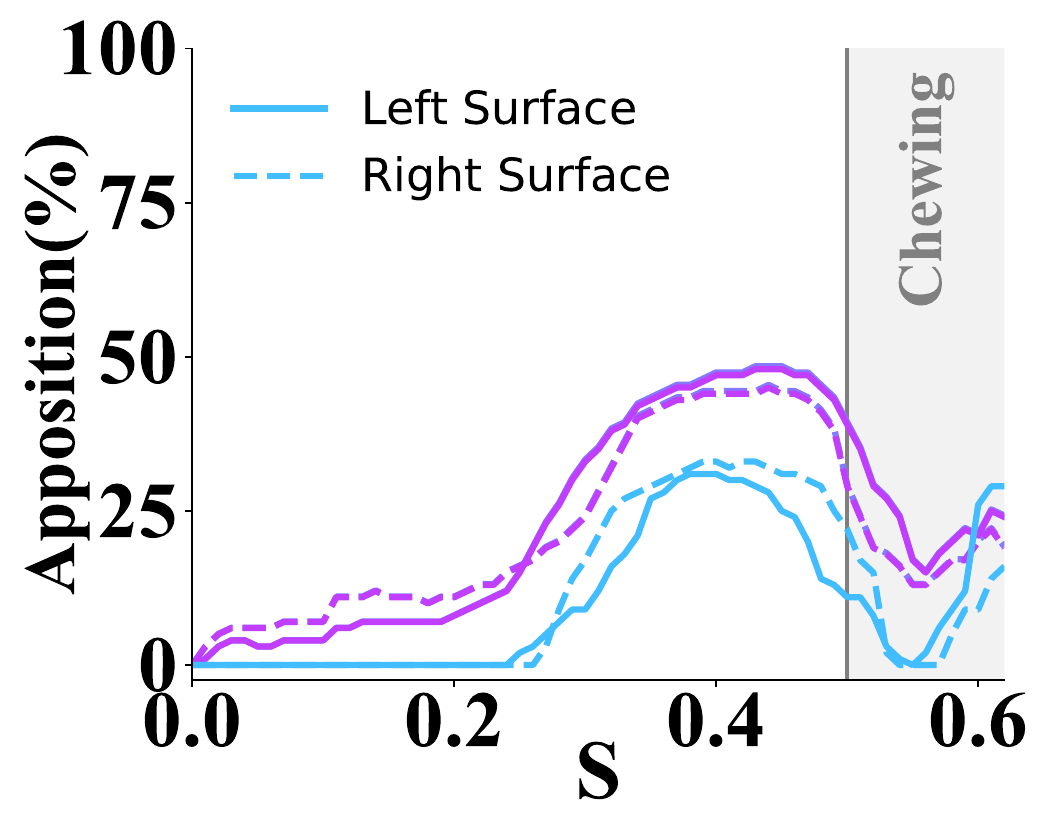}}

\caption{\textbf{Optimization outcomes for the generic defect cases.} Rows show B, S, and RB defects; columns show optimized design parameters mapped to \([-100,100]\), mean donor-mandible apposition over one chewing cycle, and the corresponding apposition trajectory. The gray interval marks bolus engagement. Results are shown for \textcolor{myhexcolor3}{baseline} reconstructions, representing common surgical practice with zero design-variable offsets, and for reconstructions optimized using \textcolor{myhexcolor1}{\(F_{\mathrm{opt}}\)} and \textcolor{myhexcolor2}{\(F_{\mathrm{SF}}\)}. All optimized values are averaged over five independent optimization trials with different random seeds. Percentage differences are reported as absolute values. Reproduced from \citet{aftabi2025osteoopt}.}
\label{fig:fig_param_generic}

\vspace{7mm}

\includegraphics[width=1\textwidth]{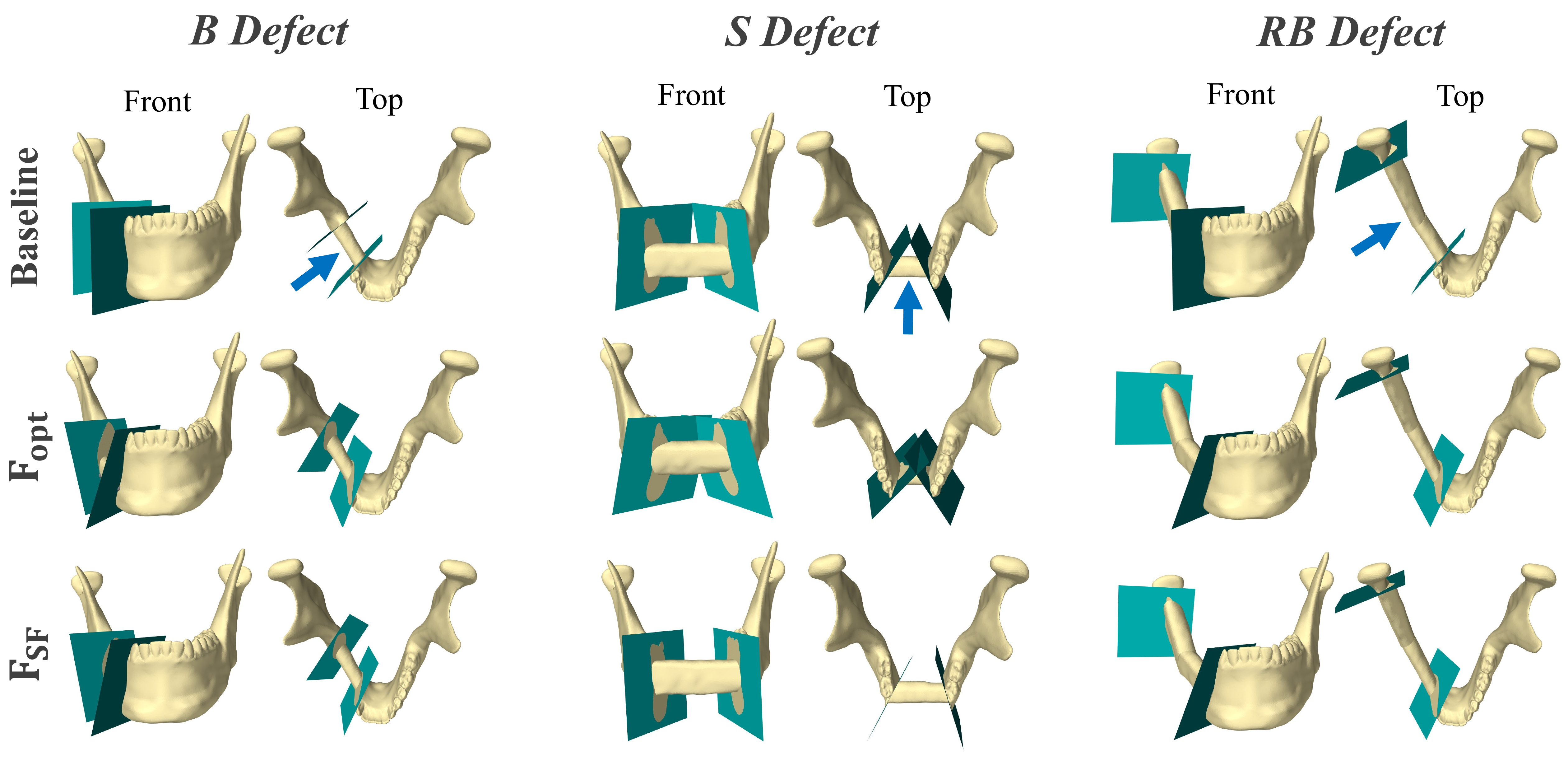}

\caption{\textbf{Generic reconstruction configurations.} Baseline reconstructions, representing common surgical practice, and the optimized \(F_{\mathrm{opt}}\) and \(F_{\mathrm{SF}}\) reconstructions are shown for the B, S, and RB defect cases from front and top views; the blue arrow in the baseline row marks the resection interface.}
\label{fig:generic_recon}

\end{figure*}

Across all three Urken defect classes, Bayesian optimization produced reconstructions with substantially higher cycle-averaged donor-mandible apposition than the baseline configurations representing common surgical practice. Figure~\ref{fig:fig_param_generic} summarizes the optimized design parameters, the mean apposition over one chewing cycle, and the corresponding apposition trajectories for the B, S, and RB defects, all averaged over five independent optimization trials with different random seeds, while Figure~\ref{fig:generic_recon} shows the corresponding three-dimensional reconstructions. Relative to the baseline, the $F_{\mathrm{opt}}$ configurations increased mean apposition by roughly $24$--$29$ percentage points on the two interfaces of the B defect, $17$--$23$ percentage points on the S defect, and $10$--$13$ percentage points on the more anatomically constrained RB defect. We report these effects as absolute percentage-point gains, which are more directly interpretable across different baseline apposition levels. The $F_{\mathrm{SF}}$ remained close to $F_{\mathrm{opt}}$ in terms of improvement and yielded comparable cycle-averaged apposition. In all three defect classes, the difference between left and right interface apposition was reduced relative to baseline, consistent with the imbalance penalty in Eq.~\ref{eq:structopt_fopt} that explicitly favors symmetric loading across the resection sides~\citep{aftabi2025osteoopt}.

\begin{figure*}[t]
    \centering
	    \hspace{-.5cm}\begin{subfigure}{0.36\textwidth}
	        \centering
	        \includegraphics[width=\linewidth]{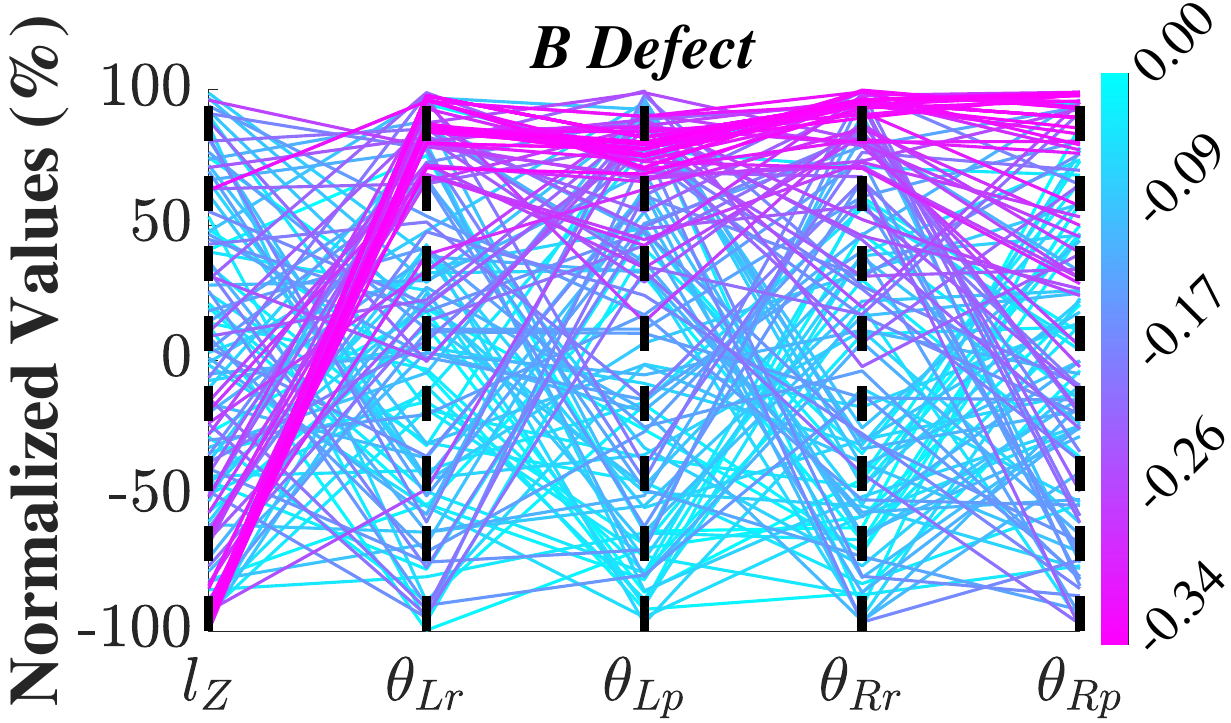}
	        \caption*{(a)}
	        \label{fig_par_b}
	    \end{subfigure}
	    \hspace{1cm}
	    \begin{subfigure}{0.33\textwidth}
	        \centering
	        \includegraphics[width=\linewidth]{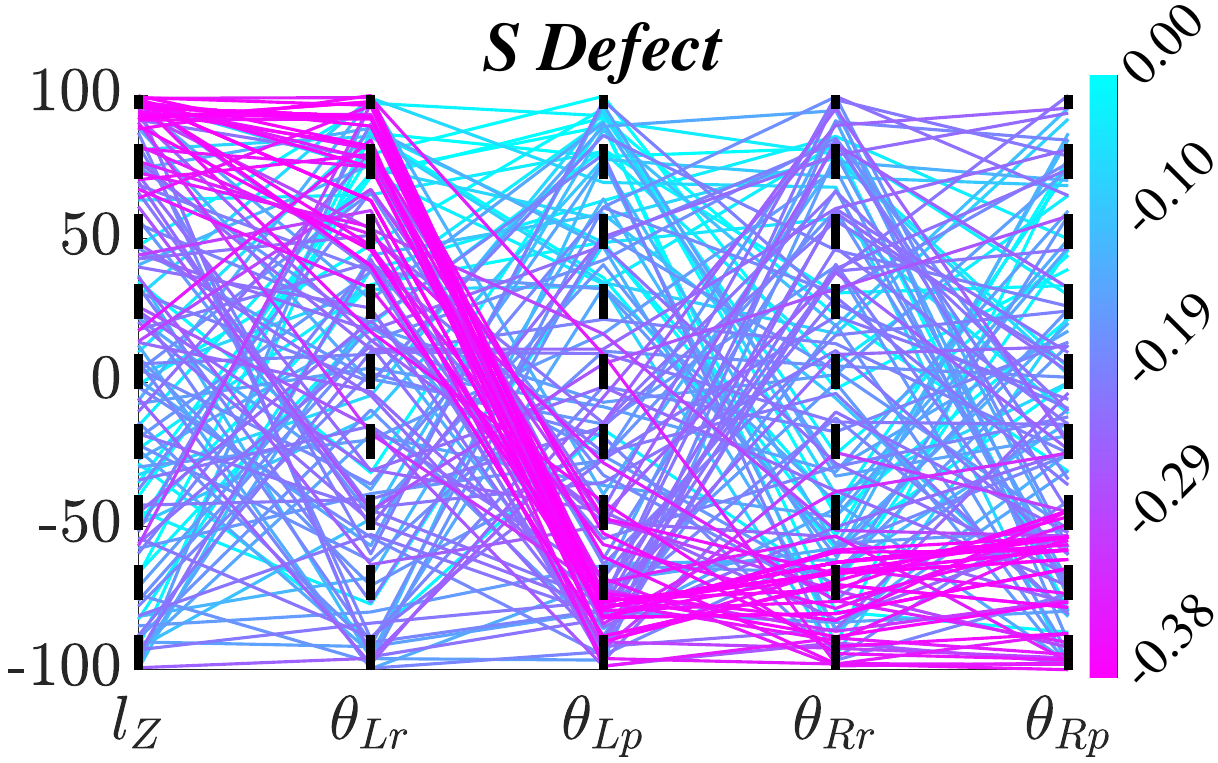}
	        \caption*{(b)}
	        \label{fig_par_s}
	    \end{subfigure}
	    \begin{subfigure}{0.33\textwidth}
	        \centering
	        \includegraphics[width=\linewidth]{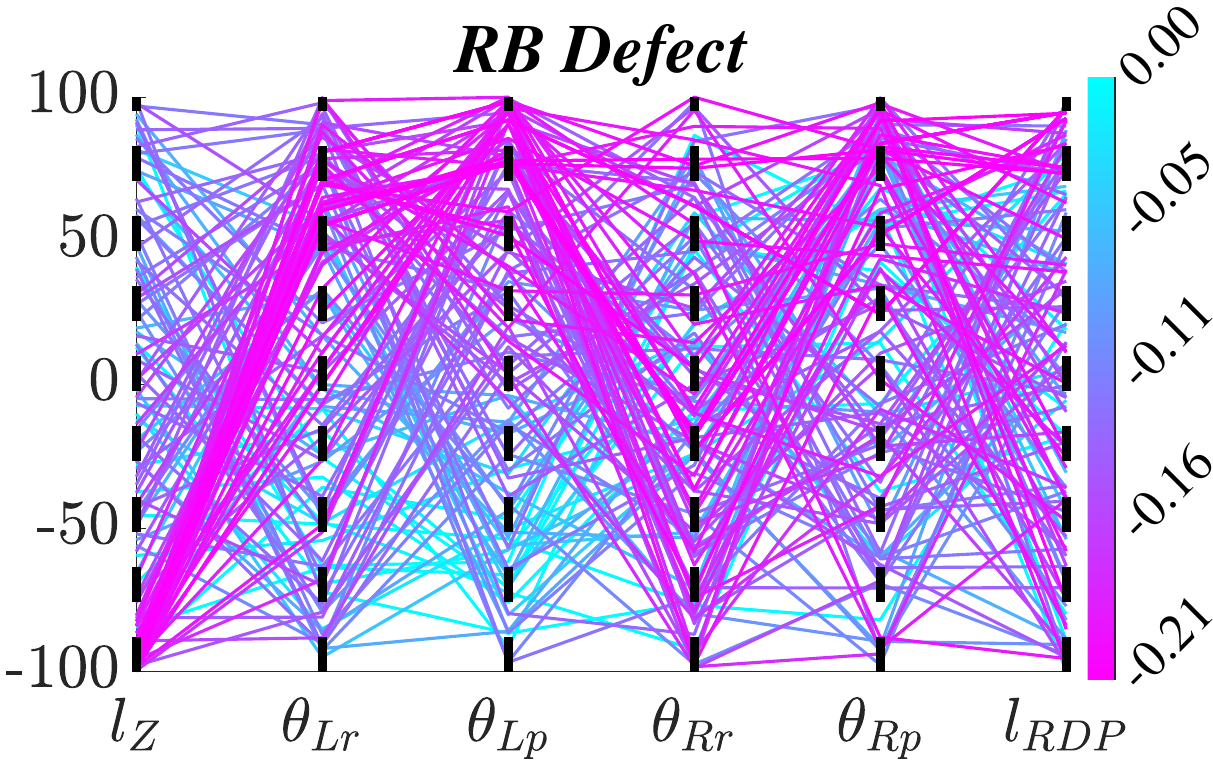}
	        \caption*{(c)}
	        \label{fig_par_rb}
	    \end{subfigure}
	    \hspace{.9cm}
	    \begin{subfigure}{0.345\textwidth}
	        \centering
	        \includegraphics[width=\linewidth]{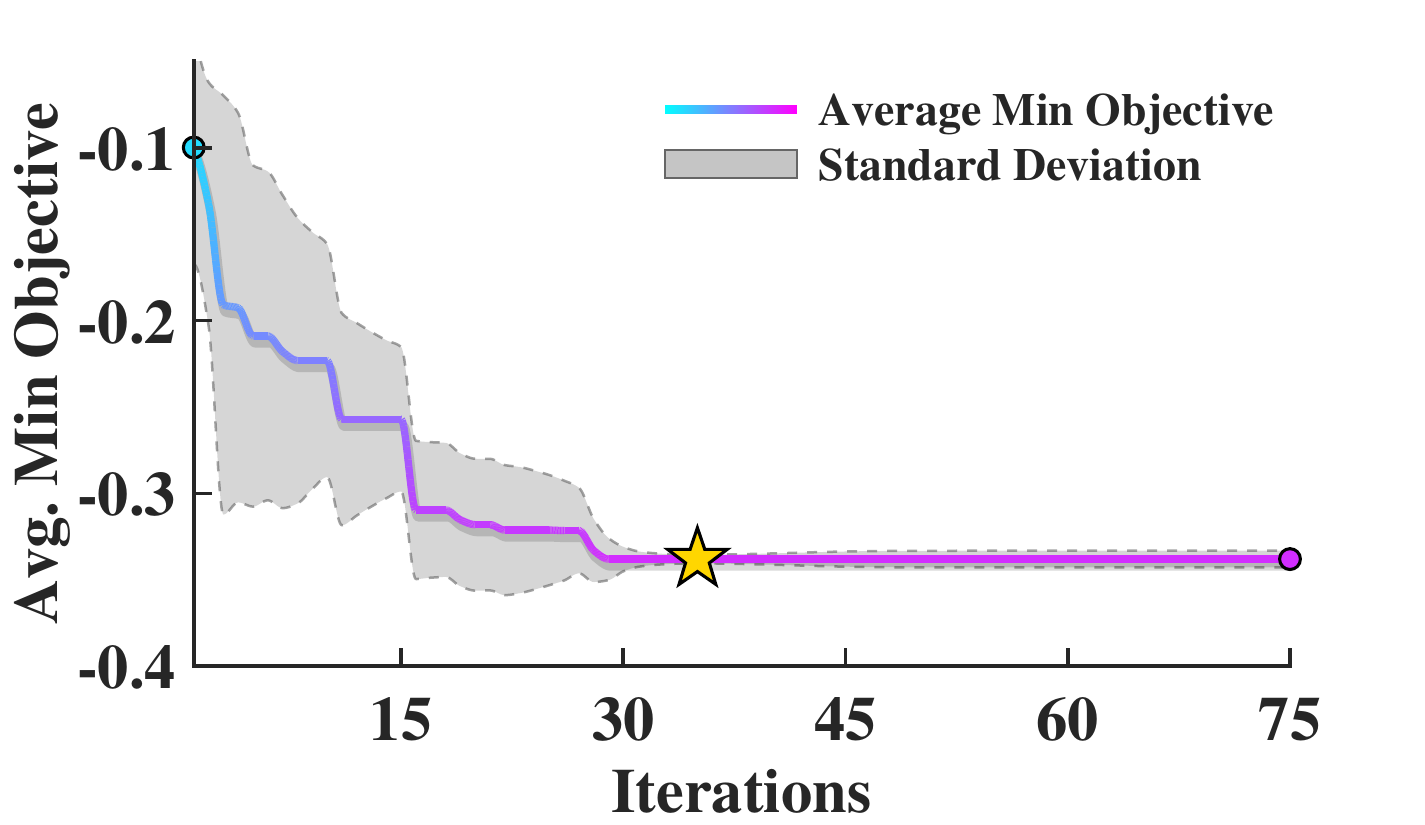}
	        \caption*{(d)}
	        \label{fig_par_conv}
	    \end{subfigure}
	    \caption{\textbf{Optimization behavior for the \(F_{\mathrm{opt}}\) objective.} Panels (a)--(c) show 150-iteration parallel-coordinate analyses from a single representative trial, relating normalized design parameters to the objective value for the generic B, S, and RB defect cases, respectively. Parameters are normalized to \([-100,100]\), and \textcolor{myhexcolor0}{magenta} highlights the optimal region. Panel (d) shows an example of a best-so-far convergence profile for the B defect over 75 iterations across five optimization trials. All panels show the negative of the optimized objective function. Panels (a)--(c) reproduced from \citet{aftabi2025osteoopt}.}
    \label{fig_par}
\end{figure*}

Before interpreting the optimized configurations, we first verify that the optimizer is well behaved. The 150-iteration parallel-coordinate analyses in Figure~\ref{fig_par}(a)--(c) show how the design variables interact with the cost function. For each defect, the magenta high-quality region concentrates within a narrow band of cut-plane angles and donor offsets rather than spanning the entire feasible region, indicating that the optimization landscape is informative rather than flat and that the union-related objective is not equally satisfied across the surgically reachable design space. The B and S defects reveal compact optimal clusters dominated by the angular variables, whereas the RB defect, which adds the intermediate cut-plane offset $l_{RDP}$ as a sixth degree of freedom, exhibits a broader optimal region--reflecting the additional geometric coupling introduced by two donor segments. The example convergence profile for B defect in Figure~\ref{fig_par}(d) shows that the best-so-far objective stabilizes by iteration~$\sim$35 and remains essentially unchanged thereafter, while the inter-trial standard deviation contracts steadily over iterations. This is the expected behavior of an EI+ acquisition rule on an expensive but smooth surrogate: early Sobol' coverage establishes the global landscape, and the EI+ safeguard prevents premature exploitation by inflating posterior uncertainty whenever it falls below $t_{\sigma}\sigma$. Because each evaluation requires a full geometry-generation, remeshing, and chewing simulation, this convergence speed is central to the workflow's clinical practicality.

The trajectories in the third column of Figure~\ref{fig:fig_param_generic} clarify where these gains originate, and the picture is defect-dependent rather than uniform. For the B defect, the optimized configurations achieve their largest apposition during the early-to-middle phase of the chewing cycle, before bolus engagement (gray interval), with peak instantaneous apposition reaching roughly $75$--$80\%$ compared with $25$--$40\%$ at baseline. For the S defect, however, the gains are not confined to the pre-bolus phase: $F_{\mathrm{opt}}$ increases apposition during bolus engagement, whereas $F_{\mathrm{SF}}$ adopts a markedly different strategy by reducing apposition in the bolus-engagement interval while simultaneously increasing the lateral, pre-bolus apposition, yielding a flatter and safer loading profile in the regime where peak principal stresses would otherwise approach yield. This contrast illustrates that $F_{\mathrm{opt}}$ and $F_{\mathrm{SF}}$ are not redundant objectives: they may redistribute interface loading differently across the chewing cycle, depending on the scenario.

The radar charts in the first column of Figure~\ref{fig:fig_param_generic} summarize the optimized design-variable values for each defect on the normalized $[-100,100]$ scale. The corresponding values in physical units, with zero denoting the surgeon-reference configuration, are listed in Appendix Table~\ref{tab:optimized_values}. For the B and RB defects, $F_{\mathrm{opt}}$ and $F_{\mathrm{SF}}$ recover nearly identical configurations, with the angular cut-plane variables settling toward the upper end of their feasible ranges and the donor offsets converging on a shared optimum. For the S defect, in contrast, the two objectives diverge: their angular patterns are essentially mirrored, while both still push $l_Z$ to its upper bound. This contrast confirms that the safety-factor regularization is not a redundant rescaling of $F_{\mathrm{opt}}$ but can select a qualitatively different region of the surgical design space. In the generic setting, $F_{\mathrm{SF}}$ may therefore yield a suboptimal apposition relative to $F_{\mathrm{opt}}$ for certain defects; this divergence, however, largely disappears in the patient-specific cases discussed in Section~\ref{subsec:results_patient}, where the $F_{\mathrm{SF}}$ optima sit very close to those of $F_{\mathrm{opt}}$ and the two objectives are not in conflict. The optimized configurations are also consistent with the surgical observation that better bony contact correlates with successful union~\citep{kim2024optimizing,wang2016mandibular}.

Taken together, the generic-model results indicate that the proposed cost functions are well posed, that meaningful improvements over common-practice baselines are recoverable in only tens of iterations, and that the optima tighten around clinically interpretable cut-plane configurations rather than collapsing onto edge cases~\citep{aftabi2025optimizing,aftabi2025osteoopt}.

\subsection{Patient-Specific Optimization}\label{subsec:results_patient}

\begin{figure*}[t]
\centering

\hspace{-1cm}\includegraphics[width=.21\textwidth]{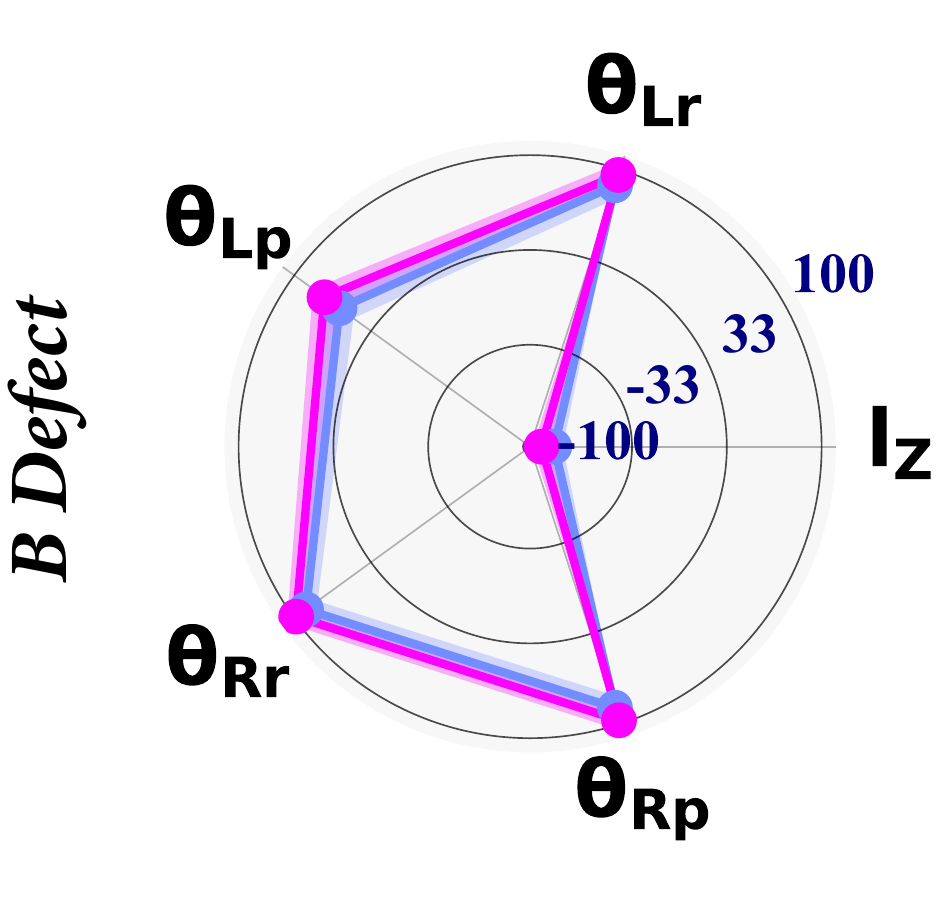}
\hspace{.6cm}
\includegraphics[width=.23\textwidth]{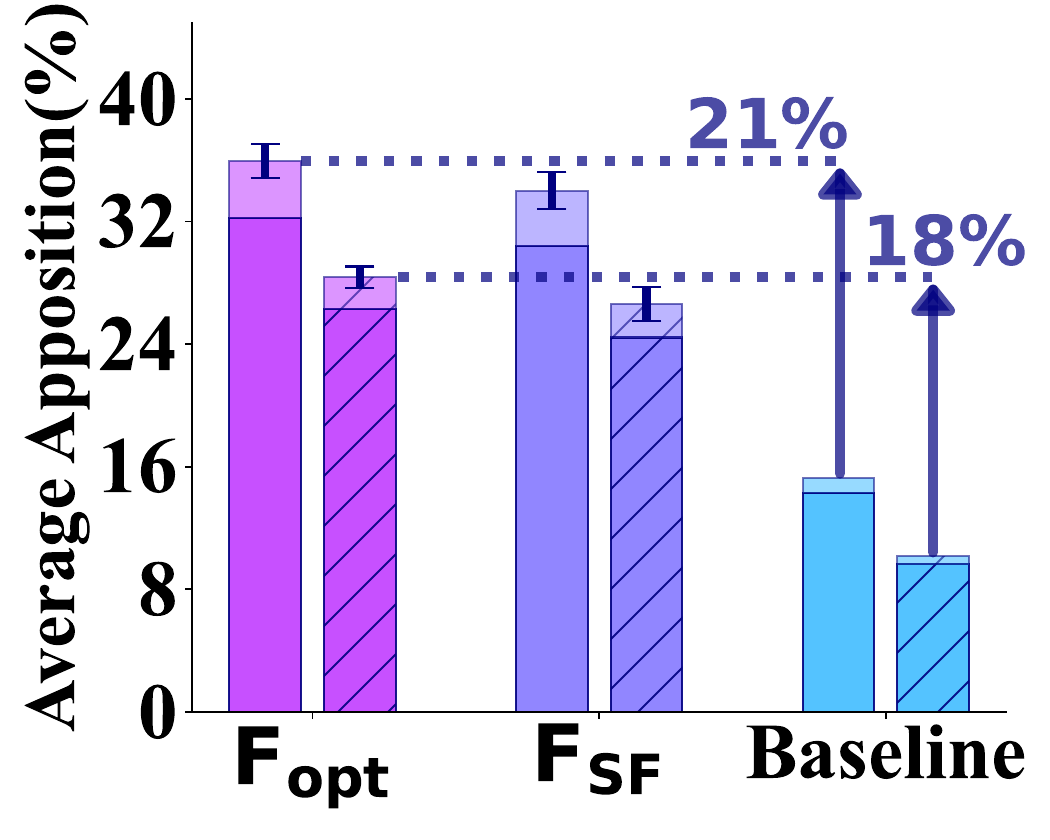}
\hspace{.7cm}
\raisebox{-1mm}{\includegraphics[width=.23\textwidth]{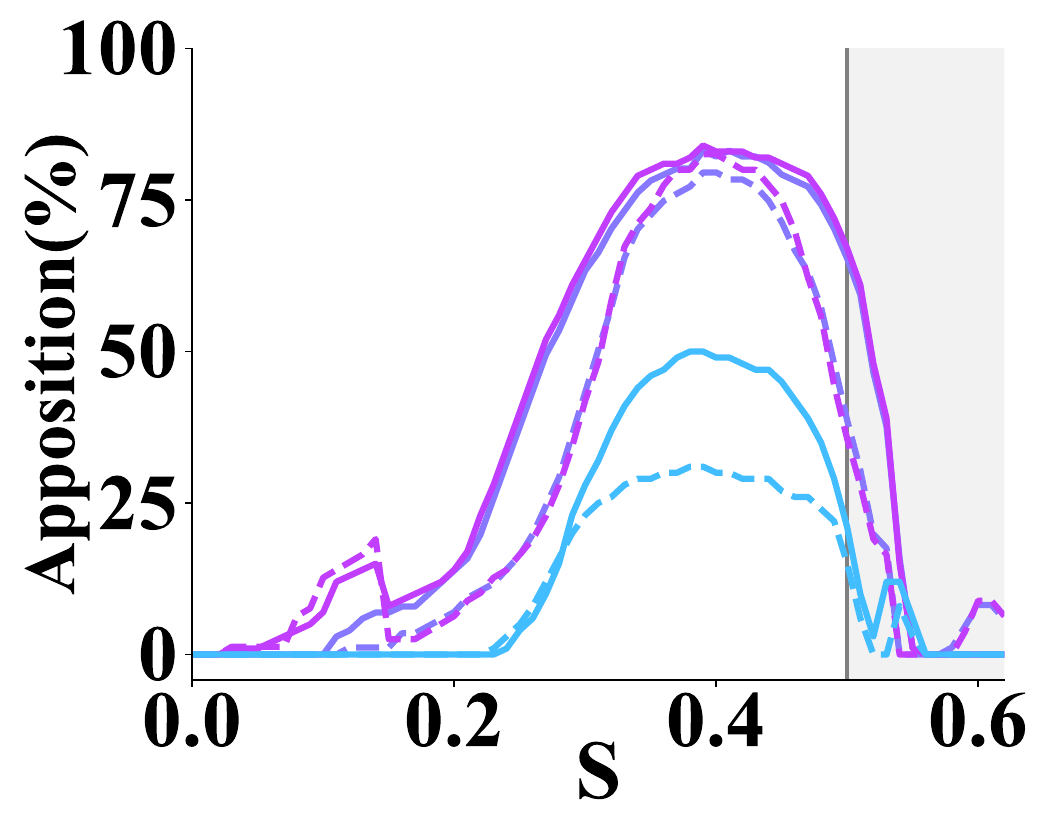}}

\vspace{-.05cm}

\hspace{-1cm}\includegraphics[width=.21\textwidth]{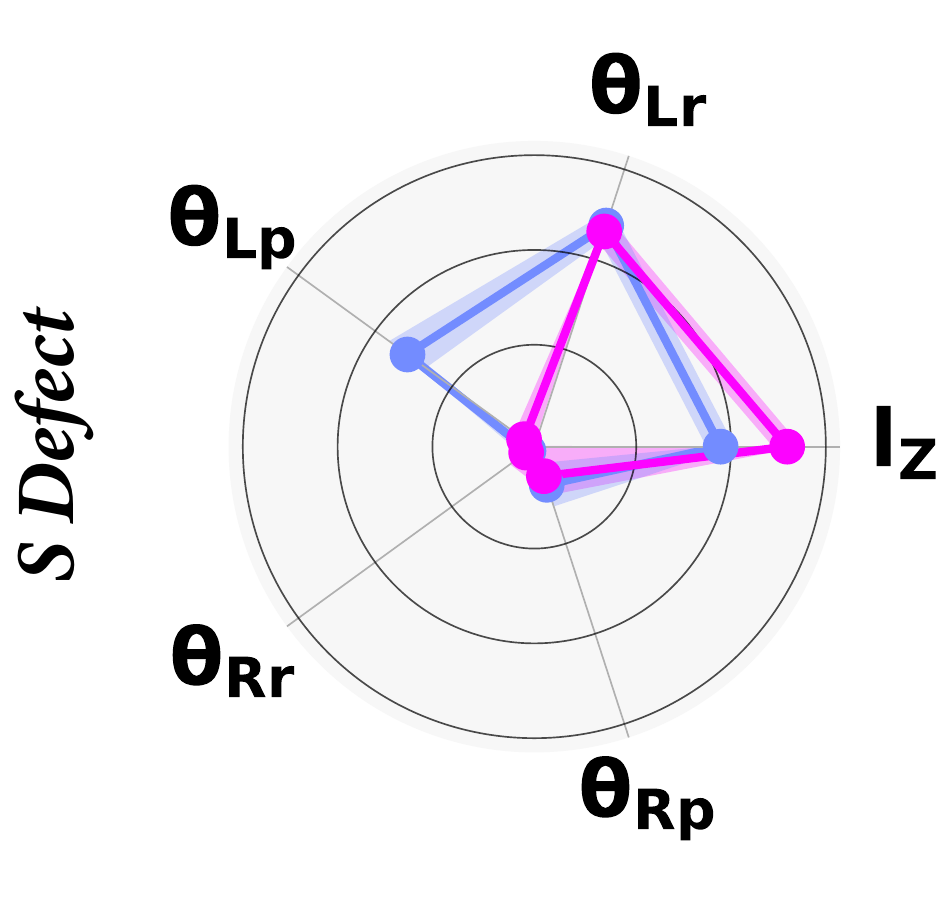}
\hspace{.6cm}
\includegraphics[width=.23\textwidth]{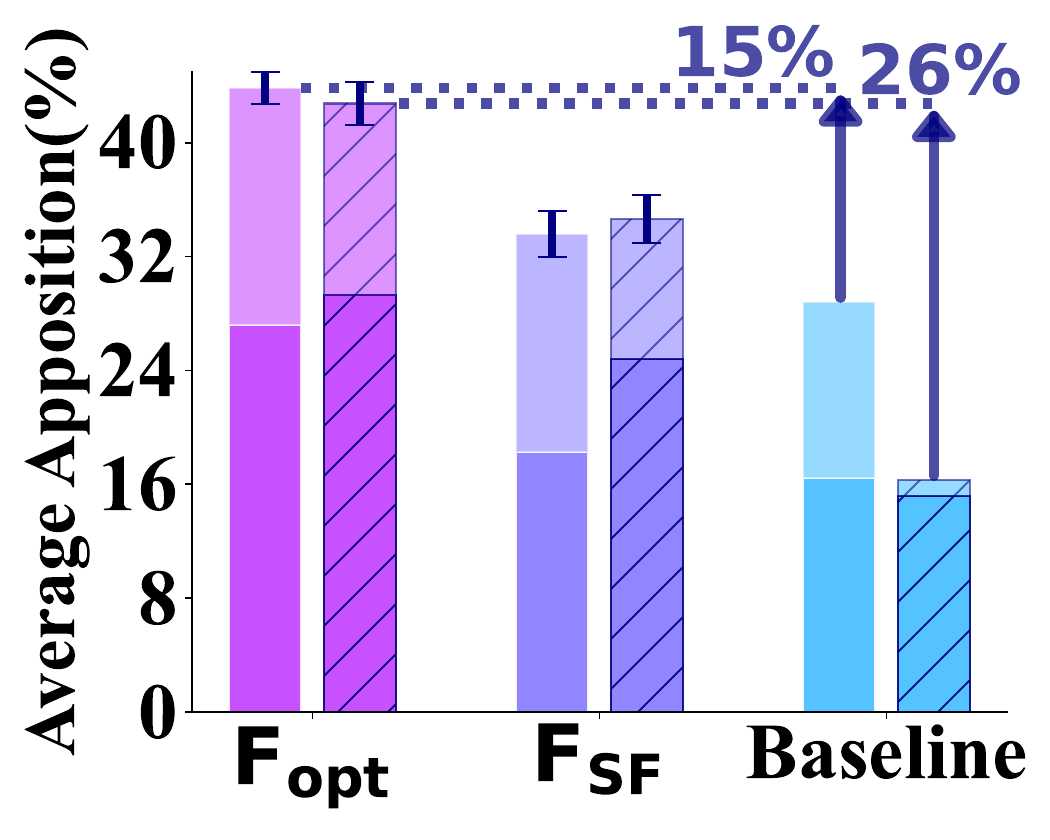}
\hspace{.7cm}
\raisebox{-1mm}{\includegraphics[width=.23\textwidth]{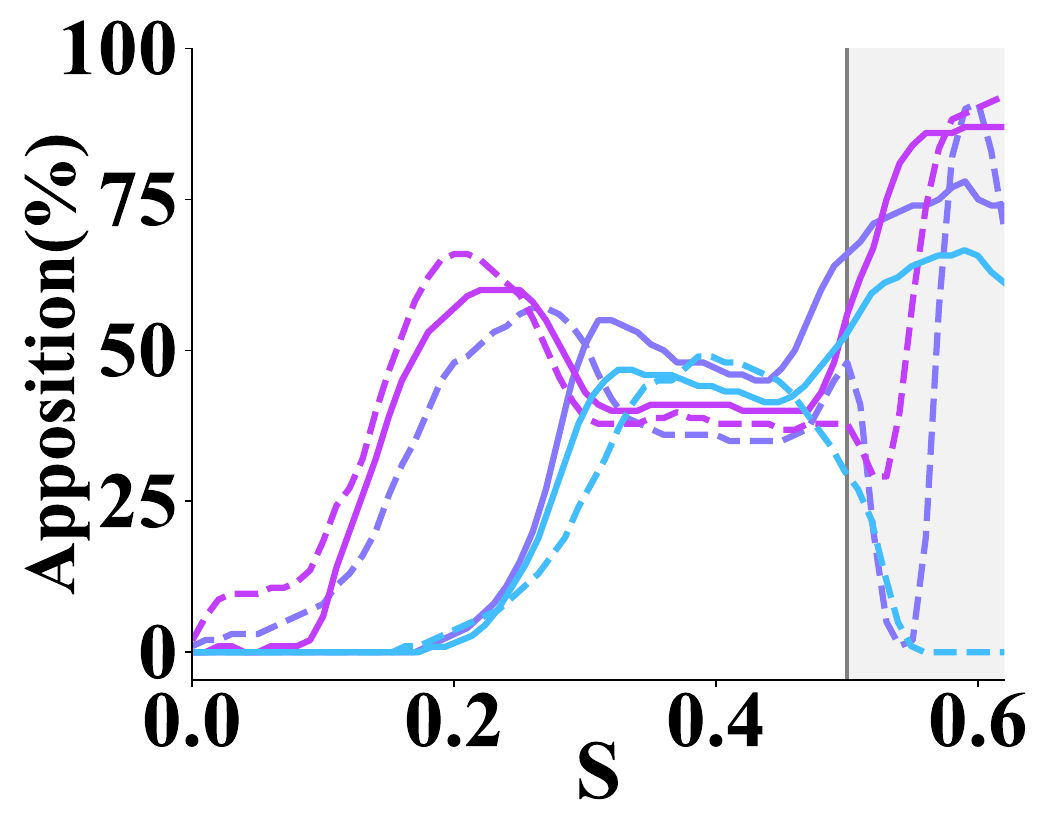}}

\vspace{-.05cm}

\hspace{-1cm}\includegraphics[width=.22\textwidth]{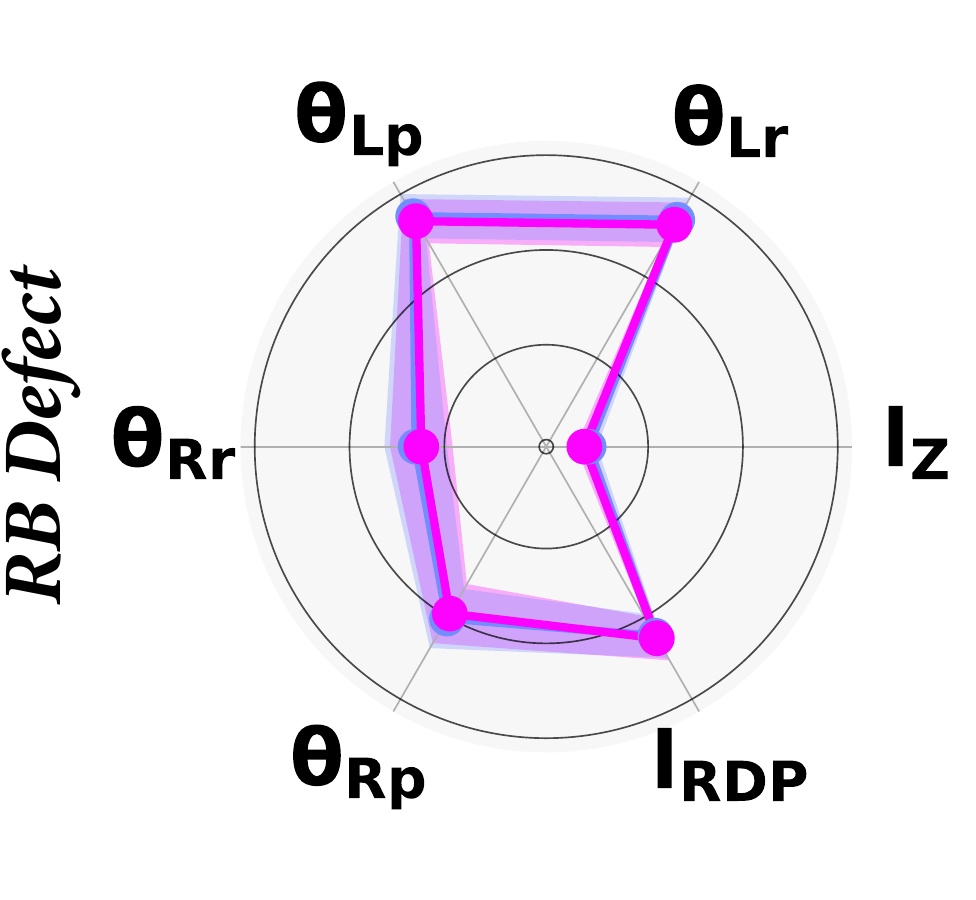}
\hspace{.6cm}
\includegraphics[width=.23\textwidth]{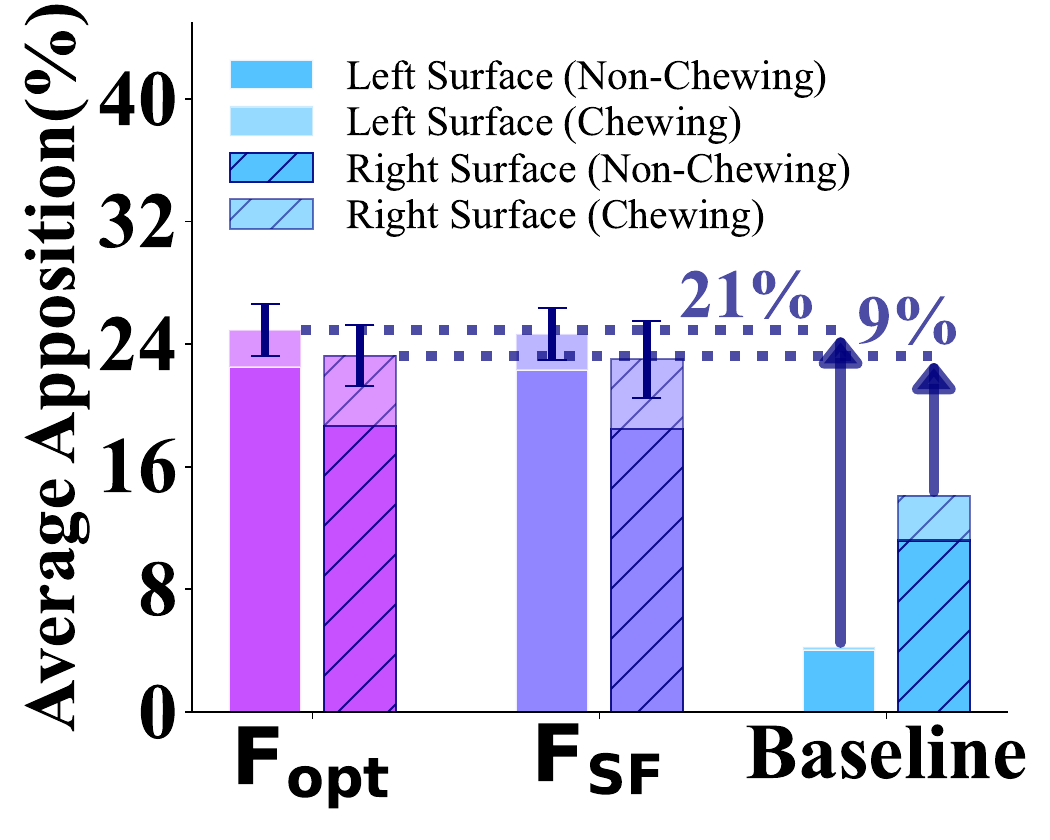}
\hspace{.7cm}
\raisebox{-1mm}{\includegraphics[width=.23\textwidth]{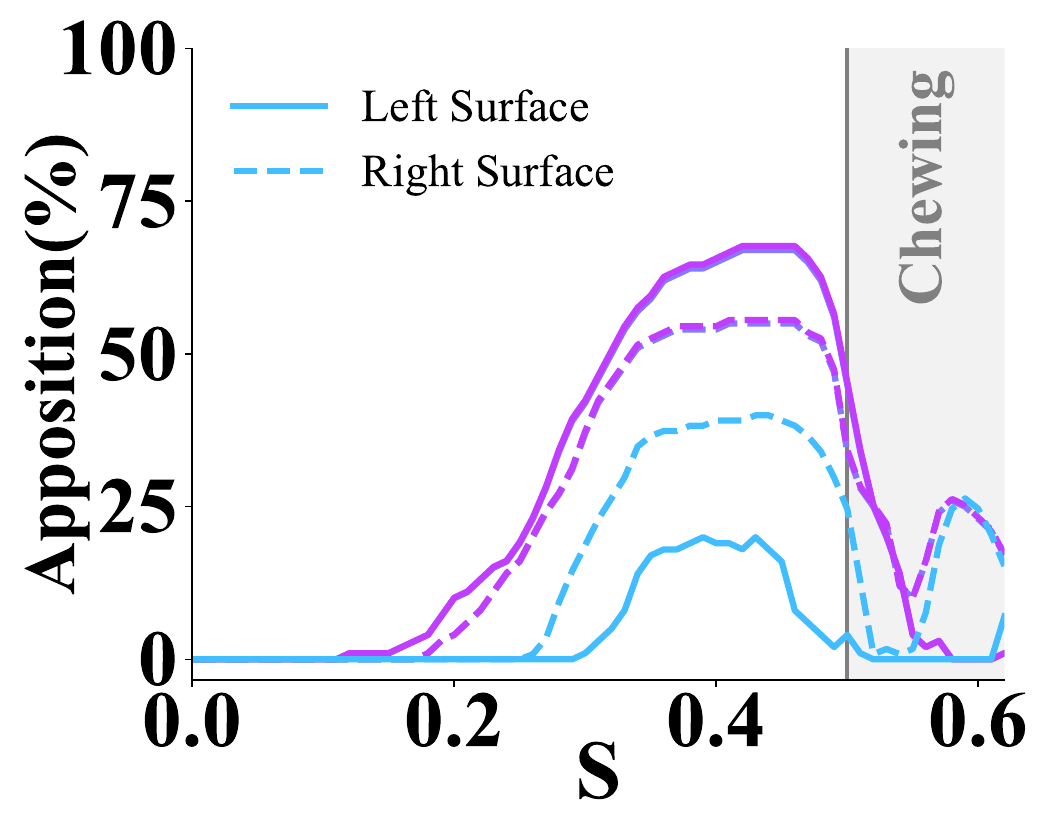}}

\caption{\textbf{Optimization outcomes for the patient-specific cases.} Rows show the patient-specific B, S, and RB cases; columns show optimized design parameters mapped to \([-100,100]\), mean donor-mandible apposition over one chewing cycle, and the corresponding apposition trajectory. The gray interval marks bolus engagement. Results are shown for \textcolor{myhexcolor3}{baseline} reconstructions, defined by the surgeon-implemented day-5 post-op CT configuration with zero design-variable offsets, and for reconstructions optimized using \textcolor{myhexcolor1}{\(F_{\mathrm{opt}}\)} and \textcolor{myhexcolor2}{\(F_{\mathrm{SF}}\)}. All optimized values are averaged over five optimization trials with different random seeds. Percentage differences are reported as absolute values.}
\label{fig:fig_param_patient}

\vspace{7mm}
\includegraphics[width=1\textwidth]{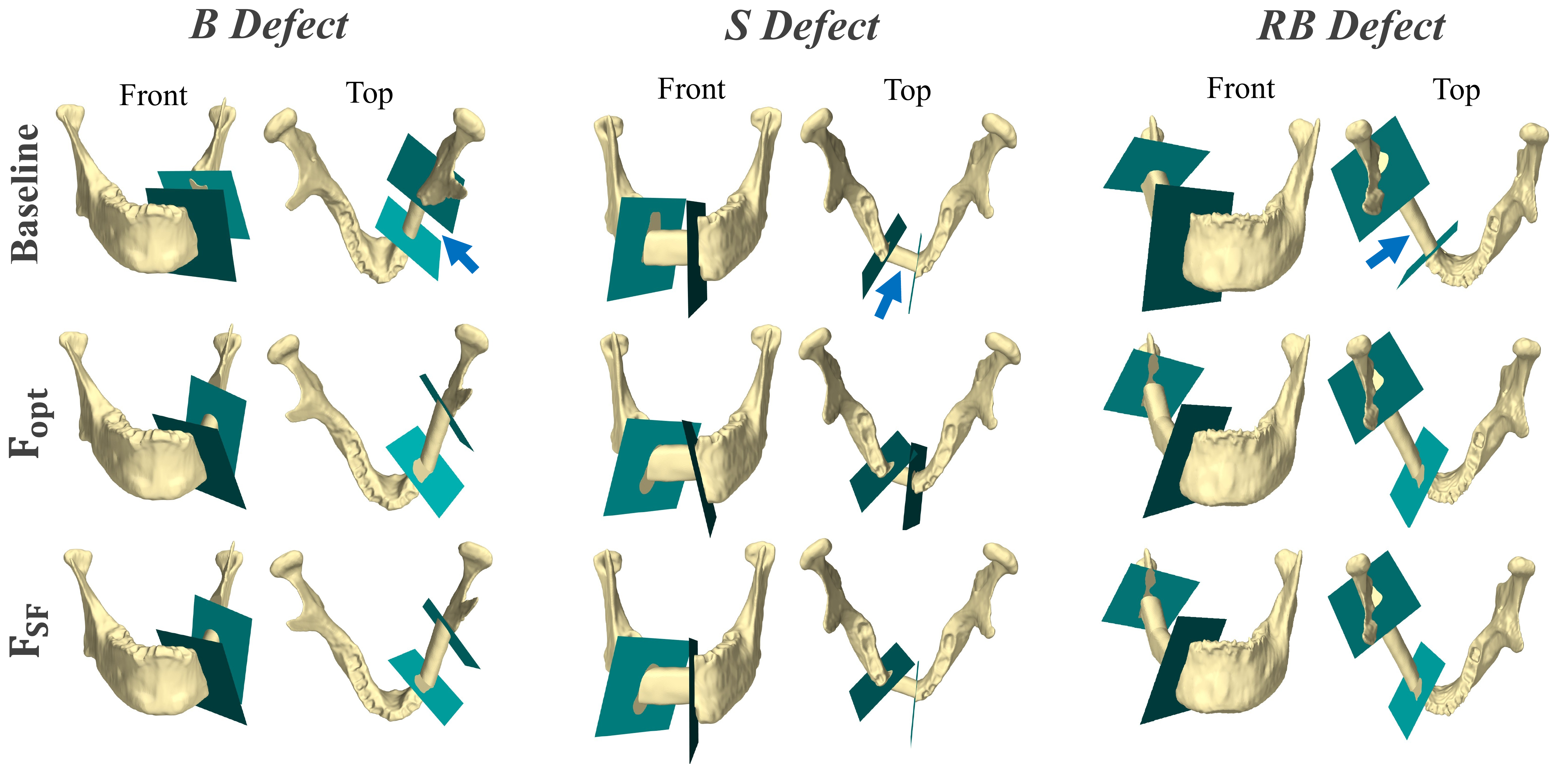}

\caption{\textbf{Patient-specific reconstruction configurations.} Baseline reconstructions, corresponding to the surgeon-implemented configuration recovered from day-5 post-op CT, and the optimized \(F_{\mathrm{opt}}\) and \(F_{\mathrm{SF}}\) reconstructions are shown for the patient-specific B, S, and RB cases from front and top views; the blue arrow in the baseline row marks the resection interface.}
\label{fig:patient_recon}

\end{figure*}

Applying the same optimization workflow to CT-derived patient-specific digital twins produced results consistent with the generic-model trends in these three cases, while exposing the case-to-case variability that motivates patient-specific planning in the first place. Figure~\ref{fig:fig_param_patient} reports the optimized parameters, mean apposition, and trajectories for the three patient cases (B, S, and RB defects), again averaged over five independent optimization trials with different random seeds, and Figure~\ref{fig:patient_recon} shows the corresponding reconstructions. Compared with the surgeon-implemented day-5 post-op baseline, $F_{\mathrm{opt}}$ increased mean apposition by approximately $18$--$21$~percentage points on the two interfaces of the patient B defect, $15$--$26$~percentage points on the patient S defect, and $9$--$21$~percentage points on the patient RB defect, with $F_{\mathrm{SF}}$ producing slightly smaller but qualitatively consistent gains. These values are of the same order as the corresponding generic-model improvements, suggesting that in these three cases the workflow's relative benefit carried over when the underlying anatomy, donor properties, and musculoligamentous parameters were personalized through the pipeline of Algorithm~\ref{alg:psm_pipeline}.

The patient-specific apposition trajectories in the third column of Figure~\ref{fig:fig_param_patient} mirror the generic-model behavior within each defect class: gains concentrate in the pre-bolus phase for the B case, while the S case retains the dual strategy in which $F_{\mathrm{opt}}$ also raises bolus-phase apposition and $F_{\mathrm{SF}}$ instead lifts the lateral, pre-bolus phase for a flatter and safer profile. The consistency of this behavior between generic and patient-specific runs of the same defect type reinforces the warm-start observation discussed below.

The patient-specific radar charts in Figure~\ref{fig:fig_param_patient} also show that $F_{\mathrm{opt}}$ and $F_{\mathrm{SF}}$ lie close to one another for a given case. Their values in physical units are likewise listed in Appendix Table~\ref{tab:optimized_values}. A further observation is that, within the same defect class, the patient-specific optimum is broadly similar to the corresponding generic optimum: the dominant pattern of cut-plane angles and donor offsets is largely set by the defect region, while patient anatomy modulates the precise location within that region. This has a useful practical implication. The generic optima can be used as warm-start initializations for the patient-specific Bayesian optimization, allowing the search to begin in an already promising region of the design space and significantly reducing the number of expensive simulations required for patient-specific planning. Across cases, however, the design vectors still differ enough that a single set of generic angular and length recommendations would either over- or under-rotate the donor segments depending on subject anatomy; the proposed approach therefore complements rather than replaces case-by-case adaptation. The improvement of $F_{\mathrm{opt}}$ over the surgeon-implemented baseline in all three cases is particularly noteworthy because the baseline is not arbitrary; it is the configuration actually delivered in the operating room and recovered from day-5 post-op CT. Even against this clinically realized reference, the optimizer finds configurations within the feasibility bounds (Table~\ref{tab:design_bounds}) that are predicted to be more favorable for union, suggesting a genuine planning headroom that current visually driven VSP workflows do not exploit. Earlier clinical reports have documented that even with VSP, mandibular reconstruction outcomes are heterogeneous and nonunion rates remain non-negligible~\citep{swendseid2020natural,knitschke2022osseous}, and the present results suggest that biomechanically informed search over the same surgically controllable variables can complement rather than replace the geometric VSP toolchain~\citep{vyas2022virtual}. To distinguish optimization from random perturbation, Table~\ref{tab:sobol_reference} reports the mean \(F_{\mathrm{opt}}\) over the 25 Sobol initialization points as a random-feasible reference; it is comparable to or below the surgeon baseline for the B and S defects and only modestly above for RB, and about three times smaller than the optimized value, confirming that the optimizer, not arbitrary perturbation, drives the gains. Different surgeons may also plan a given defect differently, though they share this geometric basis; the limited high-quality region in Fig.~\ref{fig_par} indicates that most feasible plans fall short of the optimum, but inter-surgeon variability may still contribute, and a more extensive multi-surgeon evaluation would be needed to fully separate it from the optimization gain.

A further point of interpretation concerns the $F_{\mathrm{opt}}$/$F_{\mathrm{SF}}$ trade-off. Across all six cases, $F_{\mathrm{SF}}$ produced apposition values that were consistently within a few percent of $F_{\mathrm{opt}}$ while the underlying configurations differed in their worst-case principal stresses on cortical and cancellous bone. Importantly, in every $F_{\mathrm{SF}}$ run the worst-case safety factor remained above unity at every time point of the chewing cycle on both interface sides, confirming that the safety-factor regularization defined in Eq.~\ref{eq:structopt_penalty} achieves its intended effect of keeping the local maximum principal stress below the bone yield limit throughout the simulation. This shows that the safety-factor regularization is not in zero-sum competition with apposition: in the present feasible region there exist configurations that are both nearly optimal in apposition and free of locally unfavorable loading, in line with reported bone yield-stress limits~\citep{morgan2018bone}. Conversely, the small absolute differences between $F_{\mathrm{opt}}$ and $F_{\mathrm{SF}}$ caution against interpreting either as a single ground-truth recommendation in isolation; in practice they should be presented together so that surgeons can weigh maximal predicted apposition against margin to bone yield.

\begin{figure*}[tp]
    \centering
        \hspace{-.4cm}
    \begin{subfigure}{0.49\textwidth}
        \centering
        \includegraphics[width=\linewidth]{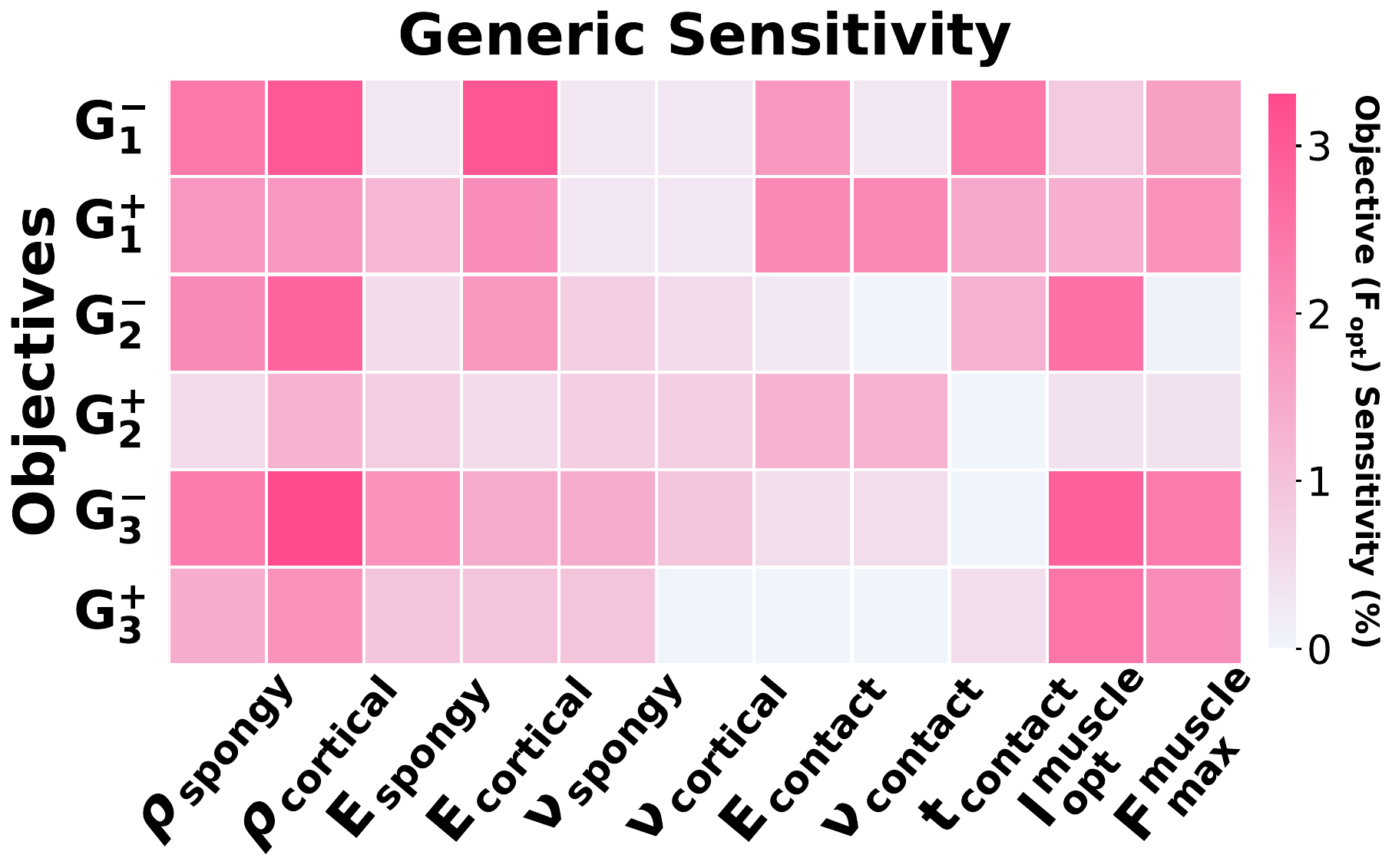}
        \caption*{(a)}
        \label{fig:sensitivity_generic}
    \end{subfigure}
    \hspace{.4cm}
    \begin{subfigure}{0.49\textwidth}
        \centering
        \includegraphics[width=\linewidth]{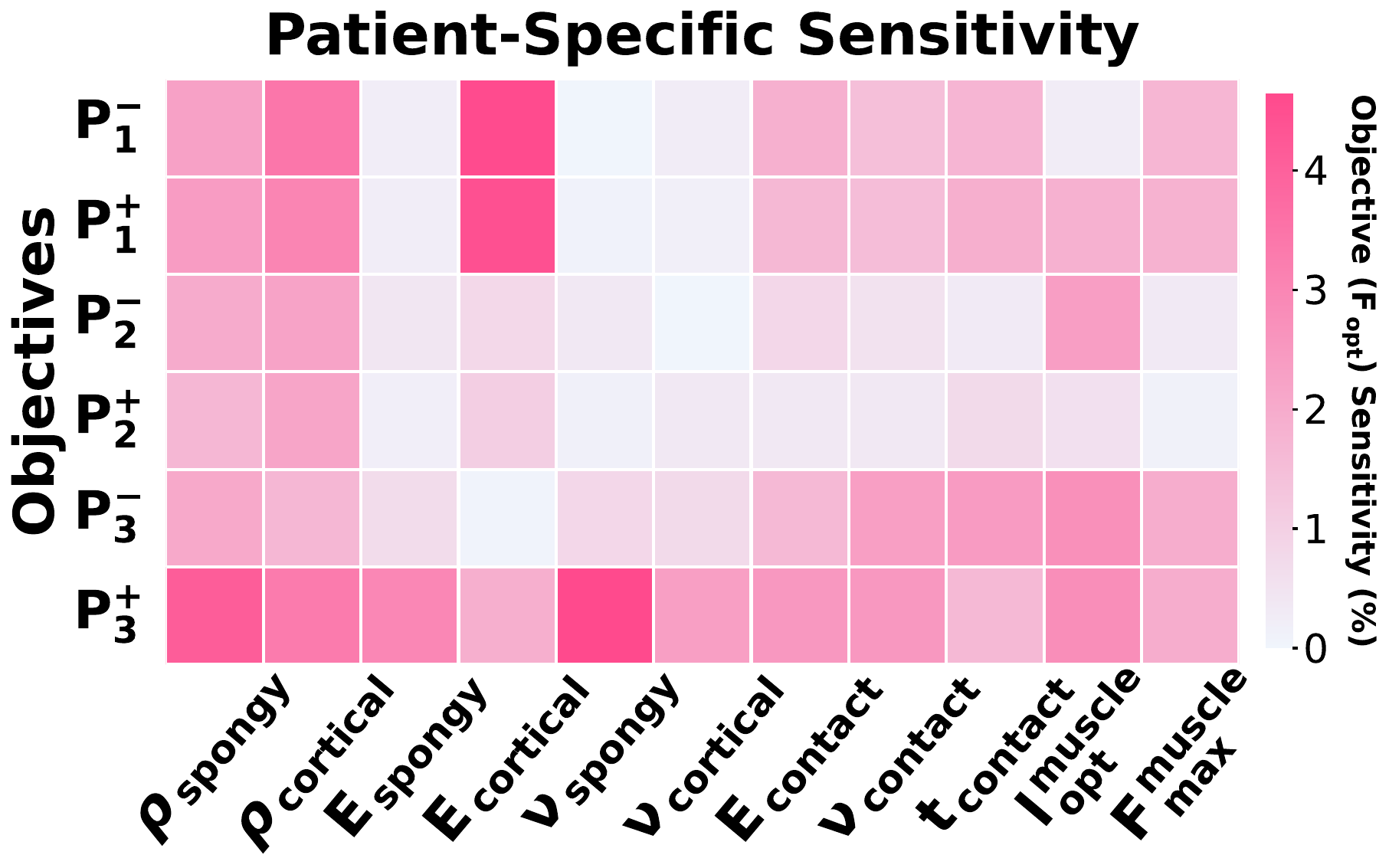}
        \caption*{(b)}
        \label{fig:sensitivity_patient}
    \end{subfigure}

\hspace{7mm}
    \begin{subfigure}{0.95\textwidth}
        \centering
        \includegraphics[width=\linewidth]{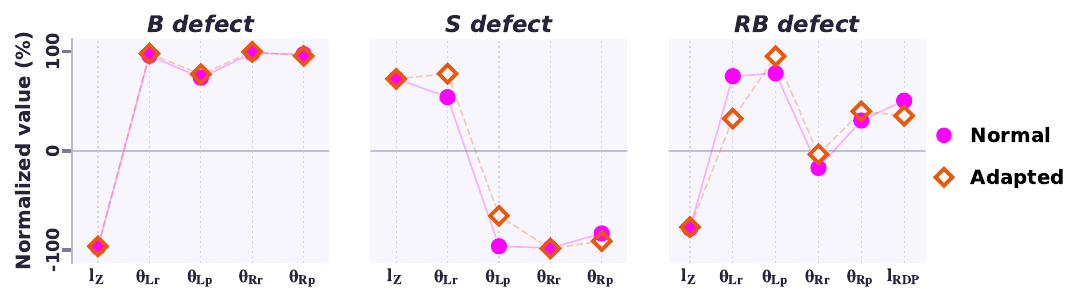}
        \caption*{(c)}
    \end{subfigure}
    \caption{\textbf{Sensitivity to model parameters and chewing activation.} Average effect, over five runs, of \(-10\%\) and \(+10\%\) perturbations in 11 modeling parameters on the \(F_{\mathrm{opt}}\) objective for (a) the generic B, S, and RB defect models and (b) the patient-specific models with similar defect types. G denotes the generic model, P denotes the patient-specific model, and superscripts \(-\) and \(+\) indicate \(-10\%\) and \(+10\%\) perturbations, respectively. Across these perturbations, the change in \(F_{\mathrm{opt}}\) was capped at \(\sim\)\(3\%\) for the generic models and \(\sim\)\(4\%\) for the patient-specific models. (c) Robustness to the chewing-activation assumption: optimized design variables for the B, S, and RB defects on the normalized \([-100,100]\) scale under the normal early-stage activation (\(F_{\mathrm{opt}}\)) and the adapted activation from inverse simulation (Appendix~\ref{app:forward_inverse}); the adapted optimum stays close to the normal optimum and preserves the sign and pattern of every variable.}
    \label{fig:sensitivity_analysis}
\end{figure*}

\begin{table}[t]
\centering
\caption{Comparison of the surgeon baseline, the random-feasible reference, and the optimized \(F_{\mathrm{opt}}\) (Eq.~\ref{eq:structopt_fopt}) for the patient-specific cases. The random-feasible reference is the mean over the 25 Sobol initialization points, and the optimized value is averaged over five trials. All values are percentages, reported as mean \(\pm\) SD.}
\label{tab:sobol_reference}
\footnotesize
\setlength{\tabcolsep}{6pt}
\renewcommand{\arraystretch}{1.2}
\begin{tabular*}{\columnwidth}{@{\extracolsep{\fill}}lcccc}
\toprule
Case & Defect & Baseline & Random feasible & Optimized \\
\midrule
Patient 1 & B  & 10.3 & $9.5 \pm 6.8$  & $28.6 \pm 0.7$ \\
Patient 2 & S  & 16.5 & $14.6 \pm 7.4$ & $42.7 \pm 0.9$ \\
Patient 3 & RB & 3.2  & $7.5 \pm 5.3$  & $22.6 \pm 1.3$ \\
\bottomrule
\end{tabular*}
\end{table}

\subsection{Sensitivity Analysis}\label{subsec:results_sensitivity}
The robustness of the optimization to modeling uncertainty was examined by perturbing eleven dominant parameters by $\pm10\%$ around their nominal values and reevaluating $F_{\mathrm{opt}}$ for all three generic and all three patient-specific cases. The averaged outcomes are summarized in Figure~\ref{fig:sensitivity_analysis}(a,b). Across all $660$ simulations, the change in $F_{\mathrm{opt}}$ was capped at $\sim$$3\%$ for the generic models and $\sim$$4\%$ for the patient-specific models, indicating that the recovered optima are not artifacts of a particular parameter setting but persist within a reasonable neighborhood of the calibrated values.

Within this small overall range, the relative ordering of sensitivities is informative. The largest contributions came from the cortical bone parameters $\rho_{\mathrm{cortical}}$ and $E_{\mathrm{cortical}}$ (Eqs.~\ref{eq:structopt_density}--\ref{eq:structopt_sed}) and from the muscle parameters $\ell_{\mathrm{opt}}$ and $F_{\max}$~\citep{hill1953mechanics}. The cancellous density $\rho_{\mathrm{cancellous}}$ also produced visible sensitivity in the generic models, although smaller than its cortical counterpart. This ordering is biomechanically plausible: cortical bone supports the dominant load path through the donor at the resection interface, and the muscle parameters scale the magnitude of the chewing-cycle stimulus, both of which directly modulate the strain-energy-density signal that drives the union objective~\citep{field2010prediction}.

The patient-specific sensitivities are slightly larger than the generic ones, and the most sensitive parameter shifts toward $E_{\mathrm{cortical}}$ in some cases. Even in this less constrained regime, the absolute change in $F_{\mathrm{opt}}$ remains a small fraction of the optimization-induced improvement reported in Sections~\ref{subsec:results_generic}--\ref{subsec:results_patient}, suggesting that the recovered configurations are robust to the kind of parameter uncertainty that routinely arises from CT-based calibration. We restricted the sensitivity analysis to $F_{\mathrm{opt}}$ because $F_{\mathrm{SF}}$ adds a worst-case safety-factor penalty to the same apposition-driven signal in Eq.~\ref{eq:structopt_penalty} and is therefore expected to share the dominant trends; this is a deliberate scoping choice rather than a limitation of the analysis.

Robustness to the chewing-activation assumption was assessed by repeating the optimization under an adapted activation profile recovered by inverse simulation (Appendix~\ref{app:forward_inverse}), in which the muscles remaining after resection track the pre-reconstruction jaw trajectory. The optimized design variables recovered under this adapted profile stayed close to the normal early-stage optimum and preserved the sign and pattern of every cut-plane angle and donor offset (Fig.~\ref{fig:sensitivity_analysis}(c)). On the normalized \([-100,100]\) scale the two optima agreed to within a mean of about 2\% for the B defect, the donor offsets changing by at most about 1~mm; for the S and RB defects the optima remained close, with mean agreement of about 13\% and 17\% and every variable retaining its sign, the largest single difference in each falling on one angular variable. The optimizer therefore recovers the same reconstruction strategy under both loading profiles, supporting the robustness of the optimized configuration to the activation assumption.

\subsection{Longitudinal Spatial Analysis and TMJ-Disc Approximation}\label{subsec:results_validation}

\begin{figure*}[t]
    \centering
    \begin{subfigure}{.6\textwidth}
        \centering
        \includegraphics[width=\linewidth]{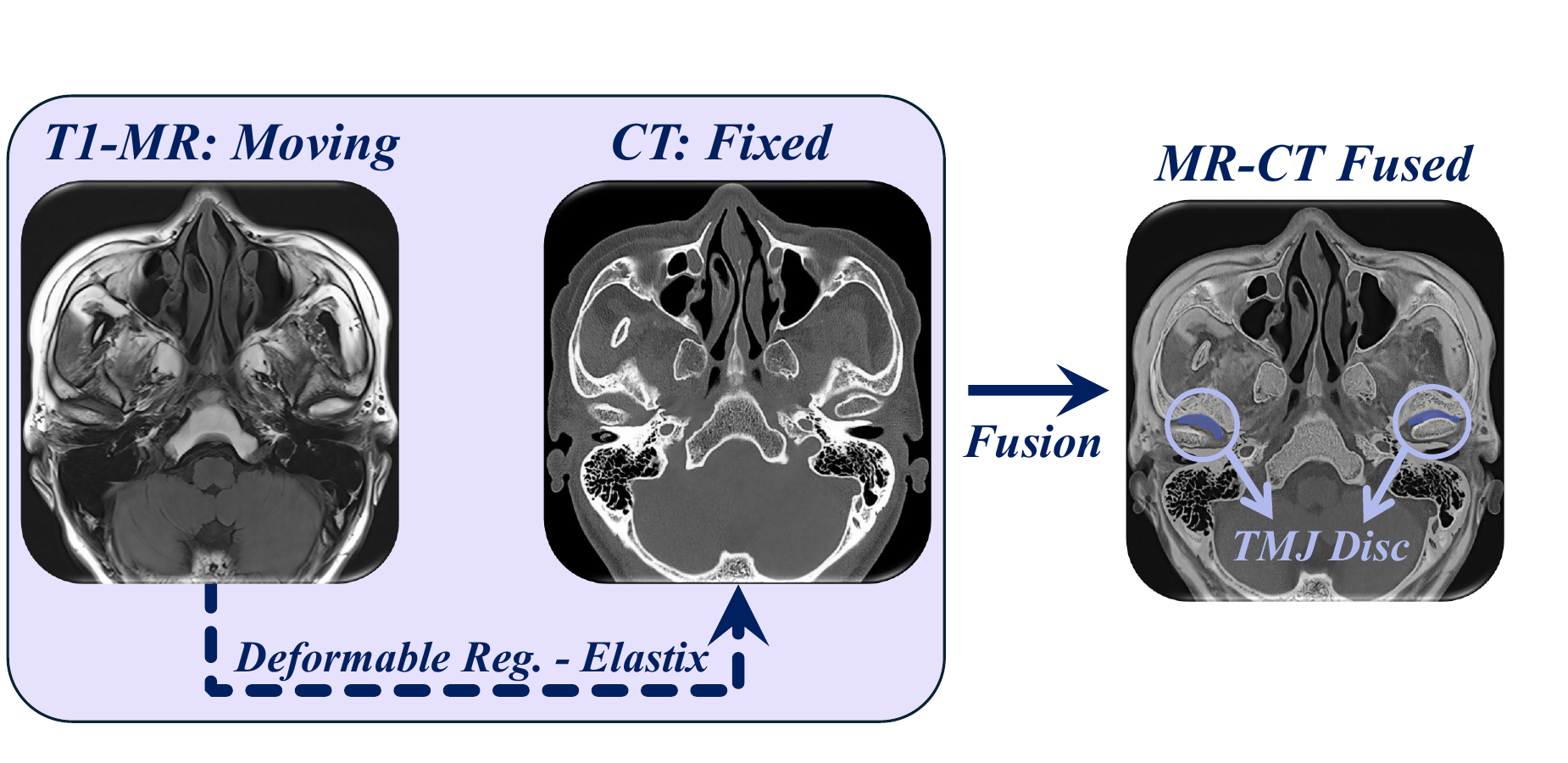}
        \caption*{(a)}
    \end{subfigure}
    \begin{subfigure}{0.39\textwidth}
        \centering
        \includegraphics[width=\linewidth]{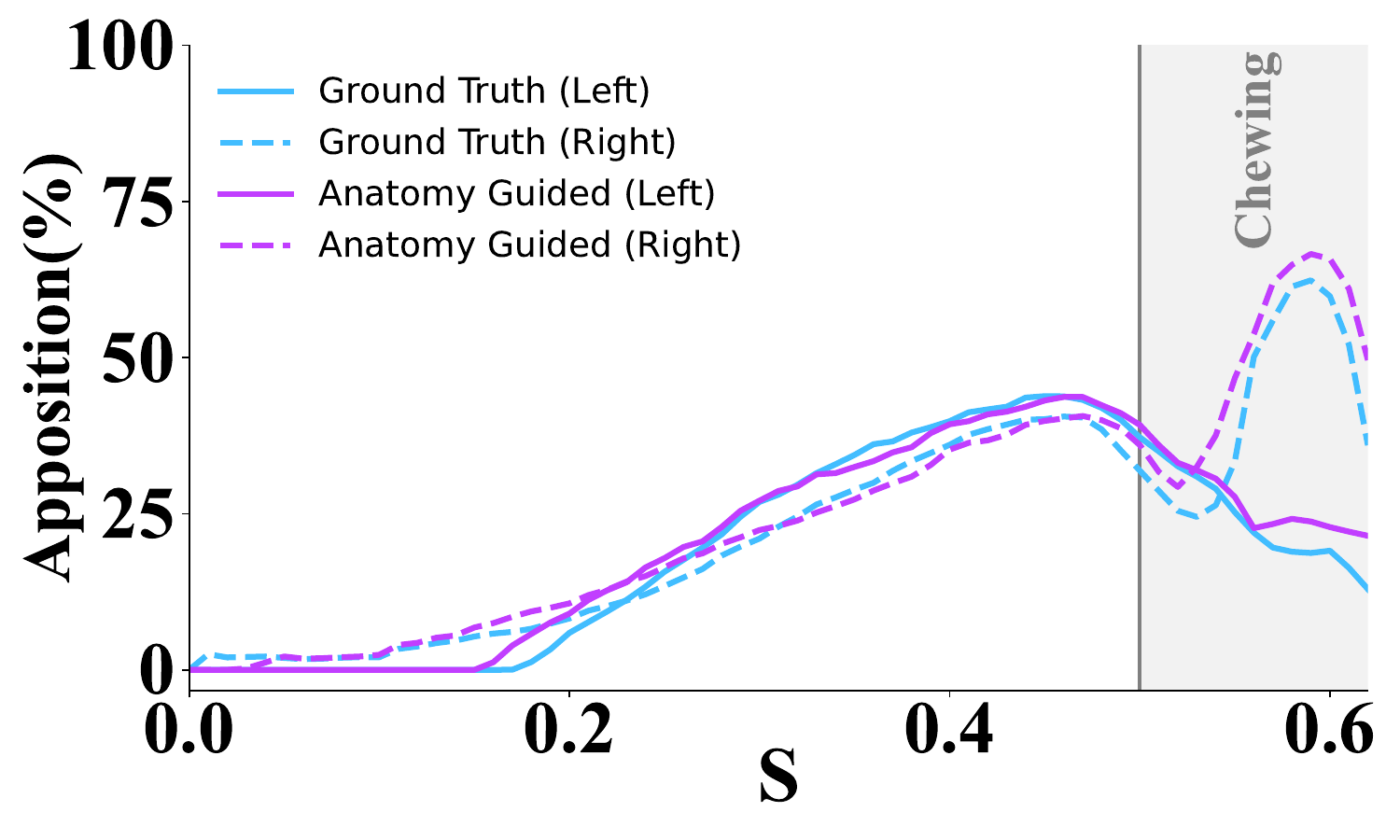}
        \caption*{(b)}
    \end{subfigure}
    \caption{\textbf{TMJ-disc sensitivity analysis.} (a) T1-weighted MRI was registered to the CT image using Elastix, and the fused image was used to guide temporomandibular joint disc segmentation. (b) Donor-mandible apposition trajectories over one chewing cycle are compared between the real segmented disc and the anatomy-guided disc approximation used in the patient-specific model. The gray interval marks bolus engagement. The two trajectories differ by an RMS of 2.4\% and 3.9\% at the left and right interfaces.}
    \label{fig_TMJ}
\end{figure*}

Here, the four longitudinal cases are used for retrospective CT-based spatial comparison of model-predicted apposition with year-1 bone formation, while the MRI-based TMJ-disc comparison is limited to Patient~4, the only case with postoperative MRI. Figure~\ref{fig_TMJ}(a) shows the Elastix-based MRI-to-CT registration that was used to obtain a real segmented disc geometry. Figure~\ref{fig_TMJ}(b) compares the resulting donor-mandible apposition trajectories obtained with the MRI-segmented disc and with the anatomy-guided disc approximation produced by the condyle/fossa-blended deformation field of Eq.~\ref{eq:psm_tmj_blend}--Eq.~\ref{eq:psm_tmj_weights}. Across the cycle the two trajectories differ by an RMS of 2.4\% and 3.9\% at the left and right interfaces, with the larger right-side deviation concentrated in the bolus-engagement interval. This implies that the cost-function value used to drive the optimizer is largely insensitive to whether the disc is derived from direct MRI segmentation or from osseous anatomy alone, supporting the use of the anatomy-guided approximation when MRI is unavailable---which is the typical clinical situation for mandibular reconstruction patients~\citep{ahmadi2026computational}. The agreement is also consistent with the design intent of Eq.~\ref{eq:psm_tmj_blend}, where the blended disc field is anchored to the same condyle and fossa surfaces that constrain the disc kinematics in vivo. Since the chewing simulation includes this patient-specific joint, configuration-dependent changes in TMJ loading enter the objective implicitly; the condyle, disc, and capsule retain full six-degree-of-freedom motion rather than constrained or planar kinematics, and condylar motion in our base model was validated previously~\citep{aftabi2024extent}. Explicit TMJ outcomes after reconstruction, such as spatial deviations of the joint~\citep{yang2022spatial}, are not optimized here.

\begin{figure*}[t!]
	\centering
	\captionsetup{skip=3pt}
	\begin{subfigure}{0.49\textwidth}
		\centering
		\includegraphics[width=\linewidth]{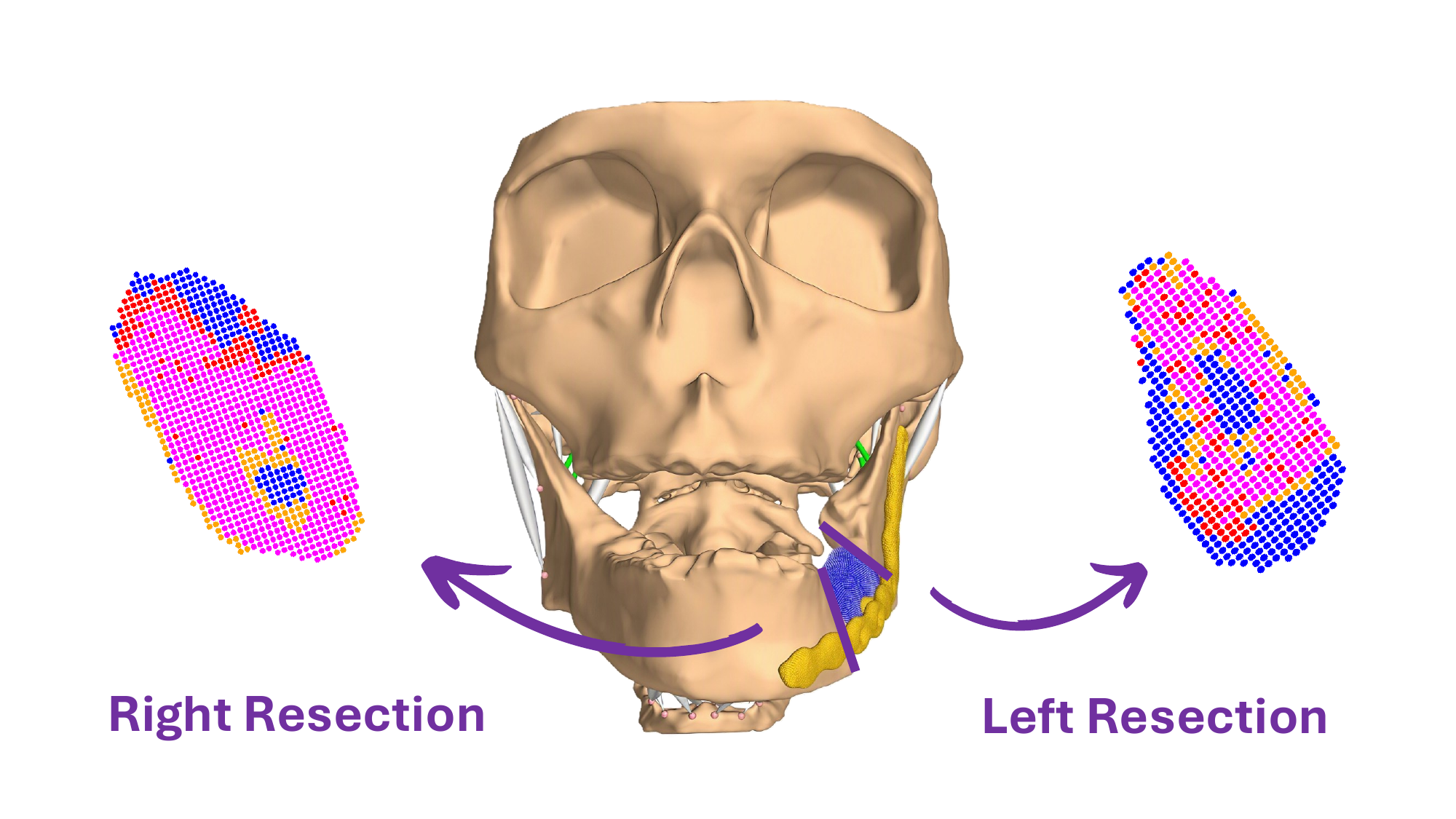}
		\caption*{(a) Patient 1 (B, fibula)}
	\end{subfigure}\hfill
	\begin{subfigure}{0.49\textwidth}
		\centering
		\includegraphics[width=\linewidth]{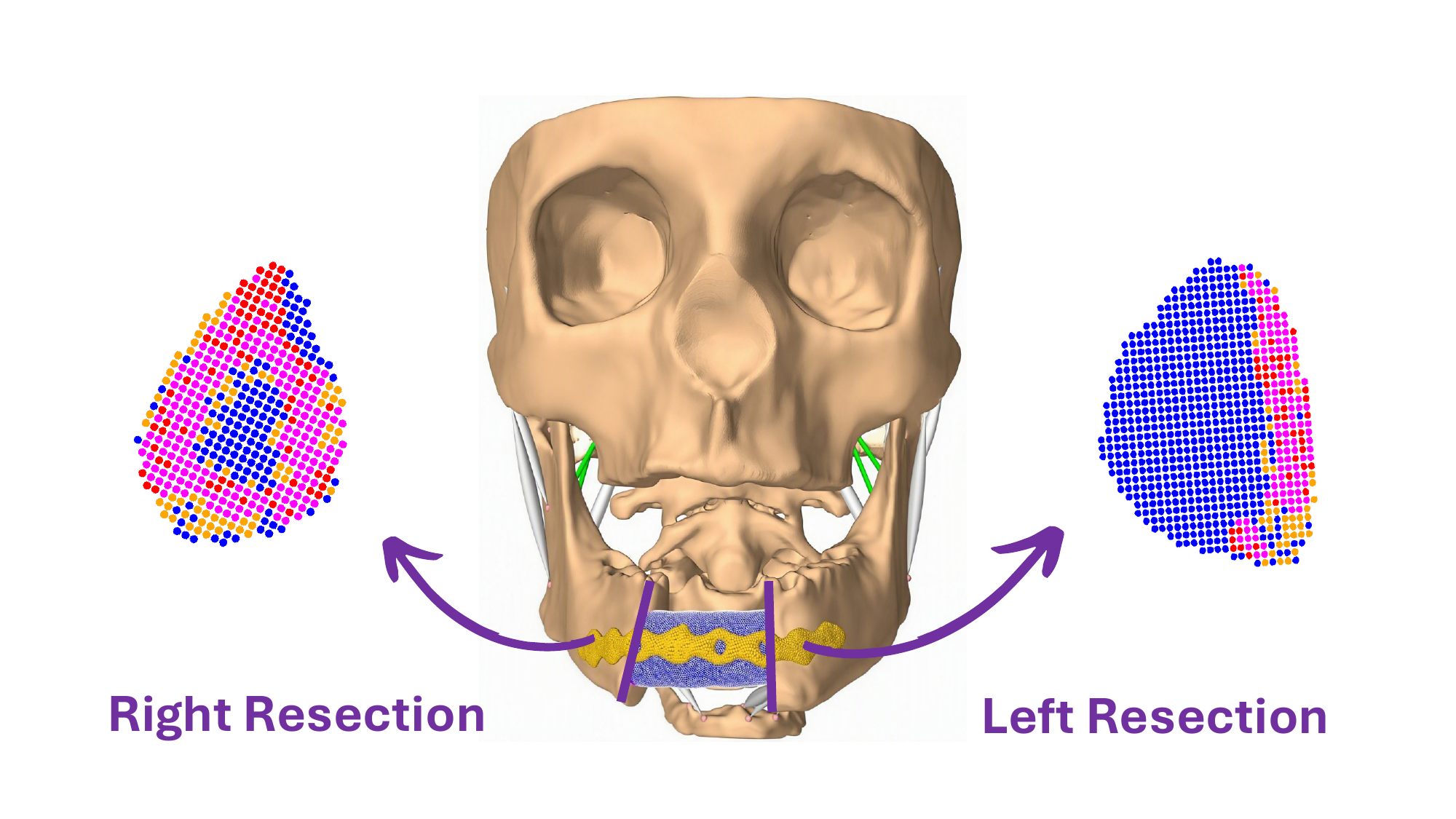}
		\caption*{(b) Patient 2 (S, fibula)}
	\end{subfigure}

	\vspace{-0.4em}
	\begin{subfigure}{0.49\textwidth}
		\centering
		\includegraphics[width=\linewidth]{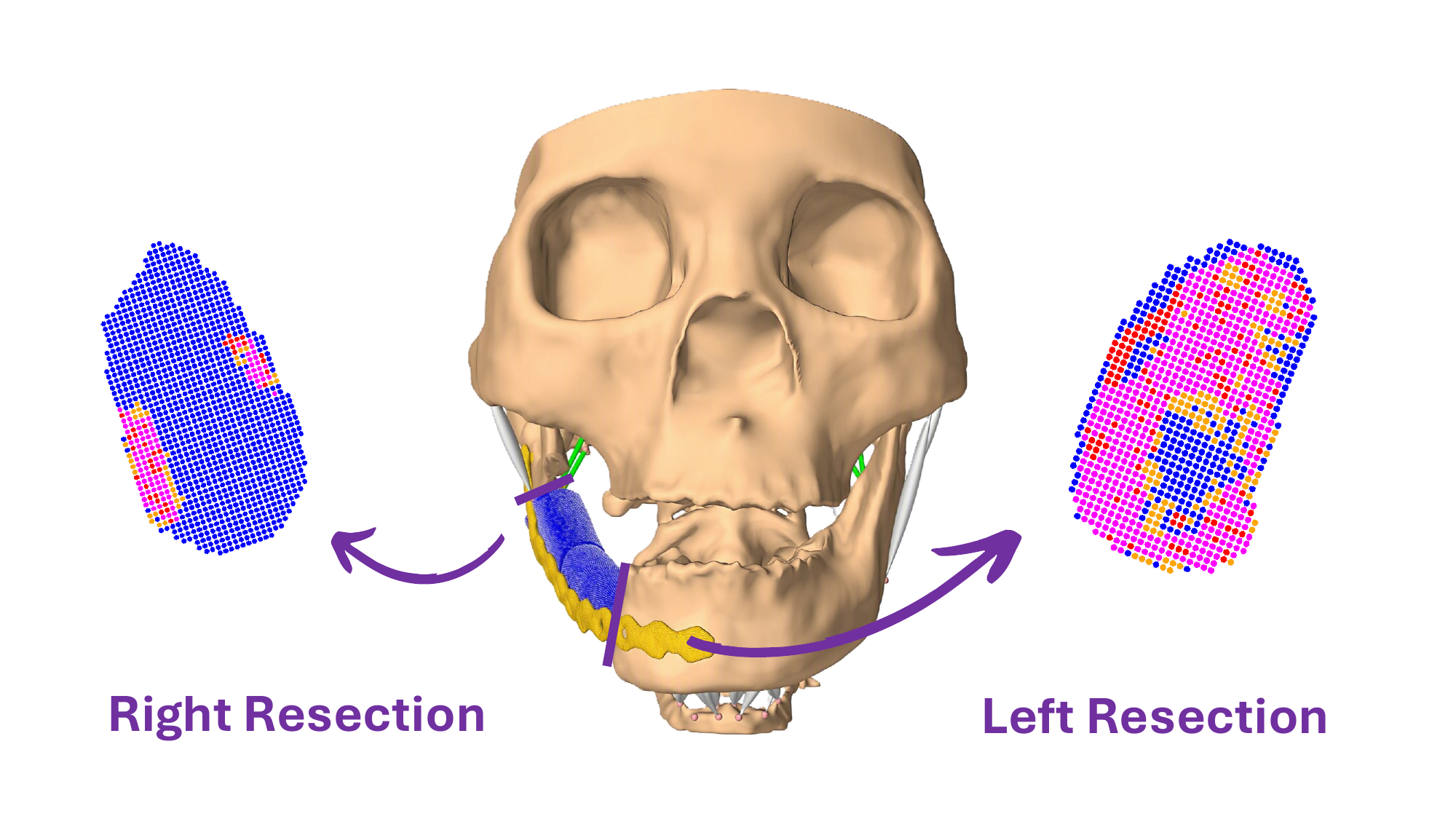}
		\caption*{(c) Patient 3 (RB, fibula)}
	\end{subfigure}\hfill
	\begin{subfigure}{0.49\textwidth}
		\centering
		\includegraphics[width=\linewidth]{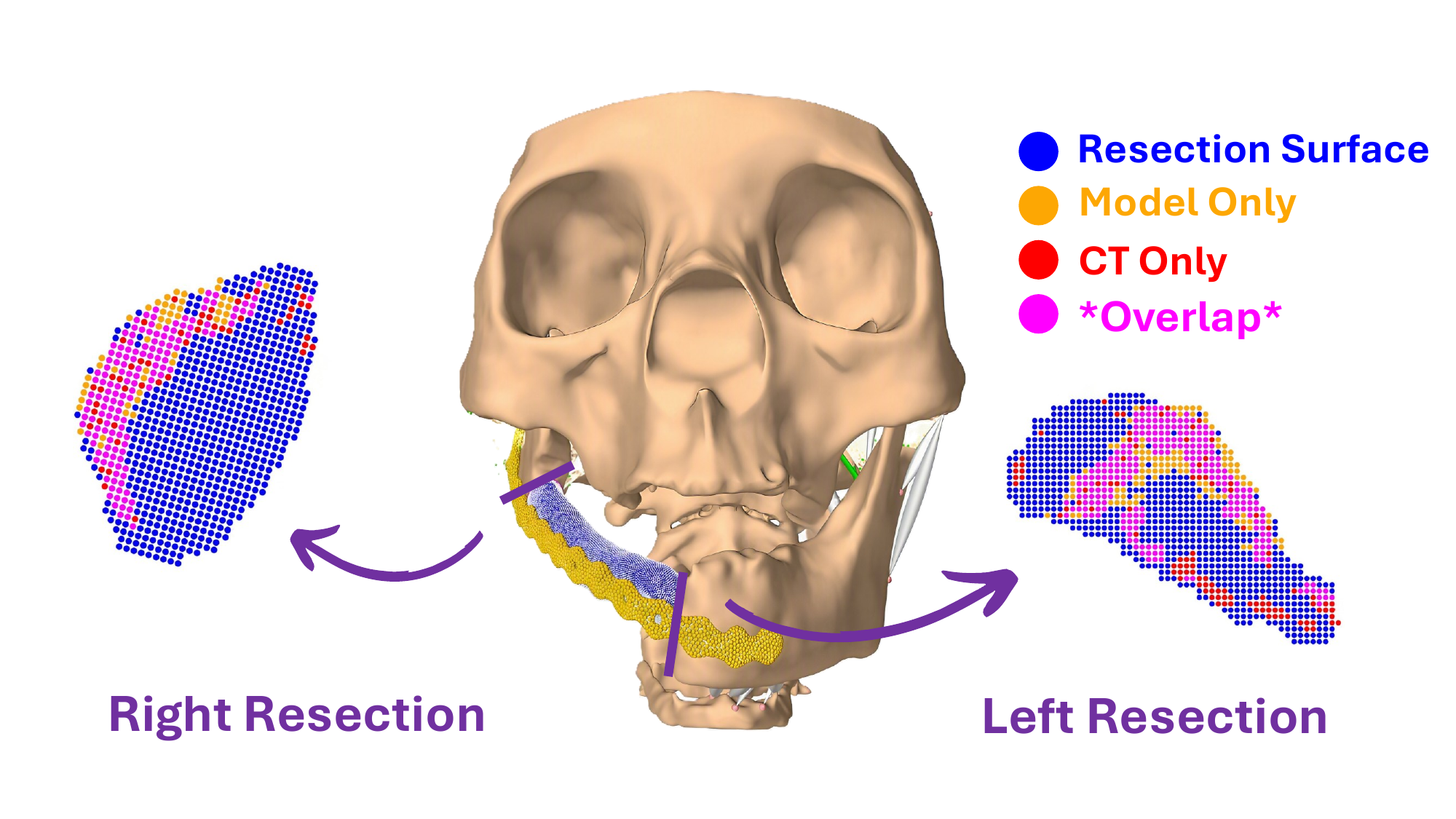}
		\caption*{(d) Patient 4 (RB, scapula)}
	\end{subfigure}
	\caption{\textbf{Longitudinal spatial comparison of model-predicted apposition and year-1 bone formation.} The patient-specific model was constructed from day-5 post-op CT and used to predict donor-mandible apposition near the resection interfaces. On each panel the blue surface is the resected bone surface at the interface, on which the predicted apposition pattern is compared with the bone formation observed on year-1 post-op CT. Panels use the nominal settings (1000~HU threshold and a layer thickness matched to the interface element edge length); the resulting Dice overlap and centroid difference, and their robustness to these two choices, are reported in Table~\ref{tab:val_robustness}. Panel (d) reproduced from \citet{aftabi2025osteoopt}.}
	\label{fig:longitudinal_validation}
	\vspace{0.2em}
	\centering
	\footnotesize
	\setlength{\tabcolsep}{11.5pt}
	\renewcommand{\arraystretch}{1.2}
	\providecommand{\ms}[2]{#1\kern0.12em{\scriptsize$\pm$\kern0.1em#2}}
	\captionof{table}{Longitudinal spatial analysis and its robustness to two analysis choices: the HU Threshold used to segment mature bone and the reconstruction Layer Thickness. For each case, Dice overlap and the centroid difference between the predicted-apposition and observed-bone-formation regions are reported on both resection interfaces. Each entry is the mean $\pm$ SD over three settings of the perturbed parameter: the HU Threshold varied by $\pm 10\%$ about its nominal $1000$~HU (left block), and the Layer Thickness taken as one, two, and three times the interface element edge length (right block).}
	\label{tab:val_robustness}
	\begin{tabular}{@{}l cccc cccc@{}}
	\toprule
	 & \multicolumn{4}{c}{HU Threshold} & \multicolumn{4}{c}{Layer Thickness} \\
	\cmidrule(lr){2-5}\cmidrule(lr){6-9}
	 & \multicolumn{2}{c}{Dice ($\%$)} & \multicolumn{2}{c}{Centroid $\Delta$ (mm)} & \multicolumn{2}{c}{Dice ($\%$)} & \multicolumn{2}{c}{Centroid $\Delta$ (mm)} \\
	\cmidrule(lr){2-3}\cmidrule(lr){4-5}\cmidrule(lr){6-7}\cmidrule(lr){8-9}
	Case & Left & Right & Left & Right & Left & Right & Left & Right \\
	\midrule
	Patient 1 & \ms{72.5}{1.01} & \ms{84.9}{0.59} & \ms{0.64}{0.08} & \ms{1.45}{0.05} & \ms{73.5}{0.80} & \ms{84.5}{0.67} & \ms{0.50}{0.08} & \ms{1.64}{0.04} \\
	Patient 2 & \ms{71.5}{1.32} & \ms{71.7}{1.12} & \ms{1.11}{0.31} & \ms{1.05}{0.15} & \ms{73.0}{1.76} & \ms{71.6}{1.11} & \ms{0.81}{0.28} & \ms{1.82}{0.29} \\
	Patient 3 & \ms{80.1}{2.27} & \ms{72.8}{1.09} & \ms{0.43}{0.07} & \ms{0.28}{0.07} & \ms{82.4}{1.93} & \ms{74.4}{0.22} & \ms{0.24}{0.05} & \ms{0.37}{0.09} \\
	Patient 4 & \ms{70.4}{0.84} & \ms{74.6}{1.35} & \ms{1.22}{0.25} & \ms{0.39}{0.09} & \ms{70.1}{0.82} & \ms{75.2}{1.18} & \ms{1.27}{0.23} & \ms{0.41}{0.10} \\
	\bottomrule
	\end{tabular}
\end{figure*}

The longitudinal comparison itself is summarized in Figure~\ref{fig:longitudinal_validation} and Table~\ref{tab:val_robustness}. Figure~\ref{fig:longitudinal_validation} shows the nominal visual comparison for each of the four longitudinal cases, using the 1000~HU bone-formation threshold and an interface-layer thickness matched to the simulation element edge length. The patient-specific model was constructed from day-5 post-op CT and used to predict donor-mandible apposition near the resection interfaces. The predicted patterns were then compared with the bone-formation distributions observed on year-1 post-op CT, where regions with $\mathrm{HU}>1000$ were operationally identified as cortical bone, a recognized marker of mature bone formation and successful union under sustained interface loading~\citep{mahesh2013essential,zheng2022}. Table~\ref{tab:val_robustness} reports the corresponding Dice overlap and centroid difference for all four cases on both resection interfaces, and summarizes robustness across the HU-threshold and layer-thickness choices described in Section~\ref{subsec:experimental_design}. Across these robustness summaries, Dice overlap ranged from 70.1\% to 84.9\%, and centroid differences ranged from 0.24 to 1.82~mm. This agreement is encouraging because it is consistent with the strain-energy-density formulation adopted in Section~\ref{subsec:opt_workflow} acting as a comparative optimization surrogate in this workflow, and because no year-1 imaging was used during model construction or optimization.

It is important to interpret these results in proper scope. The aim of the analysis was not to re-derive or re-validate the SED-based remodeling formulation, which has already been examined in dedicated bone-remodeling studies~\citep{field2010prediction,zheng2022}, but to assess whether this formulation remains useful as the optimization signal of a patient-specific pipeline that is constrained by routinely available imaging. Equivalently, the analysis does not attempt to forecast the patient-specific remodeling trajectory or absolute clinical union, which would require additional intermediate-time-point imaging to inverse-identify subject-specific remodeling parameters~\citep{zheng2022,zheng2019investigation}; instead, the spatial agreement between predicted apposition and year-1 bone formation supports the cycle-averaged signal as a comparative surrogate for predicted bone-union propensity. The longitudinal comparison remains limited in size because paired day-5/year-1 imaging is rarely available in head-and-neck reconstruction cohorts, and the MRI-based disc reference was available for Patient~4 only; each patient also receives only one realized reconstruction, making counterfactual comparisons within the same subject impossible~\citep{kumar2016mandibular}. Within these constraints, the convergence of three independent indicators--the predicted apposition trajectories, the MRI-versus-anatomy-guided disc comparison, and the year-1 bone-formation overlap--is consistent with the proposed workflow behaving in line with the available longitudinal imaging across the four cases.

\section{Conclusion, Limitations, and Future Directions}\label{sec:conclusion}

We presented \textit{OsteoOpt++}, a patient-specific Bayesian optimization workflow for image-guided mandibular reconstruction planning. Across three generic and three patient-specific defect cases, the optimizer identified configurations that increased cycle-averaged donor-mandible apposition over both common-practice and surgeon-implemented baselines. These gains were stable under parameter perturbations. In the four longitudinal cases, the predicted apposition patterns overlapped with year-1 bone formation. Generic and patient-specific optima also occupied similar regions of the design space within each defect class. Generic optima may therefore provide useful warm-start initializations for patient-specific searches and reduce computational cost.

This study should be interpreted according to its intended scope. It does not seek to re-validate the SED-based bone-remodeling stimulus itself, which has been examined in prior work. Rather, it uses this established stimulus within an image-to-decision platform for a planning problem that remains largely unaddressed: ranking surgically controllable reconstruction configurations by mechanically favorable donor-host interface conditions. This problem differs fundamentally from data-rich medical-image-analysis tasks suited to deep learning. In mandibular reconstruction, each patient receives a single realized surgical configuration. The alternative configurations remain counterfactual and unobserved. Large paired datasets linking alternative plans to biological outcomes are therefore difficult to obtain by construction. Contemporary shape-model, geometry-completion, and learning-based planning methods constitute the geometry-planning step before this biomechanical optimization rather than competing with it~\citep{nakao2017automated,guo2025automated}; \textit{OsteoOpt++} starts from such a geometry-matched plan and tests whether it is also union-favorable under loading, modestly trading geometric overlap for improved interface mechanics.

\paragraph{Limitations.} Several limitations should be considered when interpreting these results. The patient-specific evaluation comprised three optimization cases and four longitudinal cases; Patient~4 was also used for MRI-based TMJ-disc evaluation because it was the only case with postoperative MRI. These data support feasibility assessment and retrospective consistency analysis, while larger cohorts are intrinsically difficult to assemble in this setting, since the analysis requires paired early- and late-postoperative imaging together with a recoverable cut-plane configuration per patient. Such longitudinal acquisitions are seldom obtained as part of routine head-and-neck reconstruction follow-up. Each case also provided a single realized plan; multi-surgeon planning of the same defects would further quantify inter-surgeon variability and is left for future work. The SED/apposition objective is a comparative surrogate for mechanically favorable interface conditions; it does not model vascularization, graft viability, inflammatory response, patient metabolic status, fixation biology, or long-term functional adaptation. Although the workflow itself is not restricted by donor-segment count, defect type, or flap choice, the present evaluation considered three defect classes (B, S, RB) and used a fibula donor for all optimization cases, with the scapula appearing only in Patient~4; broader donor and defect coverage is therefore left for future evaluation. The workflow proposes configurations that are favorable for donor-host bone union but does not prescribe the full reconstruction: plate geometry, screw number and placement, vascular pedicle routing, avoidance of interference with vital structures, and soft-tissue management remain outside the search space and are surgical decisions. It also assumes that planned osteotomies are executed accurately in both the donor bone and native mandible. Finally, donor-bone properties are derived from CT attenuation without resolving trabecular microarchitecture, and the chewing simulation uses an early-postoperative scar-tissue representation that may underrepresent late-stage remodeling.

\paragraph{Future Directions.} Future work will prioritize prospective evaluation in larger and more anatomically diverse cohorts, including non-fibula donors, with direct clinical significance assessed through surgeon preference studies, expert plan review, larger outcome correlations, and prospective functional and union endpoints. It will also extend the search space to co-optimize fixation hardware, including plate geometry and screw placement, with the cut-plane and donor variables. Longer-horizon healing models, intraoperative guidance, and data-driven priors learned from larger imaging cohorts are additional directions. Finally, because the current implementation is primarily a research platform, future work will package the workflow in a Docker/containerized environment to reduce installation barriers and improve reproducibility.

\printcredits

\section*{Data availability}
Data will be made available on request.

\section*{Acknowledgments}
We gratefully acknowledge financial support from the Terry Fox Research Institute (TFRI; Grant PPG 1126), the Lotte \& John Hecht Memorial Foundation, and the UBC Friedman Award for Scholars in Health. We would also like to express our gratitude to the ISTAR Group, the Institute for Computing, Information and Cognitive Systems (ICICS), and the Centre for Aging SMART for supporting this work.

\section*{Disclosure of AI Use}
Claude Opus 4.7 was used for writing and language editing only. All scientific content, interpretations, conclusions, and citations were checked, revised, and approved by all authors.

\section*{Declaration of competing interest}
The authors declare that they have no known competing financial interests or personal relationships that could have appeared to influence the work reported in this paper.

\bibliographystyle{cas-model2-names}

\bibliography{cas-refs}

@INPROCEEDINGS{aftabi2025osteoopt,

SerialNo={1},
author={Aftabi, Hamidreza and Lloyd, John E. and Ding, Amanda and Sagl,
Benedikt and Prisman, Eitan and Hodgson, Antony and Fels, Sidney},
year={2025a},
title={{OsteoOpt}: {A} {{Bayes}ian} optimization framework for enhancing bone union
likelihood in mandibular reconstruction surgery},
booktitle={MICCAI},   
pages={448--458},
}

@INPROCEEDINGS{aftabi2025optimizing,
SerialNo={2},
author={Aftabi, Hamidreza and Lloyd, John E. and Sagl, Benedikt and Ding,
Amanda and Prisman, Eitan and Hodgson, Antony and Fels, Sidney},
year={2025b},
title={Optimizing bone cuts enhances predicted bone union propensity in
mandibular body reconstruction},
booktitle={ISBI},  
pages={1--4},
}

@article{aftabi2024extent,
SerialNo={3},
author={Aftabi, Hamidreza and Sagl, Benedikt and Lloyd, John E and Prisman,
Eitan and Hodgson, Antony and Fels, Sidney},
year={2024a},
title={To what extent can mastication functionality be restored following
mandibular reconstruction surgery? {A} computer modeling approach},
journal={Computer Methods and Programs in Biomedicine},
volume={250},
pages={108174},
publisher={Elsevier},
}

@article{aftabi2024computational,
SerialNo={4},
author={Aftabi, Hamidreza and Zaraska, Katrina and Eghbal, Atabak and
McGregor, Sophie and Prisman, Eitan and Hodgson, Antony and Fels, Sidney},
year={2024b},
title={Computational models and their applications in biomechanical analysis
of mandibular reconstruction surgery},
journal={Computers in Biology and Medicine},
volume={169},
pages={107887},
publisher={Elsevier},
}

@article{ahmadi2026computational,
SerialNo={5},
author={Ahmadi, Farhad and Sun, Shuchun and Zhao, Jichao and Chen, Jian and
Wilson, Marshall B and Damon, Brooke and Wu, Yongren and Almpani,
Konstantinia and Chung, Rachel and Jani, Priyam and others},
year={2026},
title={A Computational Framework for Simulating Patient-Specific {TMJ}
Biomechanics Using a Combined Multibody Dynamics and Finite Element
Approach},
journal={Annals of Biomedical Engineering},
pages={1--14},
publisher={Springer},
}

@article{akita2019masticatory,
SerialNo={6},
author={Akita, Keiichi and Shimokawa, Toshio and Sato, Tatsuo},
year={2019},
title={Masticatory Muscles and Branches of Mandibular Nerve: {P}ositional
Relationships Between Various Muscle Bundles and Their Innervating Branches},
journal={The Anatomical Record},
volume={302},
number={4},
pages={609--619},
}

@article{bei2004multibody,
SerialNo={7},
author={Bei, Yanhong and Fregly, Benjamin J.},
year={2004},
title={Multibody dynamic simulation of knee contact mechanics},
journal={Medical engineering \& physics},
volume={26},
number={9},
pages={777--789},
publisher={Elsevier},
}

@article{besl1992method,
SerialNo={8},
author={Besl, Paul J. and McKay, Neil D.},
year={1992},
title={A method for registration of {{{3}}-D} shapes},
journal={IEEE Transactions on Pattern Analysis and Machine Intelligence},
volume={14},
number={2},
pages={239--256},
}

@inproceedings{bettin2026identifying,
SerialNo={9},
 author={Bettin, Merlin and Aftabi, Hamidreza and Lloyd, John E. and Prisman, Eitan and Fels, Sidney and Hodgson, Antony},
  title={Identifying Mini-Plate Configurations with High Predicted Bone Union Propensity for Mandibular Reconstruction Surgery},
  booktitle={Proceedings of The 25th Annual Meeting of the International Society for Computer Assisted Orthopaedic Surgery},
  series={EPiC Series in Health Sciences},
  volume={8},
  pages={44--49},
  year={2026},
  publisher={EasyChair}
}

@article{brown2017mandibular,
SerialNo={10},
author={Brown, JS and Lowe, Derek and Kanatas, Anastasios and Schache,
Andrew},
year={2017},
title={Mandibular reconstruction with vascularised bone flaps: a systematic
review over 25 years},
journal={British Journal of Oral and Maxillofacial Surgery},
volume={55},
number={2},
pages={113--126},
publisher={Elsevier},
}

@inbook{cignoni2008meshlab,
SerialNo={11},
author={Cignoni, Paolo and Callieri, Marco and Corsini, Massimiliano and
Dellepiane, Matteo and Ganovelli, Fabio and Ranzuglia, Guido and others},
year={2008},
title={Meshlab: an open-source mesh processing tool},
series={Eurographics Italian Chapter Conference},
volume={vol. 2008},
pages={129--136},
publisher={Salerno, Italy},
}

@article{corr2009biomechanical,
SerialNo={12},
author={Corr, David T and Gallant-Behm, Corrie L and Shrive, Nigel G and Hart,
David A},
year={2009},
title={Biomechanical behavior of scar tissue and uninjured skin in a porcine
model},
journal={Wound Repair and Regeneration},
volume={17},
number={2},
pages={250--259},
publisher={Wiley Online Library},
}

@article{dayarathna2024deep,
SerialNo={13},
author={Dayarathna, Sanuwani and Islam, Kh Tohidul and Uribe, Sergio and Yang,
Guang and Hayat, Munawar and Chen, Zhaolin},
year={2024},
title={Deep learning based synthesis of {MRI}, {CT} and {PET}: {R}eview and analysis},
journal={Medical image analysis},
volume={92},
pages={103046},
publisher={Elsevier},
}

@article{dempster1977maximum,
SerialNo={14},
author={Dempster, Arthur P. and Laird, Nan M. and Rubin, Donald B.},
year={1977},
title={Maximum likelihood from incomplete data via the {EM} algorithm},
journal={Journal of the Royal Statistical Society: Series B (Methodological)},
volume={39},
number={1},
pages={1--38},
}

@inbook{disa2000mandible,
SerialNo={15},
author={Disa, Joseph J. and Cordeiro, Peter G.},
year={2000},
title={Mandible reconstruction with microvascular surgery},
series={Seminars in Surgical Oncology},
volume={vol. 19},
number={no. 3},
pages={226--234},
publisher={Wiley Online Library},
}

@article{ferguson2022,
SerialNo={16},
author={Ferguson, Ben M. and Entezari, Ali and Fang, Jianguang and Li, Qing},
year={2022},
title={Optimal placement of fixation system for scaffold-based mandibular
reconstruction},
journal={Journal of the Mechanical Behavior of Biomedical Materials},
volume={126},
pages={104855},
publisher={Elsevier},
}

@article{field2010prediction,
SerialNo={17},
author={Field, Clarice and Li, Qing and Li, Wei and Thompson, Mark and Swain,
Michael},
year={2010},
title={Prediction of mandibular bone remodelling induced by fixed partial
dentures},
journal={Journal of biomechanics},
volume={43},
number={9},
pages={1771--1779},
publisher={Elsevier},
}

@article{gatti2022pycpd,
SerialNo={18},
author={Gatti, Anthony A. and Khallaghi, Siavash},
year={2022},
title={Pycpd: {P}ure numpy implementation of the coherent point drift
algorithm},
journal={Journal of Open Source Software},
volume={7},
number={80},
pages={4681},
}

@article{gil2015surgical,
SerialNo={19},
author={Gil, R Sieira and Roig, A Mar{\'\i} and Obispo, C Arranz and Morla, A
and Pag{\`e}s, C Mart{\'\i} and Perez, J Llopis},
year={2015},
title={Surgical planning and microvascular reconstruction of the mandible with
a fibular flap using computer-aided design, rapid prototype modelling, and
precontoured titanium reconstruction plates: a prospective study},
journal={British Journal of Oral and Maxillofacial Surgery},
volume={53},
number={1},
pages={49--53},
publisher={Elsevier},
}

@article{goh2008mandibular,
SerialNo={20},
author={Goh, Bee Tin and Lee, Shermin and Tideman, Henk and Stoelinga, Paul
JW},
year={2008},
title={Mandibular reconstruction in adults: a review},
journal={International journal of oral and maxillofacial surgery},
volume={37},
number={7},
pages={597--605},
publisher={Elsevier},
}

@article{guo2022emg,
SerialNo={21},
author={Guo, Jianqiao and Chen, Junpeng and Wang, Jing and Ren, Gexue and
Tian, Qiang and Guo, Chuanbin},
year={2022},
title={{{EMG}}-assisted forward dynamics simulation of subject-specific mandible
musculoskeletal system},
journal={Journal of Biomechanics},
volume={139},
pages={111143},
publisher={Elsevier},
}

@article{guo2025automated,
SerialNo={22},
author={Guo, Yan and Li, Chenyao and Yang, Rong and Tu, Puxun and Zeng, Bolun
and Liu, Jiannan and Ji, Tong and Chenping, Zhang and Chen, Xiaojun},
year={2025},
title={Automated planning of mandible reconstruction with fibula free flap
based on shape completion and morphometric descriptors},
journal={Medical Image Analysis},
volume={102},
pages={103544},
publisher={Elsevier},
}

@article{hanasono2013computer,
SerialNo={23},
author={Hanasono, Matthew M. and Skoracki, Roman J.},
year={2013},
title={Computer-assisted design and rapid prototype modeling in microvascular
mandible reconstruction},
journal={Laryngoscope},
volume={123},
number={3},
pages={597--604},
publisher={Wiley Online Library},
}

@article{hang2015tetgen,
SerialNo={24},
author={Hang, Si},
year={2015},
title={{TetGen}, a Delaunay-based quality tetrahedral mesh generator},
journal={ACM Trans. Math. Softw},
volume={41},
number={2},
pages={11},
}

@article{hannam2008dynamic,
SerialNo={25},
author={Hannam, Alan G. and Stavness, Ian and Lloyd, John E. and Fels, Sidney},
year={2008},
title={A dynamic model of jaw and hyoid biomechanics during chewing},
journal={Journal of Biomechanics},
volume={41},
number={5},
pages={1069--1076},
publisher={Elsevier},
}

@article{hannam2010comparison,
SerialNo={26},
author={Hannam, Alan G and Stavness, Ian K and Lloyd, John E and Fels, S
Sidney and Miller, Arthur J and Curtis, Donald A},
year={2010},
title={A comparison of simulated jaw dynamics in models of segmental
mandibular resection versus resection with alloplastic reconstruction},
journal={The Journal of Prosthetic Dentistry},
volume={104},
number={3},
pages={191--198},
publisher={Elsevier},
}

@article{hill1953mechanics,
SerialNo={27},
author={Hill, Archibald Vivian},
year={1953},
title={The mechanics of active muscle},
journal={Proceedings of the Royal Society of London. Series B-Biological Sciences},
volume={141},
number={902},
pages={104--117},
publisher={The Royal Society London},
}

@article{hundepool2008rehabilitation,
SerialNo={28},
author={Hundepool, AC and Dumans, AG and Hofer, SOP and Fokkens, NJW and
Rayat, SS and Van der Meij, EH and Schepman, Kees Pieter},
year={2008},
title={Rehabilitation after mandibular reconstruction with fibula free-flap:
clinical outcome and quality of life assessment},
journal={International journal of oral and maxillofacial surgery},
volume={37},
number={11},
pages={1009--1013},
publisher={Elsevier},
}

@article{jacek20183d,
SerialNo={29},
author={Jacek, Banaszewski and Maciej, Pabiszczak and Tomasz, Pastusiak and
Agata, Buczkowska and Wies{\l}aw, Kuczko and Rados{\l}aw, Wichniarek and
Filip, G{\'o}rski},
year={2018},
title={{{{3D}}} printed models in mandibular reconstruction with bony free flaps},
journal={Journal of Materials Science: Materials in Medicine},
volume={29},
number={2},
pages={23},
publisher={Springer},
}

@article{kazerouni2023diffusion,
SerialNo={30},
author={Kazerouni, Amirhossein and Aghdam, Ehsan Khodapanah and Heidari, Moein
and Azad, Reza and Fayyaz, Mohsen and Hacihaliloglu, Ilker and Merhof,
Dorit},
year={2023},
title={Diffusion models in medical imaging: {A} comprehensive survey},
journal={Medical Image Analysis},
pages={102846},
publisher={Elsevier},
}

@article{kim2024optimizing,
SerialNo={31},
author={Kim, Hugh Andrew Jinwook and De Biasio, Michael J. and Forte, Vito and
Gilbert, Ralph W. and Irish, Jonathan C. and Goldstein, David P. and de
Almeida, John R. and Hanasono, Matthew M. and Yu, Peirong and Chepeha,
Douglas B. and Looi, Thomas and Yao, Christopher M. K. L.},
year={2024},
title={Optimizing Osteotomy Geometries in Posterolateral Mandibulectomies},
journal={JAMA Otolaryngology--Head \& Neck Surgery},
volume={150},
number={12},
pages={1113--1120},
}

@article{klein2009elastix,
SerialNo={32},
author={Klein, Stefan and Staring, Marius and Murphy, Keelin and Viergever,
Max A and Pluim, Josien PW},
year={2010},
title={Elastix: a toolbox for intensity-based medical image registration},
journal={IEEE transactions on medical imaging},
volume={29},
number={1},
pages={196--205},
publisher={IEEE},
}

@article{knitschke2022osseous,
SerialNo={33},
author={Knitschke, Michael and Sonnabend, Sophia and Roller, Fritz Christian
and Pons-K{\"u}hnemann, J{\"o}rn and Schmermund, Daniel and Attia, Sameh and
Streckbein, Philipp and Howaldt, Hans-Peter and B{\"o}ttger, Sebastian},
year={2022},
title={Osseous union after mandible reconstruction with fibula free flap using
manually bent plates vs. {Patient}-specific implants: {A} retrospective analysis
of 89 patients},
journal={Current Oncology},
volume={29},
number={5},
pages={3375--3392},
publisher={MDPI},
}

@article{kumar2016mandibular,
SerialNo={34},
author={Kumar, Batchu Pavan and Venkatesh, V and Kumar, KA Jeevan and Yadav, B
Yashwanth and Mohan, S Ram},
year={2016},
title={Mandibular reconstruction: overview},
journal={Journal of maxillofacial and oral surgery},
volume={15},
pages={425--441},
publisher={Springer},
}

@article{li2023current,
SerialNo={35},
author={Li, Yongdi and Li, Duchenhui and Tang, Zhenglong and Wang, Dongxiang
and Yang, Zhishan and Liu, Yiheng},
year={2023},
title={Current global research on mandibular defect: {A} bibliometric analysis
from 2001 to 2021},
journal={Frontiers in Bioengineering and Biotechnology},
volume={11},
pages={1061567},
publisher={Frontiers},
}

@article{lin2010bone,
SerialNo={36},
author={Lin, Daniel and Li, Qing and Li, Wei and Swain, Michael},
year={2010},
title={Bone remodeling induced by dental implants of functionally graded
materials},
journal={Journal of Biomedical Materials Research Part B: Applied Biomaterials: An Official Journal of The Society for Biomaterials, The Japanese Society for Biomaterials, and The Australian Society for Biomaterials and the Korean Society for Biomaterials},
volume={92},
number={2},
pages={430--438},
publisher={Wiley Online Library},
}

@misc{lloyd2025modelguide,
SerialNo={37},
author={Lloyd, John E. and S{\'a}nchez, Antonio},
year={2025},
title={{ArtiSynth} Modeling Guide},
howpublished={\url{https://www.artisynth.org/doc/html/modelguide/modelguide.html}},
note={Last update: June 2025},
}

@incollection{lloyd2012artisynth,
SerialNo={38},
author={Lloyd, John E. and Stavness, Ian and Fels, Sidney},
year={2012},
title={{ArtiSynth}: {A} fast interactive biomechanical modeling toolkit combining
multibody and finite element simulation},
booktitle={Soft Tissue Biomechanical Modeling for Computer Assisted Surgery},
pages={355--394},
publisher={Springer},
}

@article{mahesh2013essential,
SerialNo={39},
author={Mahesh, Mahadevappa},
year={2013},
title={The Essential Physics of Medical Imaging, Third Edition},
journal={Medical physics},
volume={40},
number={7},
pages={077301},
publisher={Wiley Online Library},
}

@article{mezey2022masseter,
SerialNo={40},
author={Mezey, Szilvia E. and M{\"u}ller-Gerbl, Magdalena and Toranelli,
Mireille and T{\"u}rp, Jens C.},
year={2022},
title={The human masseter muscle revisited: {F}irst description of its coronoid
part},
journal={Annals of Anatomy},
volume={240},
pages={151879},
}

@article{morgan2018bone,
SerialNo={41},
author={Morgan, Elise F. and Unnikrisnan, Ginu U. and Hussein, Amira I.},
year={2018},
title={Bone mechanical properties in healthy and diseased states},
journal={Annual review of biomedical engineering},
volume={20},
number={1},
pages={119--143},
publisher={Annual Reviews},
}

@book{pymeshlab,
SerialNo={42},
author={Alessandro Muntoni and Paolo Cignoni},
year={2021},
title={{{PyMeshLab}}},
publisher={Zenodo},
}

@article{myronenko2010cpd,
SerialNo={43},
author={Myronenko, Andriy and Song, Xubo},
year={2010},
title={Point set registration: coherent point drift},
journal={IEEE Transactions on Pattern Analysis and Machine Intelligence},
volume={32},
number={12},
pages={2262--2275},
}

@article{nakao2017automated,
SerialNo={44},
author={Nakao, Megumi and Aso, Sayako and Imai, Yutaro and Ueda, Naoki and
Hatanaka, Tomoyuki and Shiba, Masaaki and Kirita, Tadaaki and Matsuda,
Tetsuya},
year={2017},
title={Automated Planning With Multivariate Shape Descriptors for Fibular
Transfer in Mandibular Reconstruction},
journal={IEEE Transactions on Biomedical Engineering},
volume={64},
number={8},
pages={1772--1785},
}

@article{nguyen2021maxillectomy,
SerialNo={45},
author={Nguyen, Sally and Tran, Khanh Linh and Wang, Edward and Britton, Heidi
and Durham, James Scott and Prisman, Eitan},
year={2021},
title={Maxillectomy defects: {V}irtually comparing fibular and scapular free
flap reconstructions},
journal={Head \& Neck},
volume={43},
number={9},
pages={2623--2633},
publisher={Wiley Online Library},
}

@article{peck2000dynamic,
SerialNo={46},
author={Peck, C.C. and Langenbach, G.E.J. and Hannam, A.G.},
year={2000},
title={Dynamic simulation of muscle and articular properties during human wide
jaw opening},
journal={Archives of Oral Biology},
volume={45},
number={11},
pages={963--982},
publisher={Elsevier},
}

@article{prasad2019invertible,
SerialNo={47},
author={Prasad, Jitendra and Goyal, Ajay},
year={2019},
title={An invertible mathematical model of cortical bone adaptation to
mechanical loading},
journal={Scientific reports},
volume={9},
number={1},
pages={5890},
publisher={Nature Publishing Group UK London},
}

@article{qin2017improving,
SerialNo={48},
author={Qin, Chao and Klabjan, Diego and Russo, Daniel},
year={2017},
title={Improving the expected improvement algorithm},
journal={Advances in Neural Information Processing Systems},
volume={30},
}

@article{renardy2021sobol,
SerialNo={49},
author={Renardy, Marissa and Joslyn, Louis R and Millar, Jess A and Kirschner,
Denise E},
year={2021},
title={To {Sobol} or not to {Sobol}? {T}he effects of sampling schemes in systems
biology applications},
journal={Mathematical biosciences},
volume={337},
pages={108593},
publisher={Elsevier},
}

@article{roumanas2006masticatory,
SerialNo={50},
author={Roumanas, Eleni D and Garrett, Neal and Blackwell, Keith E and
Freymiller, Earl and Abemayor, Elliot and Wong, Weng Kee and Beumer III, John
and Fueki, Kenji and Fueki, Warawan and Kapur, Krishan K},
year={2006},
title={Masticatory and swallowing threshold performances with conventional and
implant-supported prostheses after mandibular fibula free-flap
reconstruction},
journal={The Journal of prosthetic dentistry},
volume={96},
number={4},
pages={289--297},
publisher={Elsevier},
}

@article{rubin1984regulation,
SerialNo={51},
author={Rubin, Clinton T. and Lanyon, Lance E.},
year={1984},
title={Regulation of bone formation by applied dynamic loads},
journal={JBJS},
volume={66},
number={3},
pages={397--402},
publisher={LWW},
}

@article{rungsiyakull2011loading,
SerialNo={52},
author={Rungsiyakull, Pimduen and Rungsiyakull, Chaiy and Appleyard, Richard
and Li, Qing and Swain, Michael and Klineberg, Iven},
year={2011},
title={Loading of a single implant in simulated bone},
journal={International Journal of Prosthodontics},
volume={24},
number={2},
pages={140--143},
}

@article{sabiq2024evaluating,
SerialNo={53},
author={Sabiq, Farahna and Cherukupalli, Abhiram and Khalil, Mohammad and
Tran, Linh K and Kwon, Jamie JY and Milner, Thomas and Durham, James S and
Prisman, Eitan},
year={2024},
title={Evaluating the benefit of virtual surgical planning on bony union rates
in head and neck reconstructive surgery},
journal={Head \& Neck},
volume={46},
number={6},
pages={1322--1330},
publisher={Wiley Online Library},
}

@article{sagl2019dynamic,
SerialNo={54},
author={Sagl, Benedikt and Schmid-Schwap, Martina and Piehslinger, Eva and
Kundi, Michael and Stavness, Ian},
year={2019},
title={A dynamic jaw model with a finite-element temporomandibular joint},
journal={Frontiers in Physiology},
volume={10},
pages={1156},
publisher={Frontiers Media SA},
}

@article{sagl2021silico,
SerialNo={55},
author={Sagl, Benedikt and Schmid-Schwap, Martina and Piehslinger, Eva and
Rausch-Fan, Xiaohui and Stavness, Ian},
year={2021},
title={An in silico investigation of the effect of bolus properties on {TMJ}
loading during mastication},
journal={journal of the mechanical behavior of biomedical materials},
volume={124},
pages={104836},
publisher={Elsevier},
}

@article{yu2021morphology,
SerialNo={56},
author={Yu, Sun Kyoung and Kim, Tae-Hoon and Yang, Kwang Yeol and Bae,
Christopher J. and Kim, Heung-Joong},
year={2021},
title={Morphology of the temporalis muscle focusing on the tendinous
attachment onto the coronoid process},
journal={Anatomy \& Cell Biology},
volume={54},
number={3},
pages={308--314},
}

@article{shahriari2015taking,
SerialNo={57},
author={Shahriari, Bobak and Swersky, Kevin and Wang, Ziyu and Adams, Ryan P
and De Freitas, Nando},
year={2016},
title={Taking the human out of the loop: {A} review of {{Bayes}ian} optimization},
journal={Proceedings of the IEEE},
volume={104},
number={1},
pages={148--175},
publisher={IEEE},
}

@article{shkedy2020predicting,
SerialNo={58},
author={Shkedy, Yotam and Howlett, Joel and Wang, Edward and Ongko, Jennifer
and Scott Durham, J and Prisman, Eitan},
year={2020},
title={Predicting the number of fibular segments to reconstruct mandibular
defects},
journal={Laryngoscope},
volume={130},
number={11},
pages={E619--E624},
publisher={Wiley Online Library},
}

@article{stavness2010predicting,
SerialNo={59},
author={Stavness, Ian and Hannam, Alan G. and Lloyd, John E. and Fels, Sidney},
year={2010},
title={Predicting muscle patterns for hemimandibulectomy models},
journal={Computer Methods in Biomechanics and Biomedical Engineering},
volume={13},
number={4},
pages={483--491},
publisher={Taylor \& Francis},
}

@article{swendseid2020natural,
SerialNo={60},
author={Swendseid, Brian and Kumar, Ayan and Sweeny, Larissa and Zhan,
Tingting and Goldman, Richard A and Krein, Howard and Heffelfinger, Ryan N
and Luginbuhl, Adam J and Curry, Joseph M},
year={2020},
title={Natural history and consequences of nonunion in mandibular and
maxillary free flaps},
journal={Otolaryngology--Head and Neck Surgery},
volume={163},
number={5},
pages={956--962},
publisher={SAGE Publications Sage CA: Los Angeles, CA},
}

@article{tran2023virtual,
SerialNo={61},
author={Tran, Khanh Linh and Kwon, Jae Young and Gui, Xi Yao and Wang, Edward
and Yang, David and Durham, James Scott and Prisman, Eitan},
year={2023a},
title={Virtual surgical planning for maxillary reconstruction with the
scapular free flap: {A}n evaluation of a simple cutting guide design},
journal={Head \& Neck},
volume={45},
number={1},
pages={115--125},
publisher={Wiley Online Library},
}

@article{tran2023dental,
SerialNo={62},
author={Tran, Khanh Linh and Yang, David H and Wang, Edward and Ham, Jennifer
Inseon and Wong, Angela and Panchal, Maharshi and Dial, Harkaran Singh and
Durham, James Scott and Prisman, Eitan},
year={2023b},
title={Dental implantability of mandibular reconstructions: {C}omparing freehand
surgery with virtual surgical planning},
journal={Oral Oncology},
volume={140},
pages={106396},
publisher={Elsevier},
}

@article{urken1991oromandibular,
SerialNo={63},
author={Urken, Mark L and Weinberg, Hubert and Vickery, Carlin and Buchbinder,
Daniel and Lawson, William and Biller, Hugh F},
year={1991},
title={Oromandibular reconstruction using microvascular composite free flaps:
report of 71 cases and a new classification scheme for bony, soft-tissue, and
neurologic defects},
journal={Archives of Otolaryngology--Head \& Neck Surgery},
volume={117},
number={7},
pages={733--744},
publisher={American Medical Association},
}

@article{vyas2022virtual,
SerialNo={64},
author={Vyas, Krishna and Gibreel, Waleed and Mardini, Samir},
year={2022},
title={Virtual surgical planning ({VSP}) in craniomaxillofacial reconstruction},
journal={Facial Plastic Surgery Clinics},
volume={30},
number={2},
pages={239--253},
publisher={Elsevier},
}

@article{wan2022interaction,
SerialNo={65},
author={Wan, Boyang and Yoda, Nobuhiro and Zheng, Keke and Zhang, Zhongpu and
Wu, Chi and Clark, Jonathan and Sasaki, Keiichi and Swain, Michael and Li,
Qing},
year={2022},
title={On interaction between fatigue of reconstruction plate and
time-dependent bone remodeling},
journal={Journal of the Mechanical Behavior of Biomedical Materials},
volume={136},
pages={105483},
publisher={Elsevier},
}

@article{wang2016mandibular,
SerialNo={66},
author={Wang, YY and Zhang, HQ and Fan, S and Zhang, DM and Huang, ZQ and
Chen, WL and Ye, JT and Li, JS},
year={2016},
title={Mandibular reconstruction with the vascularized fibula flap: comparison
of virtual planning surgery and conventional surgery},
journal={International journal of oral and maxillofacial surgery},
volume={45},
number={11},
pages={1400--1405},
publisher={Elsevier},
}

@article{wasserthal2023totalsegmentator,
SerialNo={67},
author={Wasserthal, Jakob and Breit, Hanns-Christian and Meyer, Manfred T and
Pradella, Maurice and Hinck, Daniel and Sauter, Alexander W and Heye, Tobias
and Boll, Daniel T and Cyriac, Joshy and Yang, Shan and others},
year={2023},
title={{TotalSegmentator}: robust segmentation of 104 anatomic structures in {CT}
images},
journal={Radiology: Artificial Intelligence},
volume={5},
number={5},
pages={e230024},
publisher={Radiological Society of North America},
}

@article{weijs1984relationship,
SerialNo={68},
author={Weijs, W.A. and Hillen, B.},
year={1984},
title={Relationship between the physiological cross-section of the human jaw
muscles and their cross-sectional area in computer tomograms},
journal={Cells Tissues Organs},
volume={118},
number={3},
pages={129--138},
publisher={S. Karger AG Basel, Switzerland},
}

@article{weijs1985strength,
SerialNo={69},
author={Weijs, W. A. and Hillen, B.},
year={1985},
title={Cross-sectional areas and estimated intrinsic strength of the human jaw
muscles},
journal={Acta Morphologica Neerlando-Scandinavica},
volume={23},
number={3},
pages={267--274},
aetag={\xvfill\xeject},
}

@article{wong2010biomechanics,
SerialNo={70},
author={Wong, Raymond C.W. and Tideman, Henk and Kin, L. and Merkx, Matthias A.W.},
year={2010},
title={Biomechanics of mandibular reconstruction: a review},
journal={International journal of oral and maxillofacial surgery},
volume={39},
number={4},
pages={313--319},
publisher={Elsevier},
}

@article{wu2021machine,
SerialNo={71},
author={Wu, Chi and Entezari, Ali and Zheng, Keke and Fang, Jianguang and
Zreiqat, Hala and Steven, Grant P and Swain, Michael V and Li, Qing},
year={2021},
title={A machine learning-based multiscale model to predict bone formation in
scaffolds},
journal={Nat. Comput. Sci.},
volume={1},
pages={532--541},
}

@article{yang2022spatial,
SerialNo={72},
author={Yang, W-F and Choi, WS and Zhu, W-Y and Zhang, C-Y and Li, DTS and
Tsoi, JK-H and Tang, AW-L and Kwok, K-W and Su, Y-X},
year={2022},
title={Spatial deviations of the temporomandibular joint after oncological
mandibular reconstruction},
journal={International Journal of Oral and Maxillofacial Surgery},
volume={51},
number={1},
pages={44--53},
publisher={Elsevier},
}

@article{wu2025anatomical,
SerialNo={73},
author={Wu, Songying and Pu, Jingya Jane and Pow, Edmond Ho Nang and
Leung, Pui Hang and Yu, Xing-Na and Su, Yu-Xiong and Yang, Wei-Fa},
year={2025},
title={Anatomical Landmark-guided Strategy for Computer-assisted
Reconstruction of Infrastructure Maxillary Defects Using Free Fibula Flap},
journal={Plastic and Reconstructive Surgery -- Global Open},
volume={13},
number={3},
pages={e6626},
}

@article{zheng2019investigation,
SerialNo={74},
author={Zheng, Keke and Liao, Zhipeng and Yoda, Nobuhiro and Fang, Jianguang
and Chen, Junning and Zhang, Zhongpu and Zhong, Jingxiao and Peck,
Christopher and Sasaki, Keiichi and Swain, Michael V and others},
year={2019},
title={Investigation on masticatory muscular functionality following oral
reconstruction--{An} inverse identification approach},
journal={Journal of biomechanics},
volume={90},
pages={1--8},
publisher={Elsevier},
}

@article{zheng2022,
SerialNo={75},
author={Zheng, Keke and Yoda, Nobuhiro and Chen, Junning and Liao, Zhipeng and
Zhong, Jingxiao and Wu, Chi and Wan, Boyang and Koyama, Shigeto and Sasaki,
Keiichi and Peck, Christopher and others},
year={2022},
title={Bone remodeling following mandibular reconstruction using fibula free
flap},
journal={Journal of Biomechanics},
volume={133},
pages={110968},
publisher={Elsevier},
}

\clearpage
\appendix
\counterwithin{figure}{section}
\counterwithin{table}{section}
\counterwithin{equation}{section}

\setlength{\dbltextfloatsep}{10pt plus 2pt minus 2pt}
\setlength{\dblfloatsep}{10pt plus 2pt minus 2pt}

\section{Supplementary Materials}
\label{app:functional_modeling_parameters}

\subsection{Gaussian-Process Surrogate Equations}
\label{app:gp_surrogate}

Using the standard kernel matrix \(K_N\), covariance vector \(\mathbf{k}_N(\boldsymbol{\phi})\), observation vector \(\mathbf{y}_N\), and prior-mean vector \(\boldsymbol{\mu}_N\), the GP posterior mean and latent variance are
\begin{equation}
\begin{aligned}
\mu_N(\boldsymbol{\phi}) &=
\mu(\boldsymbol{\phi}) +
\mathbf{k}_N(\boldsymbol{\phi})^T
\left(K_N+\sigma^2 I\right)^{-1}
\left(\mathbf{y}_N-\boldsymbol{\mu}_N\right),\\
\sigma_F^2(\boldsymbol{\phi}) &=
k(\boldsymbol{\phi},\boldsymbol{\phi}) -
\mathbf{k}_N(\boldsymbol{\phi})^T
\left(K_N+\sigma^2 I\right)^{-1}
\mathbf{k}_N(\boldsymbol{\phi}),
\end{aligned}
\label{eq:structopt_gp_posterior}
\end{equation}
with predictive variance \(\sigma_Q^2(\boldsymbol{\phi})=\sigma_F^2(\boldsymbol{\phi})+\sigma^2\).

The automatic relevance determination (ARD) Matérn 5/2 kernel is
\begin{equation}
\begin{aligned}
k(\boldsymbol{\phi},\boldsymbol{\phi}')
&=
\sigma_f^2(1+\sqrt{5}\,r+\tfrac{5}{3}r^2)
\exp(-\sqrt{5}\,r), \\
r
&=
\left[
\sum_{m=1}^{d}
\frac{(\phi_m-\phi_m')^2}{\ell_m^2}
\right]^{1/2}.
\end{aligned}
\label{eq:structopt_ard_matern}
\end{equation}
The separate length scales \(\ell_m\) encode anisotropic sensitivity across the surgical design variables.

\subsection{Rigid and Deformable Registration Equations}
\label{app:registration_equations}

The maxilla-based rigid initialization maps a template point \(\mathbf{x}\in\mathbb{R}^3\) to
\begin{equation}
\mathbf{x}^{(0)}=\mathbf{R}_0\mathbf{x}+\mathbf{t}_0,
\label{eq:psm_rigid_init}
\end{equation}
where \(\mathbf{R}_0\in SO(3)\) and \(\mathbf{t}_0\in\mathbb{R}^3\) are obtained from centroid alignment followed by iterative closest point refinement of the generic and patient maxillary surfaces~\citep{besl1992method}.

After rigid initialization, coherent point drift (CPD) estimates a nonrigid correspondence by treating the template control points as centroids of a Gaussian mixture model~\citep{myronenko2010cpd}. For each anatomical support \(k\in\{\mathrm{mand},\mathrm{max}\}\), patient points \(\{\mathbf{x}_n\}_{n=1}^{N}\) are modeled as observations generated from deformed template centroids \(\mathcal{T}_k(\mathbf{y}^{(k)}_m)\), with an additional outlier component for unmatched points. The deformation is represented by
\begin{equation}
\begin{aligned}
\mathcal{T}_k(\mathbf{x})
&= \mathbf{x}
+ \sum_{m=1}^{M_k}
G_{\beta_k}\!\left(\mathbf{x},\mathbf{y}^{(k)}_m\right)
\mathbf{w}^{(k)}_m, \\
G_{\beta_k}\!\left(\mathbf{x},\mathbf{y}\right)
&= \exp\!\left(
-\frac{\|\mathbf{x}-\mathbf{y}\|_2^2}{2\beta_k^2}
\right),
\end{aligned}
\label{eq:psm_cpd}
\end{equation}
where \(\{\mathbf{y}^{(k)}_m\}_{m=1}^{M_k}\) are the template control points, \(\mathbf{w}^{(k)}_m\) are the displacement coefficients, and \(\beta_k\) controls the spatial smoothness of the deformation.

The CPD parameters are estimated by expectation-maximization. In the expectation step, the posterior correspondence probability is
\begin{equation}
P_{mn}
=
\frac{
\exp\!\left[-\|\mathbf{x}_n-\mathcal{T}_k(\mathbf{y}^{(k)}_m)\|_2^2/(2\sigma_{\mathrm{CPD}}^2)\right]
}{
\sum_{j=1}^{M_k}
\exp\!\left[-\|\mathbf{x}_n-\mathcal{T}_k(\mathbf{y}^{(k)}_j)\|_2^2/(2\sigma_{\mathrm{CPD}}^2)\right] + c
},
\label{eq:psm_cpd_estep}
\end{equation}
where \(c\) is the standard CPD outlier-weight term. In the maximization step, the displacement coefficients \(\mathbf{W}_k=[\mathbf{w}^{(k)}_1,\ldots,\mathbf{w}^{(k)}_{M_k}]^T\) are updated by minimizing the regularized expected mismatch,
\begin{equation}
E(\mathbf{W}_k)
=
\sum_{m=1}^{M_k}\sum_{n=1}^{N}
P_{mn}
\left\|
\mathbf{x}_n-\mathcal{T}_k(\mathbf{y}^{(k)}_m)
\right\|_2^2
\;+\;
\lambda\,\mathrm{tr}\!\left(\mathbf{W}_k^{T}G_k\mathbf{W}_k\right),
\label{eq:psm_cpd_mstep}
\end{equation}
yielding a coherent displacement field controlled by the smoothness kernel \(G_k\).

\subsection{Forward and Inverse Dynamics Simulation}
\label{app:forward_inverse}

Both modes use the ArtiSynth multibody dynamics
\begin{equation}
\bar{M}\dot{\vec{v}} = \vec{f}_p(\vec{x},\vec{v}) + \bar{A}(\vec{x},\vec{v})\,\vec{a},
\label{eq:app_fd}
\end{equation}
with mass matrix \(\bar{M}\), passive forces \(\vec{f}_p\) (muscle stretch, ligaments, scar tissue, bolus), activations \(\vec{a}\), and Hill-type activation-to-force map \(\bar{A}\)~\citep{hill1953mechanics}.

\begin{figure*}[t]
\centering
\subfloat[]{\includegraphics[width=.48\linewidth]{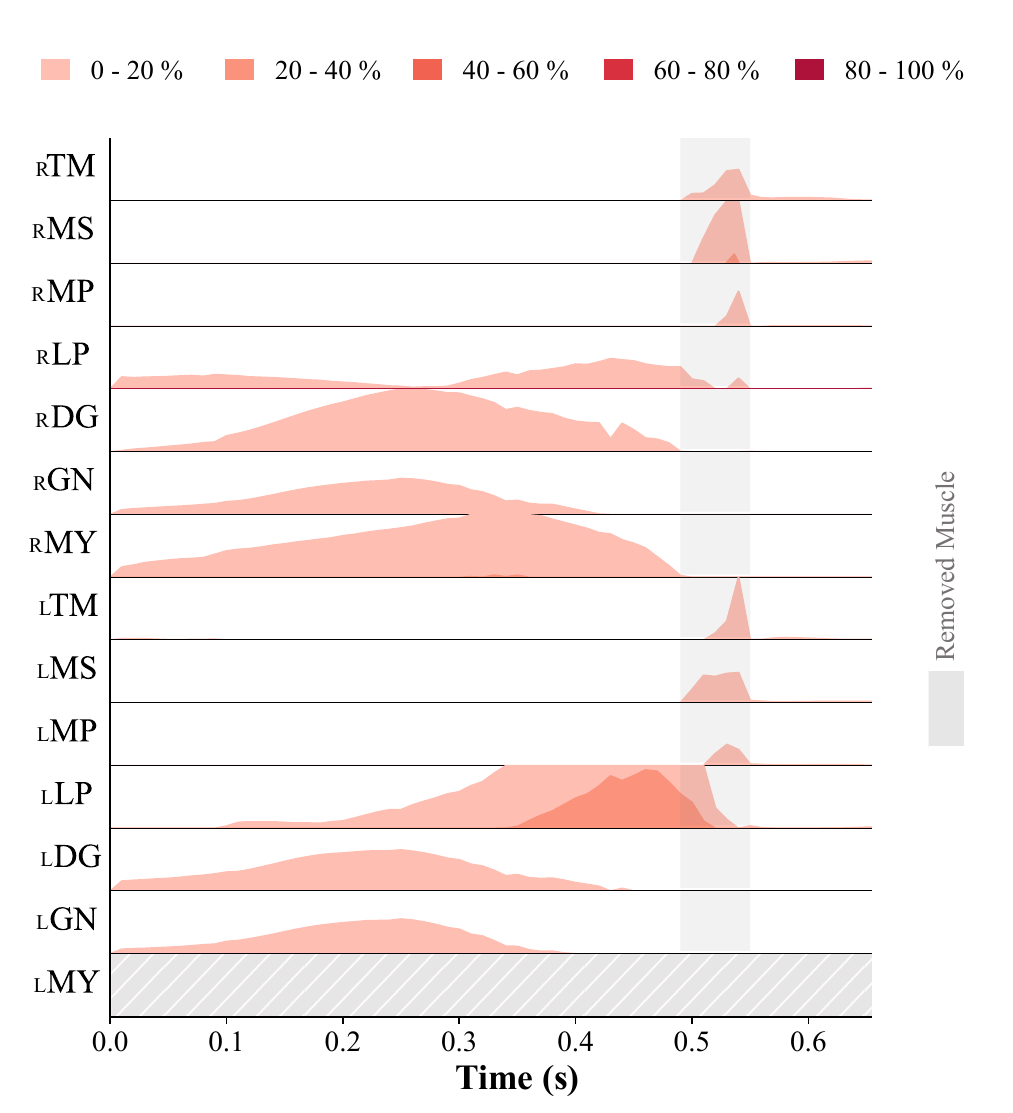}}\hfill
\subfloat[]{\includegraphics[width=.48\linewidth]{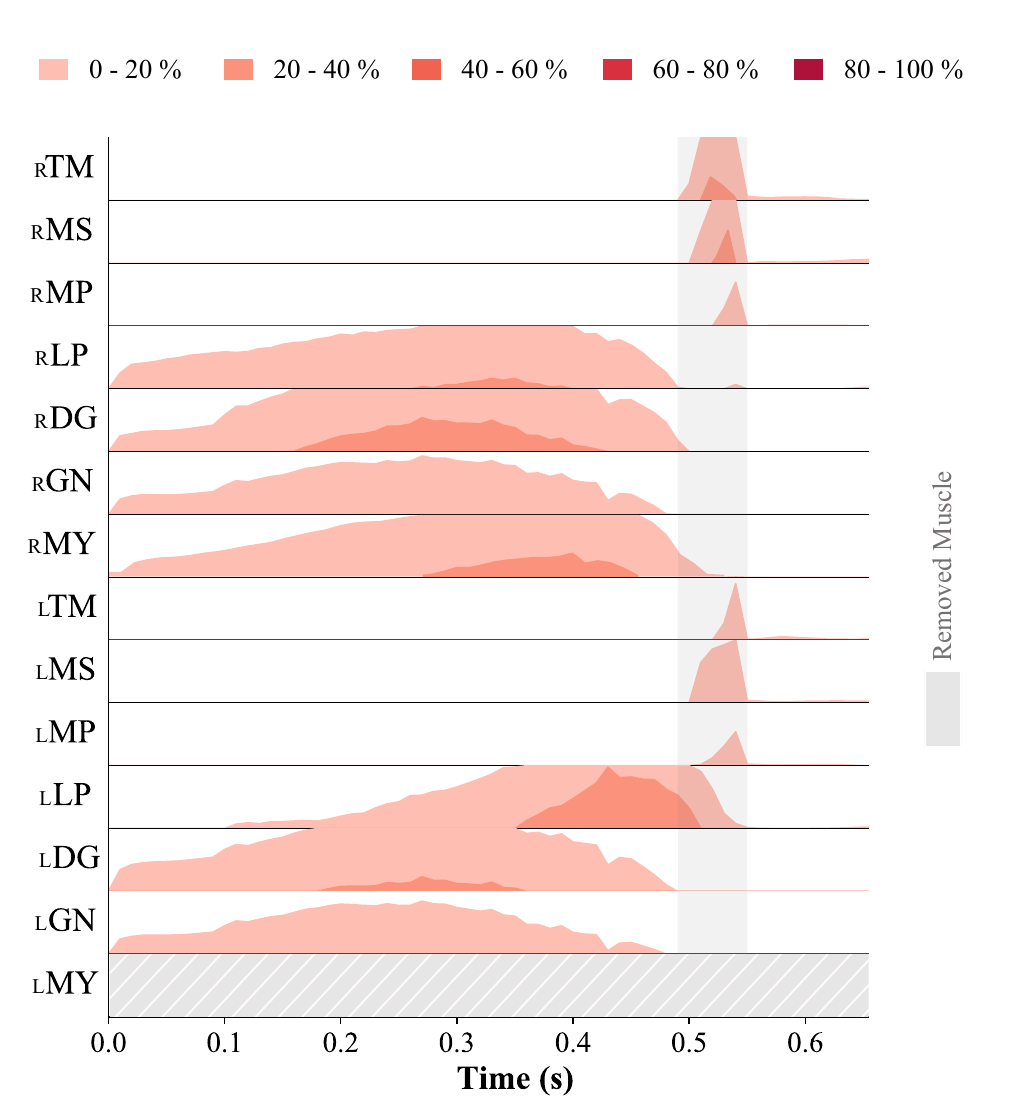}}
\caption{Muscle activation for the patient B defect: (a) normal activation used in the forward simulation; (b) adapted activation recovered by inverse simulation to follow the pre-reconstruction jaw trajectory. The gray background marks the bolus-engagement interval, and the hatched lines indicate the muscles removed for the reconstruction.}
\label{fig:fwd_inv_b}
\end{figure*}

\noindent\textbf{Forward dynamics.} Activations are taken from experimental unilateral chewing, and Eq.~\ref{eq:app_fd} is integrated forward to obtain the motion that scores each candidate, with activations at pre-reconstruction levels (Fig.~\ref{fig:fwd_inv_b}(a)).

\noindent\textbf{Inverse dynamics.} For a target motion \(\vec{v}^{*}\), the activations are recovered by
\begin{equation}
\vec{a}^{*}=\arg\min_{\vec{a}}\; w_v\lVert\vec{v}^{*}-\vec{v}^{j+1}\rVert^{2}+\tfrac{w_a}{2}\,\vec{a}^{\top}\bar{W}^{-1}\vec{a}+w_d\lVert\vec{a}-\vec{a}^{j-1}\rVert^{2},
\label{eq:app_id}
\end{equation}
subject to Eq.~\ref{eq:app_fd} and \(0\le a_i\le 1\), trading off tracking, effort, and smoothness. Taking the pre-reconstruction trajectory as \(\vec{v}^{*}\) and letting the muscles remaining after resection follow it gives the adapted activation (Fig.~\ref{fig:fwd_inv_b}(b)). Full formulations follow~\citep{stavness2010predicting,aftabi2024extent}.

\subsection{Scan Cross Section Extraction}
\label{app:pcsa_scs_workflow}

For PCSA estimation, scan cross-section planes were defined from craniofacial landmarks following the referenced extraction protocol~\citep{zheng2019investigation,weijs1984relationship}. A masseter/medial-pterygoid reference plane was defined at $30^\circ$ to the Frankfort horizontal plane using the left and right mandibular angle points, and was translated $25$ mm anterosuperiorly along its normal direction. The temporalis reference plane was positioned $10$ mm superior to the Frankfort horizontal plane along its normal. For the lateral pterygoid, the reference plane was taken perpendicular to the Frankfort horizontal plane using the left and right lateral pole points, and was translated $10$ mm anteriorly along its normal. For each muscle, the scan cross section (SCS) was measured on the reference plane and on 10 neighboring parallel planes located 1--5 mm above and below the reference plane, or anterior and posterior for the lateral pterygoid. The largest SCS was retained, since PCSA estimation error decreases as the measured SCS increases~\citep{zheng2019investigation,weijs1984relationship}.

\clearpage

\begin{table*}[!t]
\subsection{Parameter and Optimization Tables}

\vspace{0.75em}
\centering
\begin{threeparttable}
\caption{Maximum-to-optimal muscle length ratios ($r_m$) inherited from the generic template, used to update patient-specific muscle lengths via Eq.~\ref{eq:psm_muscle_lengths}.}
\label{tab:func_supp_9}
\footnotesize
\setlength{\tabcolsep}{12pt}
\renewcommand{\arraystretch}{1.1}
\begin{tabular*}{\textwidth}{@{\extracolsep{\fill}}lll}
\toprule
Abbreviation & Muscle name & $r_m$ \\
\midrule
{\scriptsize R}TM$_a$ & Right temporalis (anterior)        & 1.27 \\
{\scriptsize R}TM$_m$ & Right temporalis (medial)          & 1.42 \\
{\scriptsize R}TM$_p$ & Right temporalis (posterior)       & 1.31 \\
{\scriptsize R}MS$_s$ & Right masseter (superficial)       & 1.30 \\
{\scriptsize R}MS$_d$ & Right masseter (deep)              & 1.52 \\
{\scriptsize R}MP     & Right medial pterygoid             & 1.25 \\
{\scriptsize R}LP$_s$ & Right lateral pterygoid (superior) & 1.36 \\
{\scriptsize R}LP$_i$ & Right lateral pterygoid (inferior) & 1.42 \\
{\scriptsize R}DG     & Right anterior digastric           & 1.28 \\
{\scriptsize R}GN     & Right geniohyoid                   & 1.28 \\
{\scriptsize R}MY$_a$ & Right mylohyoid (anterior)         & 1.28 \\
{\scriptsize R}MY$_p$ & Right mylohyoid (posterior)        & 1.61 \\
{\scriptsize L}TM$_a$ & Left temporalis (anterior)         & 1.27 \\
{\scriptsize L}TM$_m$ & Left temporalis (medial)           & 1.42 \\
{\scriptsize L}TM$_p$ & Left temporalis (posterior)        & 1.31 \\
{\scriptsize L}MS$_s$ & Left masseter (superficial)        & 1.30 \\
{\scriptsize L}MS$_d$ & Left masseter (deep)               & 1.54 \\
{\scriptsize L}MP     & Left medial pterygoid              & 1.25 \\
{\scriptsize L}LP$_s$ & Left lateral pterygoid (superior)  & 1.36 \\
{\scriptsize L}LP$_i$ & Left lateral pterygoid (inferior)  & 1.32 \\
{\scriptsize L}DG     & Left anterior digastric            & 1.28 \\
{\scriptsize L}GN     & Left geniohyoid                    & 1.28 \\
{\scriptsize L}MY$_a$ & Left mylohyoid (anterior)          & 1.28 \\
{\scriptsize L}MY$_p$ & Left mylohyoid (posterior)         & 1.55 \\
\bottomrule
\end{tabular*}
\begin{tablenotes}[flushleft]
\footnotesize
\item Ratios extracted as $\text{maxLength}/\text{optLength}$ from the generic muscle-material definitions~\citep{hannam2008dynamic,aftabi2024extent}. Each side carries its own ratio, computed from the corresponding optimal and maximum fiber lengths of that muscle.
\end{tablenotes}
\end{threeparttable}
\end{table*}


\begin{table*}[!t]
\centering

\begin{minipage}[t]{0.33\textwidth}
\centering

\refstepcounter{table}
\label{tab:func_supp_10}
\noindent\parbox[t]{\linewidth}{%
\footnotesize
\textbf{Table~\thetable.}
PCSA branch allocation for the four masticatory muscle groups~\citep{hannam2008dynamic,aftabi2024extent}.%
}
\vspace{4pt}

\footnotesize
\setlength{\tabcolsep}{3pt}
\renewcommand{\arraystretch}{1.1}
\begin{tabular}{lll}
\toprule
Group & Branch & Prop. \\
\midrule
MS & Superficial & 0.70 \\
MS & Deep        & 0.30 \\
MP & --          & 1.00 \\
TM & Anterior    & 0.48 \\
TM & Medial      & 0.29 \\
TM & Posterior   & 0.23 \\
LP & Superior    & 0.30 \\
LP & Inferior    & 0.70 \\
\bottomrule
\end{tabular}

\end{minipage}%
\hspace{0.03\textwidth}%
\begin{minipage}[t]{0.62\textwidth}
\centering

\refstepcounter{table}
\label{tab:func_supp_11}
\noindent\parbox[t]{\linewidth}{%
\footnotesize
\textbf{Table~\thetable.}
PCSA regression coefficients used in Eq.~\ref{eq:psm_pcsa_regression}, with the nominal patient-specific PCSA defined by Eq.~\ref{eq:psm_pcsa_mean}.%
}
\vspace{4pt}

\footnotesize
\setlength{\tabcolsep}{4pt}
\renewcommand{\arraystretch}{1.1}
\begin{threeparttable}
\begin{tabular}{lcccccc}
\toprule
Muscle & \multicolumn{3}{c}{WPCS (Weber)} & \multicolumn{3}{c}{BPCS (Buchner)} \\
\cmidrule(lr){2-4}\cmidrule(lr){5-7}
 & Slope & Intercept & Error & Slope & Intercept & Error \\
\midrule
MS & 1.52 &  1.04 & 0.86 & 1.11 &  0.85 & 0.29 \\
TM & 2.45 & -1.85 & 1.34 & 1.87 & -1.51 & 0.75 \\
MP & 2.34 & -2.04 & 0.55 & 1.56 & -1.40 & 0.50 \\
LP & 1.55 & -3.42 & 0.41 & 0.93 & -1.51 & 0.40 \\
\bottomrule
\end{tabular}
\begin{tablenotes}[flushleft]
\footnotesize
\item WPCS: Weber-based; BPCS: Buchner-based regression~\citep{weijs1984relationship}.
\end{tablenotes}
\end{threeparttable}

\end{minipage}

\end{table*}

\begin{table*}[!t]
\centering
\begin{threeparttable}
\caption{Muscles omitted for each defect type to model muscle and tissue loss after resection, following the Urken-based classification and our prior craniofacial reconstruction model~\citep{aftabi2024extent}.}
\label{tab:resected_muscles}
\footnotesize
\setlength{\tabcolsep}{12pt}
\renewcommand{\arraystretch}{1.2}
\begin{tabular*}{\textwidth}{@{\extracolsep{\fill}}ll}
\toprule
Defect type & Omitted muscles \\
\midrule
B (body)         & Mylohyoid (anterior, posterior) \\
S (symphysis)    & Geniohyoid, anterior digastric \\
RB (ramus--body) & Masseter (superficial, deep), temporalis (anterior, medial, posterior), \\
                 & medial pterygoid, mylohyoid (anterior, posterior) \\
\bottomrule
\end{tabular*}
\begin{tablenotes}[flushleft]
\footnotesize
\item Muscles are omitted on the resected side. The RB set combines the ramus (R) and body (B) removals; the elementary R, B, and S removals follow \citet{aftabi2024extent}.
\end{tablenotes}
\end{threeparttable}
\end{table*}

\begin{table*}[t]
\centering
\begin{threeparttable}
\caption{Optimized design-variable values in physical units for the generic and patient-specific cases under the \(F_{\mathrm{opt}}\) and \(F_{\mathrm{SF}}\) objectives. Angles are in degrees and offsets in millimeters; \(l_{RDP}\) applies only to the two-segment RB defect. Values are means over five trials.}
\label{tab:optimized_values}
\footnotesize
\setlength{\tabcolsep}{6pt}
\renewcommand{\arraystretch}{1.2}
\begin{tabular*}{\textwidth}{@{\extracolsep{\fill}}lcl cccc cc}
\toprule
Case & Defect & Obj. & $\theta_{Lr}$ (\textdegree) & $\theta_{Lp}$ (\textdegree) & $\theta_{Rr}$ (\textdegree) & $\theta_{Rp}$ (\textdegree) & $l_{Z}$ (mm) & $l_{RDP}$ (mm) \\
\midrule
Generic 1 & B  & $F_{\mathrm{opt}}$ & $24.2$ & $20.4$ & $20.0$ & $18.7$ & $-3.3$ & --- \\
Generic 1 & B  & $F_{\mathrm{SF}}$  & $18.4$ & $15.4$ & $16.6$ & $16.1$ & $-3.3$ & --- \\
Generic 2 & S  & $F_{\mathrm{opt}}$ & $9.3$ & $-11.0$ & $-14.9$ & $-12.6$ & $4.9$ & --- \\
Generic 2 & S  & $F_{\mathrm{SF}}$  & $-14.8$ & $4.8$ & $8.6$ & $9.2$ & $3.7$ & --- \\
Generic 3 & RB & $F_{\mathrm{opt}}$ & $18.4$ & $24.0$ & $-3.6$ & $11.1$ & $-4.7$ & $6.7$ \\
Generic 3 & RB & $F_{\mathrm{SF}}$  & $18.4$ & $24.0$ & $-3.6$ & $11.1$ & $-4.7$ & $6.7$ \\
\midrule
Patient 1 & B  & $F_{\mathrm{opt}}$ & $19.2$ & $14.8$ & $24.6$ & $24.3$ & $-3.9$ & --- \\
Patient 1 & B  & $F_{\mathrm{SF}}$  & $17.7$ & $12.1$ & $22.4$ & $22.0$ & $-3.5$ & --- \\
Patient 2 & S  & $F_{\mathrm{opt}}$ & $5.4$ & $-14.4$ & $-9.8$ & $-12.5$ & $3.6$ & --- \\
Patient 2 & S  & $F_{\mathrm{SF}}$  & $5.9$ & $0.8$ & $-10.0$ & $-11.5$ & $1.3$ & --- \\
Patient 3 & RB & $F_{\mathrm{opt}}$ & $18.8$ & $19.5$ & $-4.3$ & $7.6$ & $-3.9$ & $3.5$ \\
Patient 3 & RB & $F_{\mathrm{SF}}$  & $18.8$ & $19.5$ & $-4.3$ & $7.6$ & $-3.9$ & $3.5$ \\
\bottomrule
\end{tabular*}
\begin{tablenotes}[flushleft]
\footnotesize
\item Angular variables are roll/pitch of the left/right resection planes; \(l_{Z}\) is the vertical donor offset and \(l_{RDP}\) the intermediate-cut offset. A dash indicates a variable not present for single-segment defects.
\end{tablenotes}
\end{threeparttable}
\end{table*}



\end{document}